\documentclass[10pt,twocolumn,letterpaper]{article}

\usepackage[pagenumbers]{wacv} % To force page numbers, e.g. for an arXiv version

\usepackage{graphicx}
\usepackage{amsmath}
\usepackage{amssymb}
\usepackage{booktabs}
\usepackage{multirow}
\usepackage{xcolor}
\usepackage{makecell}

\usepackage[pagebackref,breaklinks,colorlinks]{hyperref}

\usepackage[capitalize]{cleveref}
\crefname{section}{Sec.}{Secs.}
\Crefname{section}{Section}{Sections}
\Crefname{table}{Table}{Tables}
\crefname{table}{Tab.}{Tabs.}

\def\wacvPaperID{836} % *** Enter the WACV Paper ID here
\def\confName{WACV}
\def\confYear{2025}

\newcommand{\qcr}[1]{{\fontfamily{qcr}\selectfont{#1}}}
\DeclareMathOperator{\softmax}{Softmax}
\DeclareMathOperator{\attention}{Attention}

\begin{document}

\title{Towards Localized Fine-Grained Control for Facial Expression Generation}

%\author{Tuomas Varanka\\
%University of Oulu\\
%{\tt\small tuomas.varanka@oulu.fi}
%\and
\author{
	Tuomas Varanka$^1$ ~~~ Huai-Qian Khor$^1$ ~~~ Yante li$^1$ ~~~ Mengting Wei$^1$\\ ~~~ Hanwei Kung$^2$ ~~~ Nicu Sebe$^2$ ~~~ Guoying Zhao$^1$ \\
	$^1$ University of Oulu ~~~~~~~ $^2$ University of Trento\\
    {\tt\small tuomas.varanka@oulu.fi}
}
%Teaser
\twocolumn[{
\renewcommand\twocolumn[1][]{#1}
\maketitle
\begin{center}
    \centering
    \vspace*{-.8cm}
    \includegraphics[width=1\textwidth]{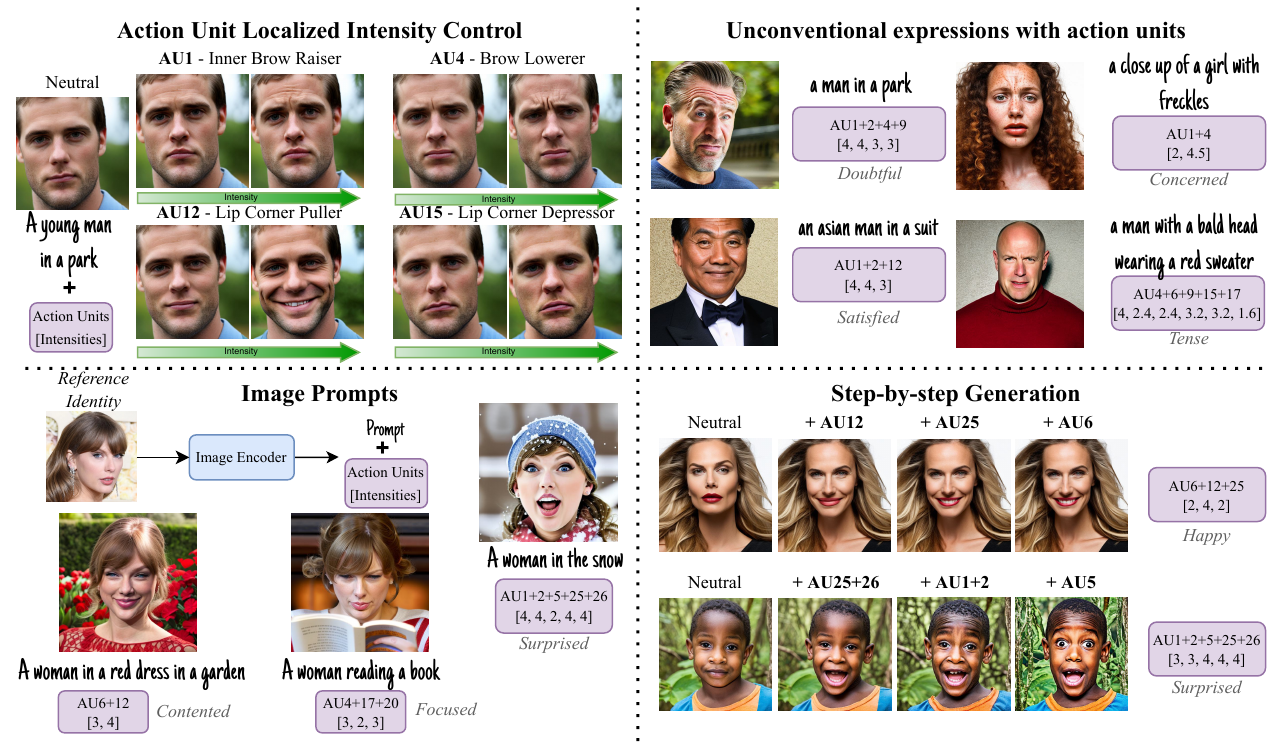}
    \vspace*{-.6cm}
    \captionof{figure}{The proposed method, FineFace, enables precise control over individual muscle movements of the face. By combining several Action Units (AUs), FineFace can generate complex and nuanced facial expressions. Our adapter architecture-based approach enables integration with image prompts using IP-Adapter \cite{ipadapter}.}
\label{fig:teaser}
\end{center}
}]

\maketitle

%%%%%%%%% ABSTRACT
\begin{abstract}
    Generative models have surged in popularity recently due to their ability to produce high-quality images and video. However, steering these models to produce images with specific attributes and precise control remains challenging. Humans, particularly their faces, are central to content generation due to their ability to convey rich expressions and intent. Current generative models mostly generate flat neutral expressions and characterless smiles without authenticity. Other basic expressions like anger are possible, but are limited to the stereotypical expression, while other unconventional facial expressions like doubtful are difficult to reliably generate.

    In this work, we propose the use of AUs (action units) for facial expression control in face generation. AUs describe individual facial muscle movements based on facial anatomy, allowing precise and localized control over the intensity of facial movements. By combining different action units, we unlock the ability to create unconventional facial expressions that go beyond typical emotional models, enabling nuanced and authentic reactions reflective of real-world expressions. The proposed method can be seamlessly integrated with both text and image prompts using adapters, offering precise and intuitive control of the generated results. Code and dataset are available in \url{https://github.com/tvaranka/fineface}.
    
   %The human face, a dynamic canvas conveying nuanced emotions and unspoken messages, remains a captivating centerpiece in visual media. Recent generative methods have yet to fully master the detailed customization, indicating a need for further development in this area to achieve optimal precision and personalization in facial generative applications. The main objective of the research is to develop powerful AU-conditioned generation methods for efficient and precise generation and editing of facial expression images and videos with high fidelity and the capability to generate co-occurred AUs with different intensities, subtleness and fleetness, and spatial-temporal consistently among multi-modal input information, which is a very new and rarely explored topic. This can open new possibilities for face generation, amplifying the richness of human-human and human-AI interactions.

   %present a disentagled representation of individual facial movements based on physical anatomy of the human face.
\end{abstract}

%Idea for writing from https://arxiv.org/pdf/2403.18978:
%In abstract only briefly mention what you are doing, leave it as a cliff hanger for readers
%Give more information in introduction, but again leave some stuff in the method section
%Contributions: 1. Showcase that AUs give more control. 2. Propose effective tool for conditioning with continuous multi-labels. 3. Integration and applicability to real world use cases

%%%%%%%%% BODY TEXT
\section{Introduction}
\label{sec:intro}
The advent of T2I (text-to-image) generative diffusion models \cite{ddpm, stablediffusion} has marked a significant milestone in content generation \cite{animate_anyone, emo, null_text, sora}, offering unprecedented tools for creativity and expression. These state-of-the-art technologies are starting to be used in the production of film and artistic pieces \cite{sora_first_impressions}, where the nuanced portrayal of facial expressions plays a pivotal role. However, despite their sophistication, current models exhibit a notable deficiency: the lack of localized, fine-grained control over the generation of facial expressions. This shortcoming restricts the breadth of artistic expression for creating nuanced emotional conveyance, a critical aspect for immersive storytelling.

ControlNet \cite{controlnet} enables users to add additional control signals to a T2I (text-to-image) model in the form of depth, human skeleton pose and canny edges to name a few. It has been widely accepted by the community as it allows for more control from the users, which is difficult with just a single text prompt. Controlling the identity of a generated person by inserting a specific face has attracted a large amount of work very recently \cite{instantbooth, instantid, consistentid, photomaker, ipadapter, flashface}. Despite these efforts, users are still stuck with neutral or a generic smile for their generations. Recent works \cite{unifiedface, granular_expression} improve on this by enabling wider choice of facial expressions over the basic facial expression like \textit{happy} and \textit{sad}. However, they still lack the ability to provide localized and intensity-specific control.

%A recent work \cite{unifiedface} changes this by introducing a new 3D emotion space which can be used for conditioning. Although the model is capable of creating 15 different compound expressions, the outputs are typically exaggerated, lack localized control and interpretability. Due to the fine-tuning of the entire UNet, the model has lost its capability of following prompts that describe scenes other than portraits of faces.

\begin{figure}[t]
    \includegraphics[width=0.5\textwidth]{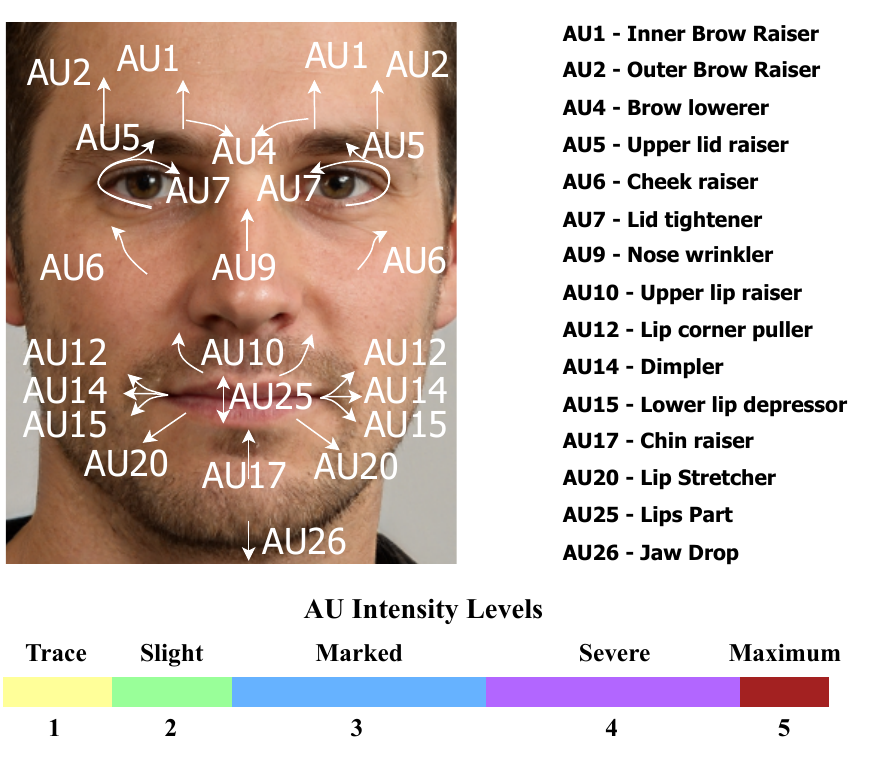}
    \caption{Display of a selection of different action units and the intensity scale. Figure repurposed from \cite{meb}. For a complete collection of AUs with videos see \cite{facs_cheat_sheet}.}
    \label{fig:aus}
\end{figure}

Stepping back and reconsidering the conditioning inputs for facial expression control, the condition should be interpretable, easy to use and enable precise localized control with adjustable intensity. The basic six emotions \cite{basix_six}, compound emotions \cite{compound}, valence-arousal emotion space \cite{valence_arousal}, and 3DMMs \cite{3dmm} are considered, but we find that AUs (Action Units) \cite{facs} best match the set requirements. AUs encode facial muscle movements, offering localized control with adjustable intensity. \Cref{fig:aus} displays a set of different AUs. Since AUs objectively represent muscle movements, they are not constrained by emotional labels. This enables the generation of non-emotion related facial expressions like focused, duckface, squinting, confused, skepticism and more. Further, emotional labels such as anger can vary largely depending on the culture, but also the situation and context. By combining multiple AUs, both simple and intricate facial expressions can be accurately generated.

AUs offer granular localized control, however, their representation as continuous multi-labels presents novel challenges when integrating them into T2I models \cite{ccgan}. We propose an \textit{AU Encoder} that addresses the continuity of intensity values and the multi-label nature of AUs by learning the interactions between different AUs. To address the issue of limited data availability, a training strategy \cite{ccgan} that mitigates this is implemented with an extended dataset using automatic annotation tools \cite{libreface, blip-2}. By avoiding the direct training of the original T2I model weights and instead using adapters, the method effectively follows text prompts while accurately adhering to AU conditions, and can be seamlessly integrated with image adapters \cite{ipadapter}, as shown in \cref{fig:teaser}.

Our contributions are as follows:
\begin{itemize}
    \item Introduction of an approach utilizing Action Units (AUs) to precisely control facial expressions for T2I generation. By leveraging AUs, the method enables localized and fine-grained manipulation of facial muscles, facilitating the generation of diverse and nuanced facial expressions.
    \item Design of an AU encoder that effectively translates input commands into intricate facial gestures. The encoder enables continuous scale and combination of several action units together.
    \item Seamless integration of the proposed method, FineFace, into real-world applications by enabling both text and image prompts while accurately adhering to AU conditions.
\end{itemize}

\section{Related Work}

\paragraph{Facial expression generation and editing}
One of the earlier works on photorealistic facial expression editing is with StarGan \cite{stargan}, which is able to edit facial images with basic expressions. GANimation \cite{ganimation} offers a more fine-grained approach to editing by using AUs. This is further improved by using a patch attentive GAN with a newly proposed discriminator in \cite{zhao2021action}. AUs are also used by ICface \cite{icface} for controllable facial expressions in facial reenactment. GANmut \cite{ganmut} learns compound facial expressions from only basic labels for wider options of facial expressions that can be used in the editing. More recently, EmoStyle \cite{emostyle} uses a StyleGAN2 \cite{stylegan2} as a basis for photorealistic outputs and the valence \& arousal -space for granular editing.

In contrast to editing, where modifications are made to an existing image, generation involves creating the output identity either based on the posterior distribution or from a conditioned prompt. This process requires additional hallucination of content to complete the scene. Stable Diffusion \cite{stablediffusion} is capable of generating scenes from text prompts with basic facial expressions. To further increase the variety of possible facial expressions Paskaleva \etal \cite{unifiedface} learn a 3-dimensional emotion space by using a combination of valence \& arousal, action units and learning using GANmut \cite{ganmut}. Liu \etal \cite{granular_expression} uses a dictionary of 135 emotional words that are used to query a database, from which the facial expressions are transferred to the generated result. Compared to these works, we enable localized and adjustable intensity control of facial expressions. A concurrent work, InstructAvatar \cite{instructavatar} creates talking faces from input images and offers control with textual descriptions of AUs, however with less intensity control.

\paragraph{Diffusion model conditioning}
\vspace{-2mm}
Diffusion models \cite{ddpm} have become prominent in generating high-quality images by iteratively refining a noisy image towards a target through a denoising process. Conditioning these models to control specific aspects of the generated output is an active research area \cite{controlnet, animate_anyone, cameractrl, camvig, emo, unifiedface, collaborative_diffusion}. Text conditioning, popularized and made accessible by Stable Diffusion \cite{stablediffusion}, is one of the earliest and most versatile tools for T2I generation. Controlnet \cite{controlnet} is a pioneering work that introduced image conditioning for diffusion models, enabling the use of human skeleton poses, depth maps and more. However, while it excels in providing structural control of the image, it struggles to preserve fine details. IP-Adapter \cite{ipadapter} instead proposes a novel conditioning mechanism that enables non-structural image prompting. For maintaining pixel level details, AnimateAnyone \cite{animate_anyone} utilizes a ReferenceNet to effectively incorporate high-resolution reference images into the generation process.

Other approaches other than text and image include camera controls such as those proposed in \cite{cameractrl, camvig}, which allow users to change the camera position to different angles. Another common conditioning input is audio as in \cite{emo}, which allows users to create talking and singing avatars. Several other custom conditions exist such as emotion \cite{unifiedface, granular_expression} or mask \cite{collaborative_diffusion}.

\section{Control for Facial Expressions}
There is a need to control facial expressions in generative models, but determining the appropriate conditioning signal is crucial. In this section, we analyze several potential options and present our argument for the most effective choice. The conditioning signal should satisfy the following criteria: 1) enable localized edits, 2) be adjustable in intensity, and 3) remain interpretable.

\paragraph{Emotional models}
\vspace{-2mm}
The most likely option that comes to mind initially for facial expression control is the set of basic six emotions\cite{basix_six} (\textit{happy}, \textit{sad}, \textit{surprise}, \textit{angry}, \textit{disgust} and \textit{fear}). Although this is common to most people, it is limited in the number of expressions and finer control. Compound expressions \cite{compound} expand the number of categories to 17 (\eg \textit{happily surprised}, \textit{happily fearful}), which is still too limited for as it only applies for emotional facial expressions. Furthermore, different people, situations and cultures may interpret text labels differently. Arousal-valence (AV) \cite{valence_arousal} is a two-dimensional model with continuous values that covers a large amount of possible facial expressions. It provides intensity control, but interpreting which facial muscles are changing with different values is not intuitive as the model lacks localized control of individual parts of the face.

\paragraph{Blendshapes and 3DMMs}
\vspace{-2mm}
In contrast to the emotional models, 3D Morphable Models (3DMMS) \cite{3dmm, deca} and blendshapes offer a fully objective approach to facial expressions, parameterized by changes in facial contours rather than emotional labels. Several methods \cite{stylerig, discofacegan, config, ren2024flexible} have utilized 3DMMs as a condition to generate faces with fine-grained control. However, the large number of parameters presents a significant challenge, as only experts are able to manually tune wanted expressions. This complexity arises from the general purpose of 3DMMs, which are also designed for creating facial identities—a capability not necessary for facial expression modification. Despite the large number of parameters, 3DMMS and blendshapes are unable to create fine-grained wrinkles \cite{qian2024gaussianavatars} and extreme facial expressions \cite{pan2023real}.

%To address this, a more streamlined and user-friendly approach is needed. By focusing exclusively on facial expression control, we can reduce the parameter space and simplify the tuning process. This can make the technology accessible to non-experts, broadening its applicability and usability in various domains such as digital art, virtual reality, and interactive media. Additionally, we propose leveraging action units (AUs) as a more intuitive and precise conditioning signal. AUs, which encode specific facial muscle movements, offer localized control with adjustable intensity and can generate a wide range of expressions by combining multiple units. This approach balances ease of use with the ability to produce both basic and complex facial expressions, making it an optimal choice for conditioning generative models.

%An alternative option, yet to be explored for T2I generation, that offers objective facial expressions with localized and fine-grained details with intuitive and easy controls are action units (AUs). 3DMMs also lack the really fine-grained details for creating wrinkles. And if one wants a novel expression they either have to manually change the thousands of points from a 3DMM or find an existing face with such expression fro mwhich the 3DMM can be extracted.

\begin{figure*}
    \centering
    \includegraphics[width=0.99\textwidth]{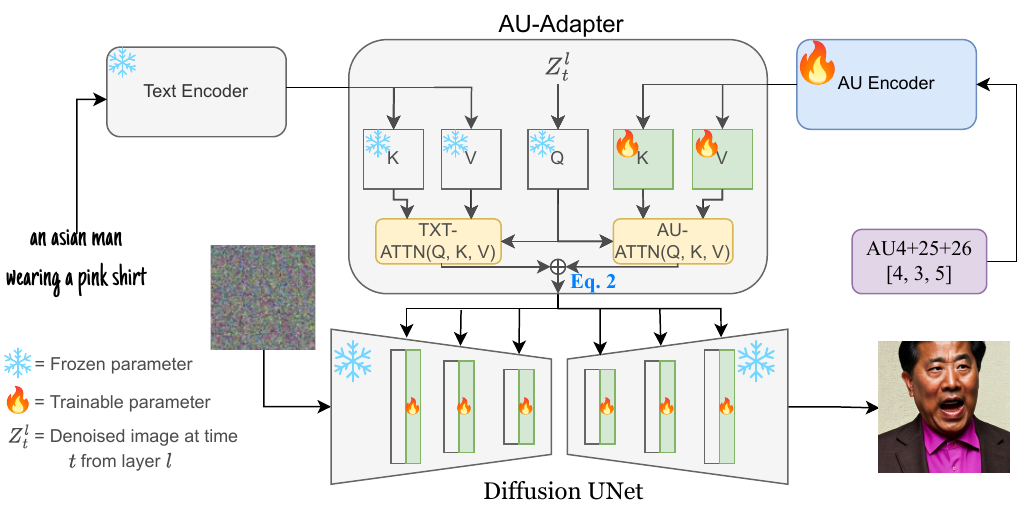}
    \caption{FineFace generates an image based on a text prompt and an AU condition. The AU condition vector is first passed to an AU encoder and subsequently to the AU-Adapter. The output of the AU attention is then added with the existing text attention. In this setup, only the AU encoder and the K and V projection matrices are trainable, while the other layers remain frozen.}
    \label{fig:architecture}
\end{figure*}

\paragraph{Action Units}
\vspace{-2mm}
The Facial Action Coding System (FACS) \cite{facs} provides a precise method for analysing and interpreting facial movements by breaking down face movements into individual muscle movements known as Action Units (AUs), see \cref{fig:aus}. For example, a facial expression like ``happily surprised'' can be represented by Inner Brow Raiser (\qcr{AU1}), Outer Brow Raise (\qcr{AU2}), Cheek Raiser (\qcr{AU6}), and Lip Corner Puller (\qcr{AU12}). There are a total of 30 atomic action units and additional 14 reserved for head movements, gazes and other miscellaneous actions \cite{facs}. Additionally, each AU has specific descriptions of appearance and geometry changes with different intensity levels, ranging from 0 (not present),  A (trace) to E (maximum) or in the numerical range of [0, 5], see bottom of \cref{fig:aus}. The specificity and clarity of AUs offer unparalleled control over facial expressions, providing users with an interpretable and accessible set of controls. 

%A large benefit of using AUs is the precise localized control with adjustable intensity. For editing applications, most of the facial expression may be good, but the just a specific part is not good. With AUs everything else can be kept the same, while only changing the not good part.

%A downside of action units is that they are not immediately intuitive to non-experts compared to basic facial expressions. However, with the help of a visual guide (see \cref{fig:aus}) non-export users are able to quickly learn the basics. As AUs are highly localized and only control certain parts of the face, combining them is relatively intuitive. Furthermore, it is possible to create existing combinations like \qcr{AU6+12} (=happiness) for even easier use.

\section{Method}
As shown in \cref{fig:architecture}, the proposed framework contains an AU encoder and an adapter to the stable diffusion model \cite{stablediffusion} that takes in the features from the AU encoder. Only the AU encoder and adapter are made trainable. This design ensures minimal changes to the strong priors of the base diffusion model, enabling strong coherence to text prompts.

\subsection{Preliminaries}

\paragraph{IP-adapter\cite{ipadapter}}
enhances text-to-image diffusion models by integrating them with image prompt capabilities. At the core of this method is a decoupled cross-attention mechanism that processes text and image features separately, thereby maintaining the integrity of the pre-trained model while enabling the addition of image prompts. Cross-attention works by having the $Q$ and $K, V$ features in
\begin{equation}
    \attention(Q, K, V) = \softmax \left(\frac{QK^T}{\sqrt{d}} \right)V,
\end{equation}
come from different sources, as opposed to self-attention where all the features $Q, K, V$ come from the same source. For each cross-attention new Key and Value projection matrices are employed for the image prompt features, while the Query comes from the original cross-attention. The outputs of each new decoupled cross-attention are added to the original cross-attention with a scaling factor as follows:
\begin{align}
    \mathbf{Z} = & \attention(Q_{noise}, K_{text}, V_{text}) \nonumber \\ 
               & + \lambda_{img} \cdot \attention(Q_{noise}, K_{img}, V_{img}),
    \label{eq:ip_adapter}
\end{align}
where $\{Q, K, V\}_{source}$ refers to the source of the feature tensor. This lightweight adapter can be applied to existing diffusion models without the need for extensive retraining or computing resources. 

%\paragraph{LoRA \cite{lora}}
%Low-Rank Adaptation, is a technique originally developed for large language models, but later applied to diffusion models. It modifies pre-trained models by introducing low-rank matrices that adapt the weights of the model's layers, rather than retraining the entire network. This approach allows for efficient fine-tuning of large models, as it only adjusts a small subset of parameters. It enables the model to learn specific features or styles with fewer computational resources, maintaining the overall structure and knowledge of the pre-trained network while adapting to new tasks with high fidelity.

\subsection{Architecture}
The design of the architecture is propelled forward by two goals: 1. Efficient injection of AU information to the model 2. Retaining capabilities of the base T2I model. An approach used by \cite{unifiedface} is to project a triplet with values $[-1, 1]$ representing a three dimensional emotional space, to the CLIP space. This projected feature can then be injected into the text prompt directly by replacing one of the tokens. However, this approach limits controllability due to its reliance on the guidance scale \cite{cfg} with the text and diminishes the base model's capabilities by fine-tuning the entire network, despite using prior preservation loss \cite{dreambooth}.

%This appro

Instead, we use IP-Adapter \cite{ipadapter}. Although the original IP-Adapter is developed for image conditions, it takes in arbitrary features as its inputs. By employing an AU encoder that projects the AUs into the feature space of the adapter, the IP-adapter can be used. From \cref{eq:ip_adapter} the features $K_{img}$ and $V_{img}$ can be simply replaced with $K_{AU}$ and $V_{AU}$, that are obtained from the cross-attention's corresponding projection matrices which take in the features from the AU encoder. This enables the injection of AU features into the pre-trained T2I model with minimal disturbance and using a limited number of parameters. Similarly as found later in \cite{ipadapter}, we also use a LoRA \cite{lora} to further enhance the results as we find that the use of IP-adapters alone is not sufficient for capturing the complexity of AUs.

\subsection{Continuous Multi-Label Conditioning}
AUs can be represented for a single instance as a vector $\mathbf{y} \in [0, 5]^{n_{au}}$, where $n_{au}$ refers to the number of AUs used and $[0, 5]$ is the range of possible intensity values. For example, with $n_{au} = 6$ and for AUs [\qcr{AU1}, \qcr{AU2}, \qcr{AU4}, \qcr{AU6}, \qcr{AU9}, \qcr{AU12}], the expression \qcr{AU6+12} (happiness), with moderate intensity (3) could be represented as $[0, 0, 0, 3, 0, 3]$, where the 0s refer to other AUs that are not active. Compared to typical class conditions that only have a single discrete value associated with them, AUs are not only continuous, but multi-label. This makes learning the entire distribution extremely challenging as most labels do not exist within the training data \cite{ccgan}. AUs commonly occur in combination with other AUs rather than independently, requiring the model to learn how to disentangle and isolate the effects of different AUs.

The model should be capable of learning the continuous nature of the labels, enabling smooth transitions across different intensity levels. Additionally, because most combinations of AUs are not present in the training data, the model needs to learn to interpolate between the existing combinations to ensure that individual AUs interact cohesively and produce accurate facial expressions. This task is further complicated by the fact that combinations of different AUs yield distinct visual results; for example, \qcr{AU4} in \qcr{AU1+4} appears different from \qcr{AU4} alone. Therefore, simple interpolations of individual AUs are insufficient; the model must learn the interactions between the AUs to generate realistic and precise expressions.

\begin{figure*}
\begin{center}
    \setlength{\tabcolsep}{1pt}
    \begin{subfigure}{1.02\textwidth}
    \hspace{-0.2cm}
    \begin{tabular}{*{14}{c}}
        \rotatebox{90}{\footnotesize{\makecell{FineFace \\(Ours)}}} &
        \includegraphics[width=0.0714\textwidth]{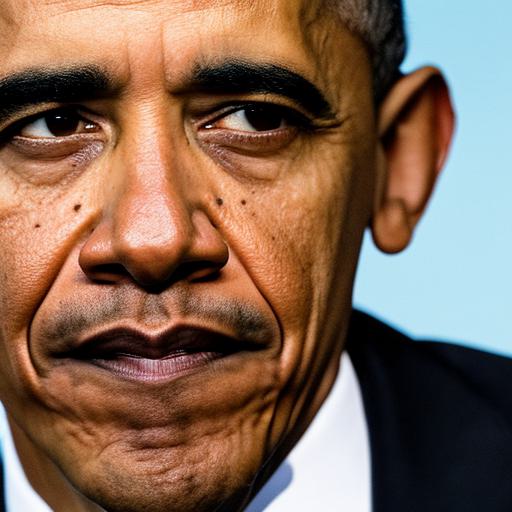} &
        \includegraphics[width=0.0714\textwidth]{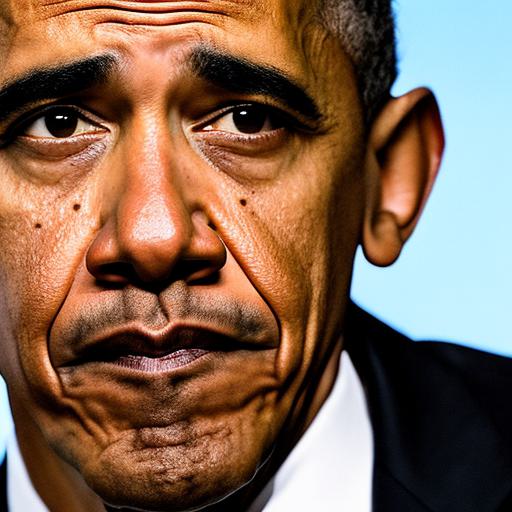} &
        \includegraphics[width=0.0714\textwidth]{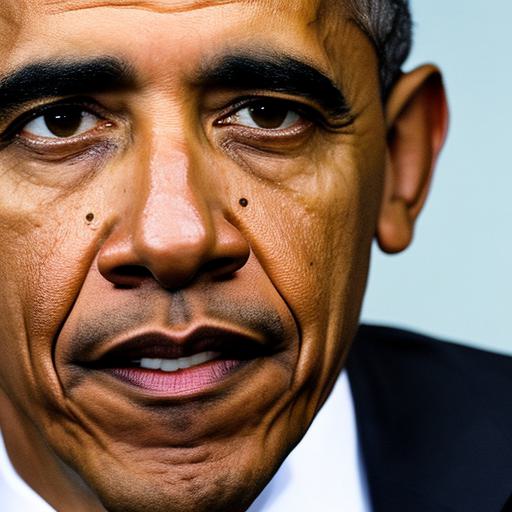} &
        \includegraphics[width=0.0714\textwidth]{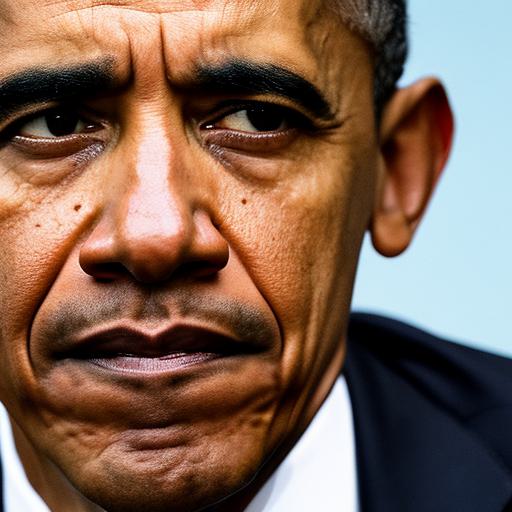} &
        \includegraphics[width=0.0714\textwidth]{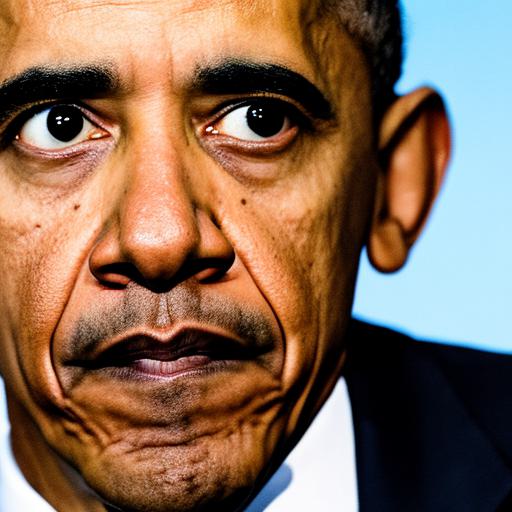} &
        \includegraphics[width=0.0714\textwidth]{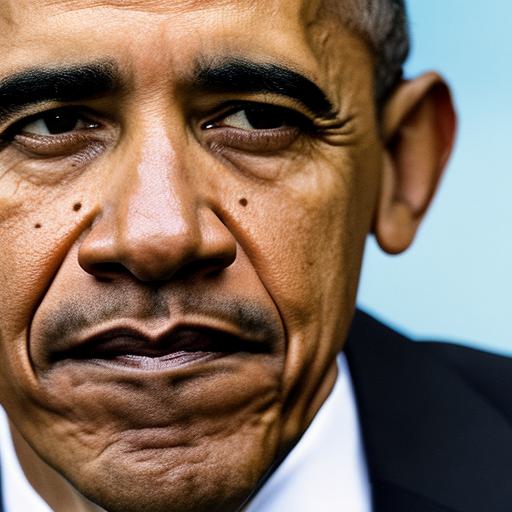} &
        \includegraphics[width=0.0714\textwidth]{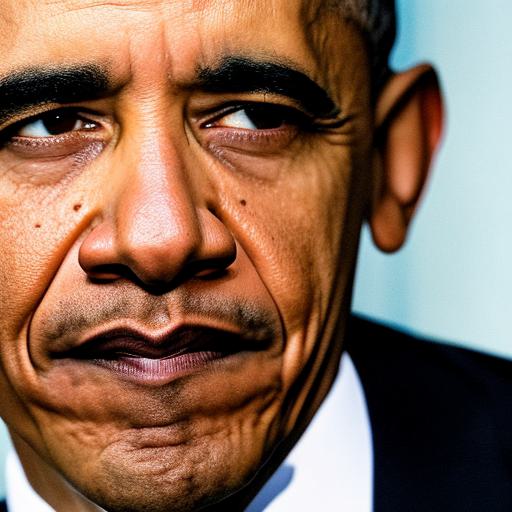} &
        \includegraphics[width=0.0714\textwidth]{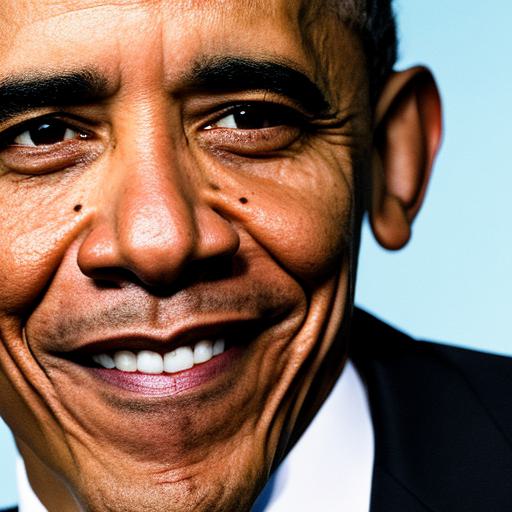} &
        \includegraphics[width=0.0714\textwidth]{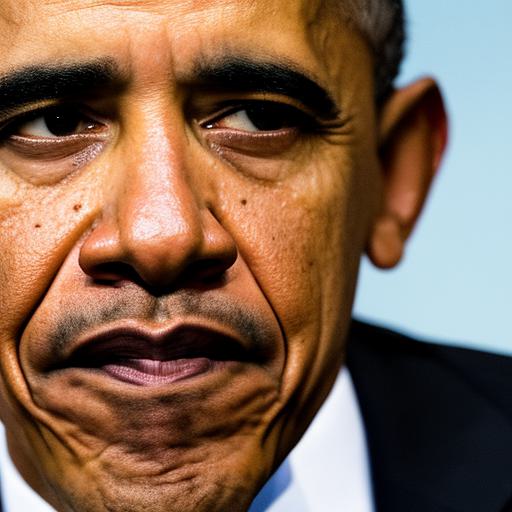} &
        \includegraphics[width=0.0714\textwidth]{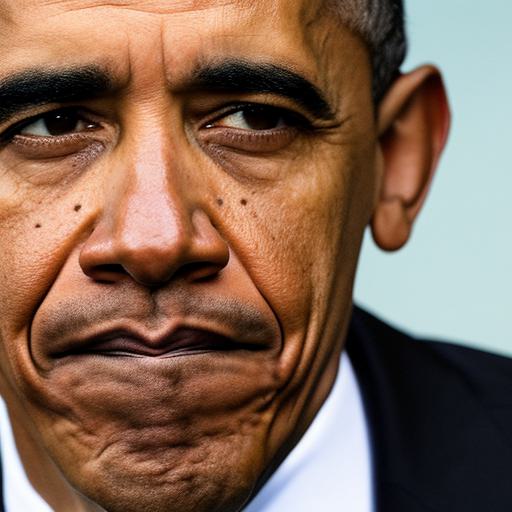} &
        \includegraphics[width=0.0714\textwidth]{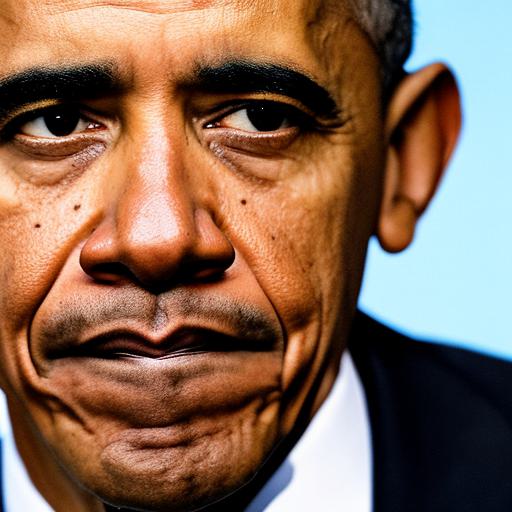} &
        \includegraphics[width=0.0714\textwidth]{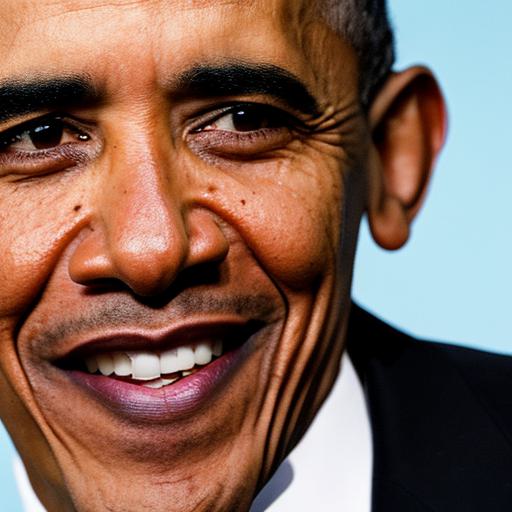} &
        \includegraphics[width=0.0714\textwidth]{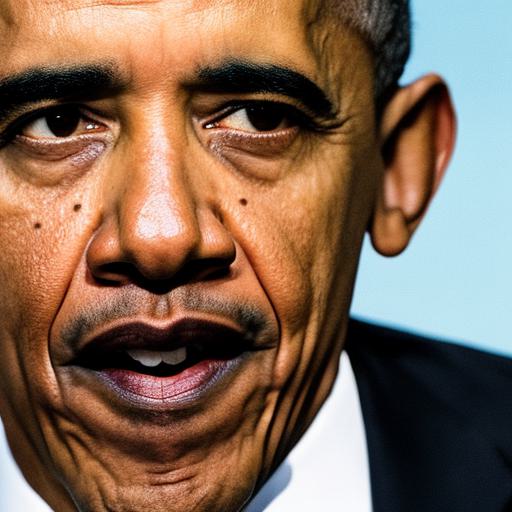} 
        \\
        \rotatebox{90}{\footnotesize{LoRA-AU}} &
        \includegraphics[width=0.0714\textwidth]{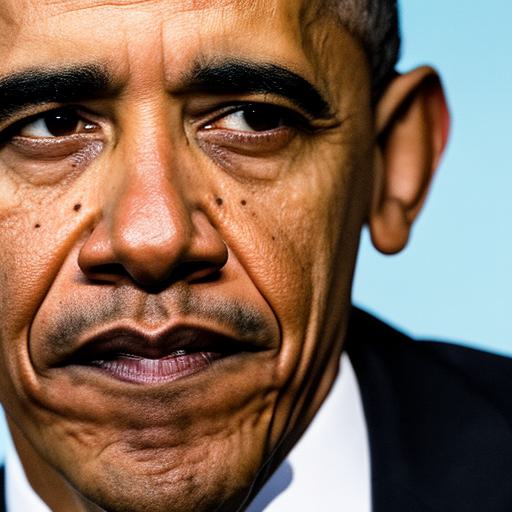} &
        \includegraphics[width=0.0714\textwidth]{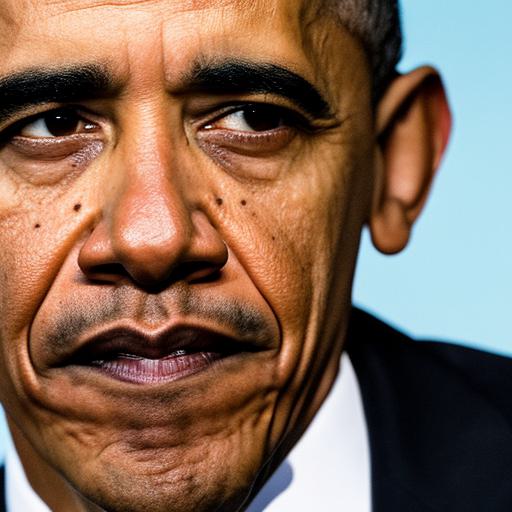} &
        \includegraphics[width=0.0714\textwidth]{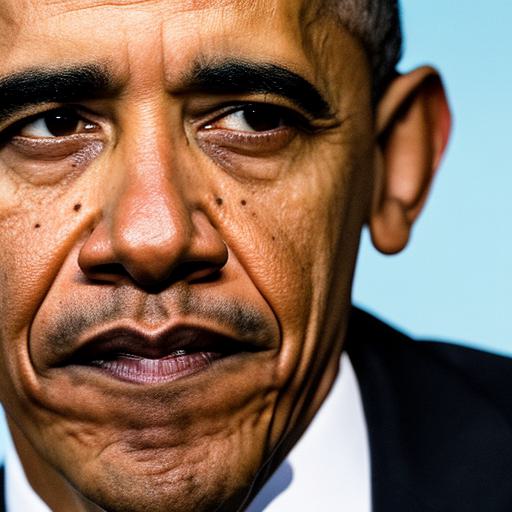} &
        \includegraphics[width=0.0714\textwidth]{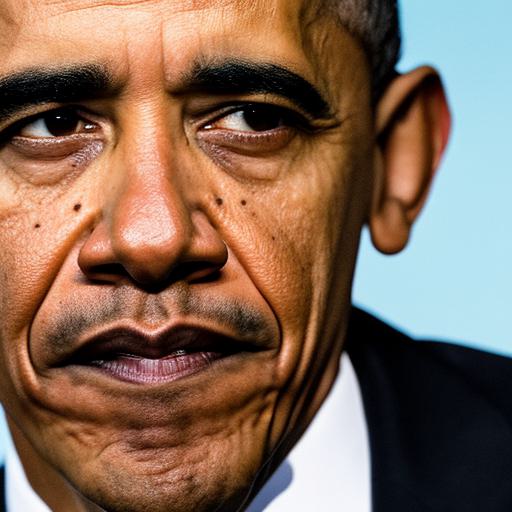} &
        \includegraphics[width=0.0714\textwidth]{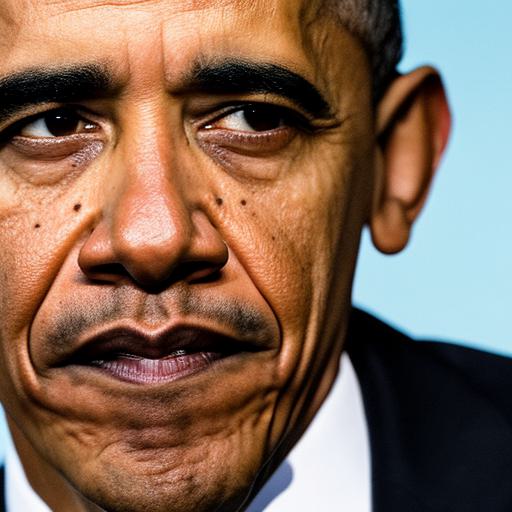} &
        \includegraphics[width=0.0714\textwidth]{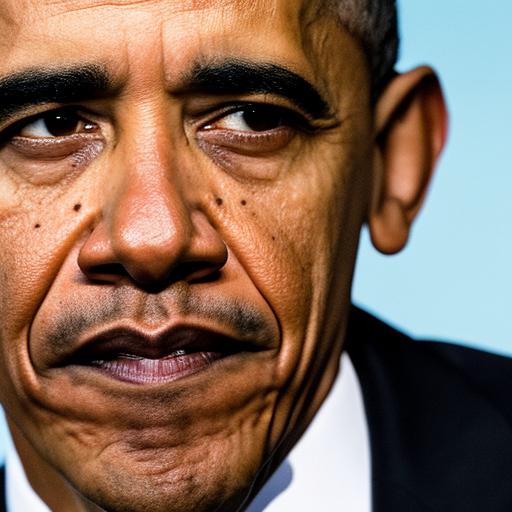} &
        \includegraphics[width=0.0714\textwidth]{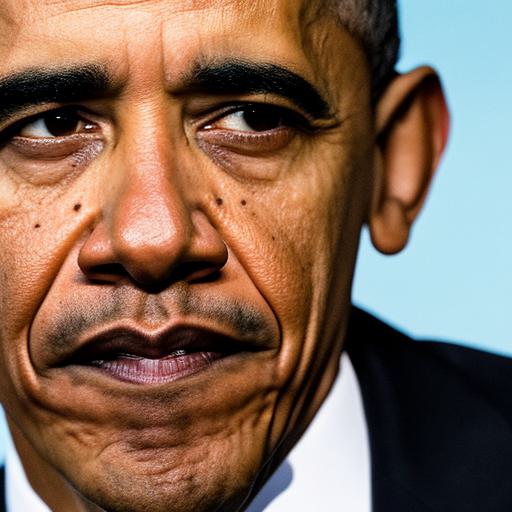} &
        \includegraphics[width=0.0714\textwidth]{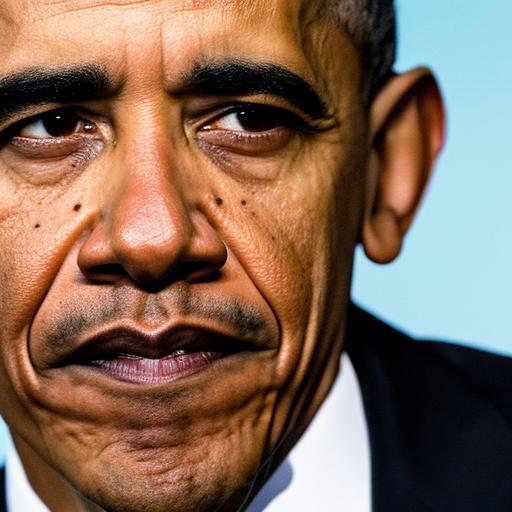} &
        \includegraphics[width=0.0714\textwidth]{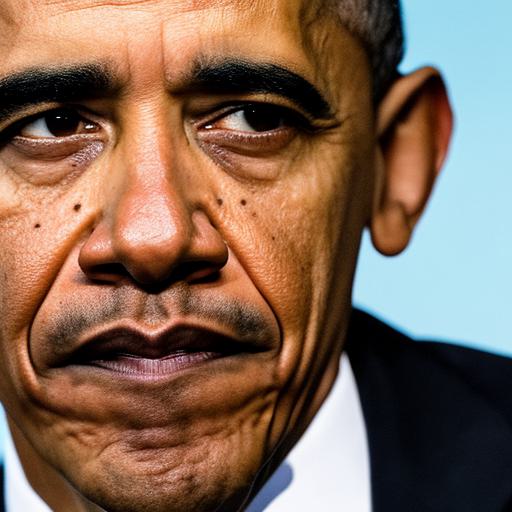} &
        \includegraphics[width=0.0714\textwidth]{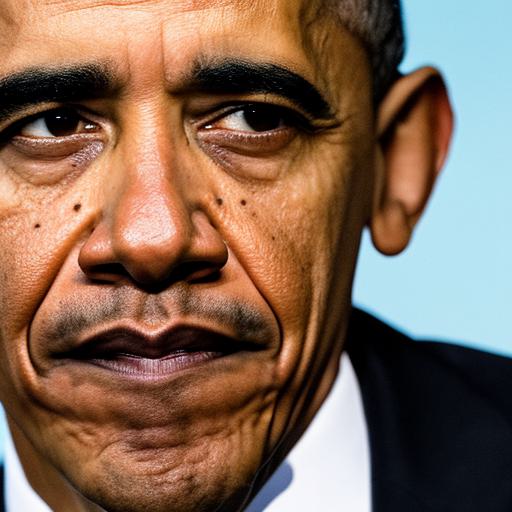} &
        \includegraphics[width=0.0714\textwidth]{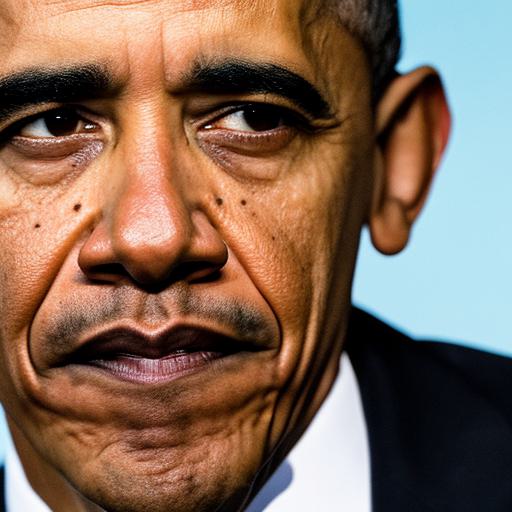} &
        \includegraphics[width=0.0714\textwidth]{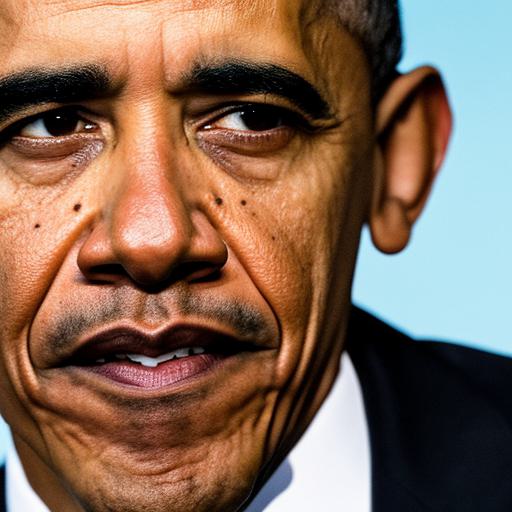} &
        \includegraphics[width=0.0714\textwidth]{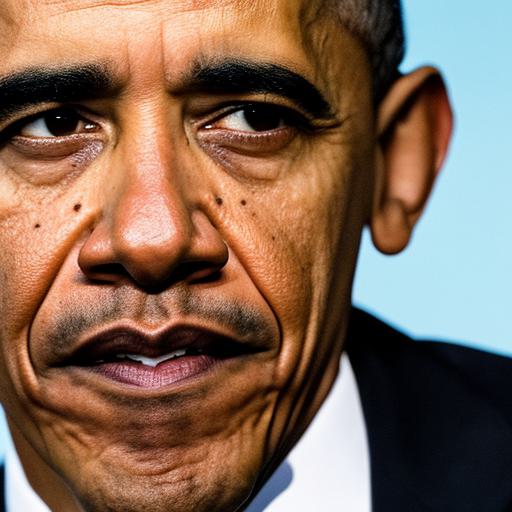}
        \\
        \rotatebox{90}{\footnotesize{LoRA-T}} &
        \includegraphics[width=0.0714\textwidth]{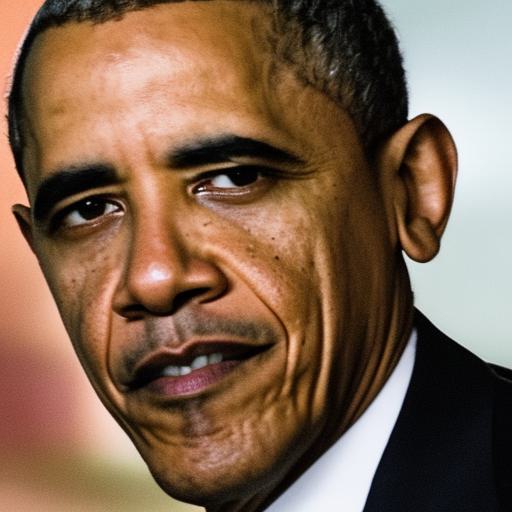} &
        \includegraphics[width=0.0714\textwidth]{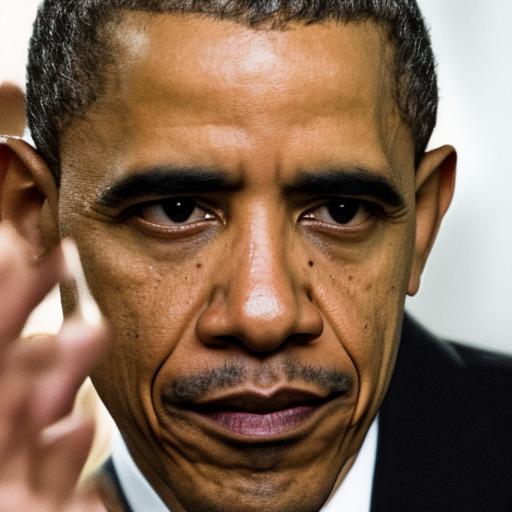} &
        \includegraphics[width=0.0714\textwidth]{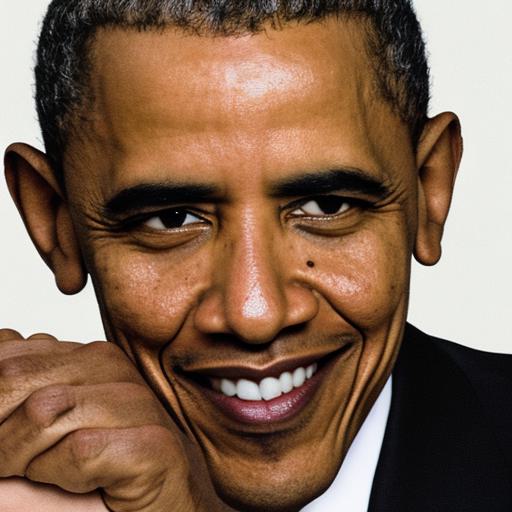} &
        \includegraphics[width=0.0714\textwidth]{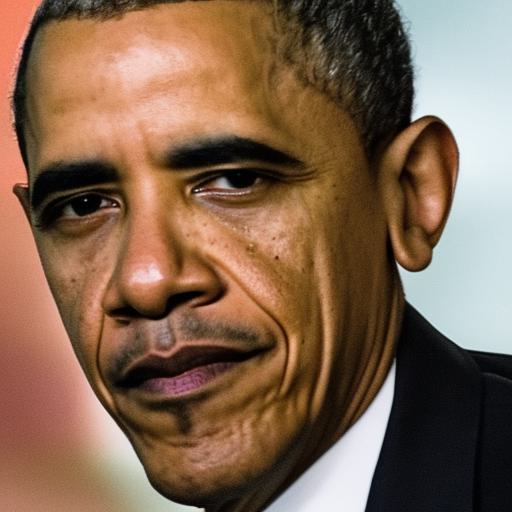} &
        \includegraphics[width=0.0714\textwidth]{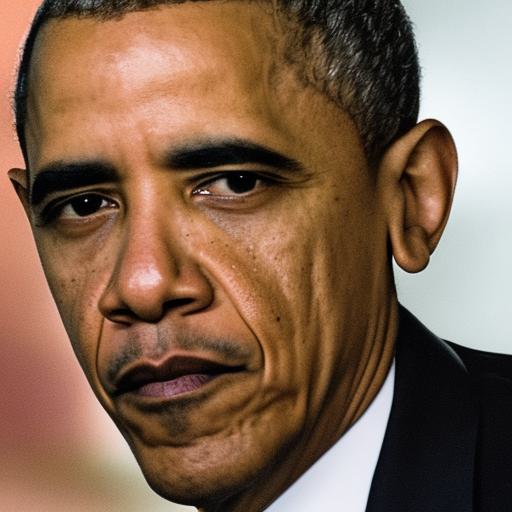} &
        \includegraphics[width=0.0714\textwidth]{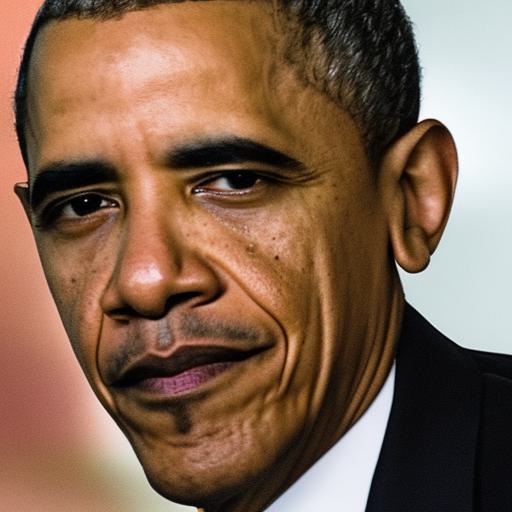} &
        \includegraphics[width=0.0714\textwidth]{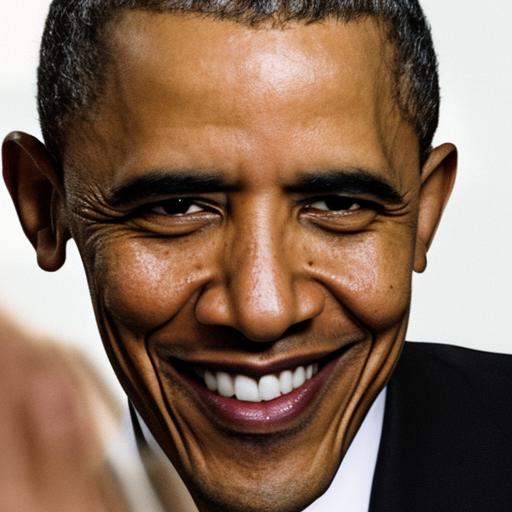} &
        \includegraphics[width=0.0714\textwidth]{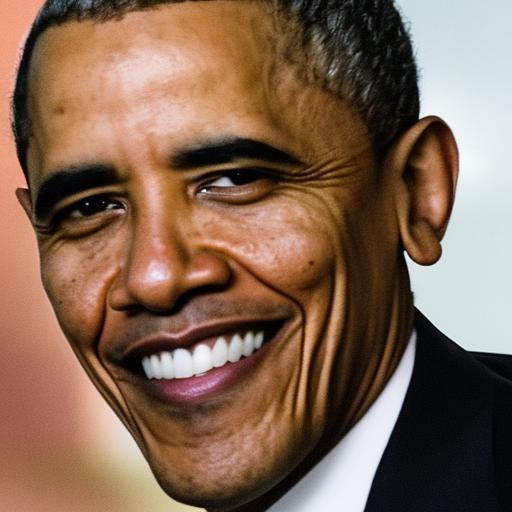} &
        \includegraphics[width=0.0714\textwidth]{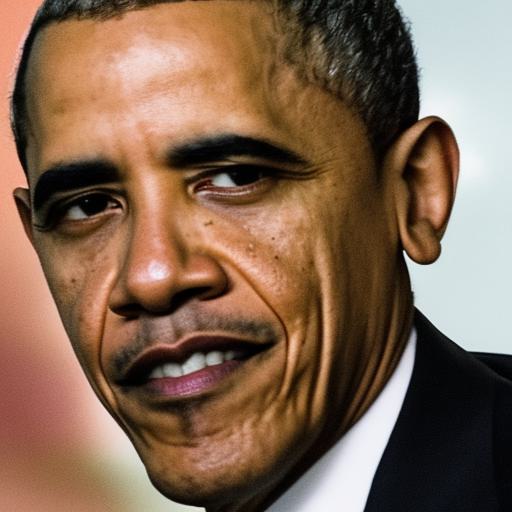} &
        \includegraphics[width=0.0714\textwidth]{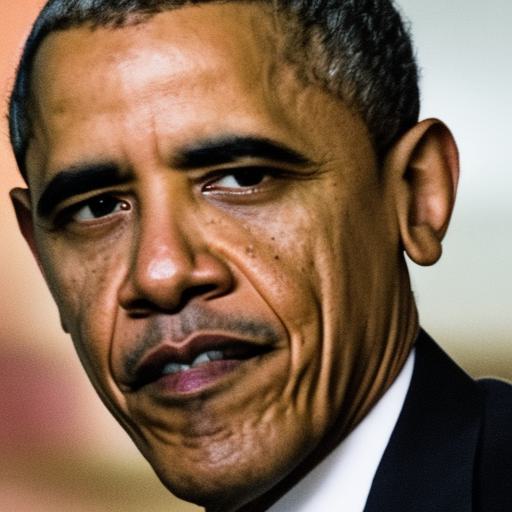} &
        \includegraphics[width=0.0714\textwidth]{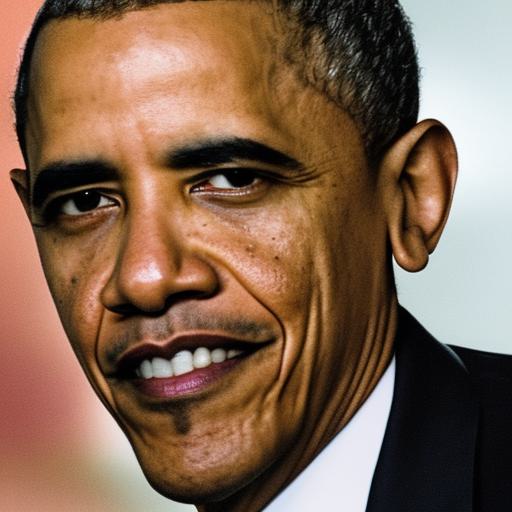} &
        \includegraphics[width=0.0714\textwidth]{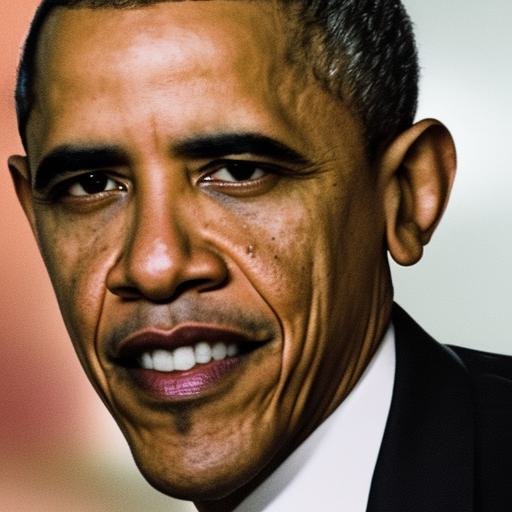} &
        \includegraphics[width=0.0714\textwidth]{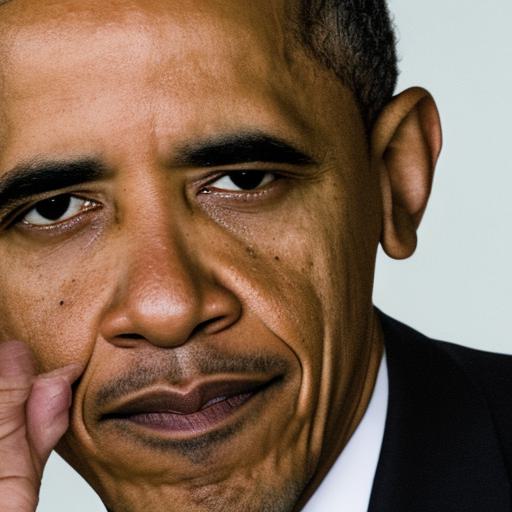}
        \\
        \rotatebox{90}{\footnotesize{DB}} &
        \includegraphics[width=0.0714\textwidth]{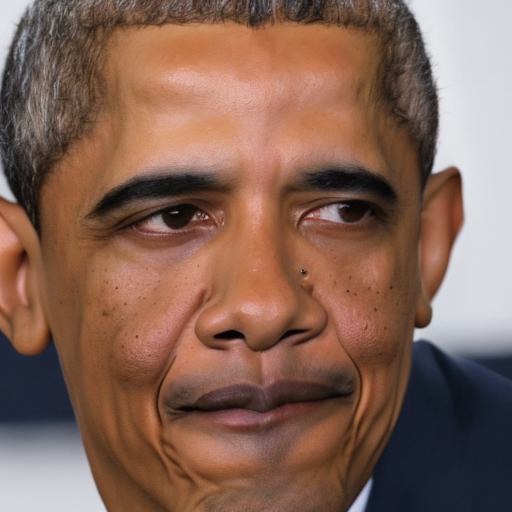} &
        \includegraphics[width=0.0714\textwidth]{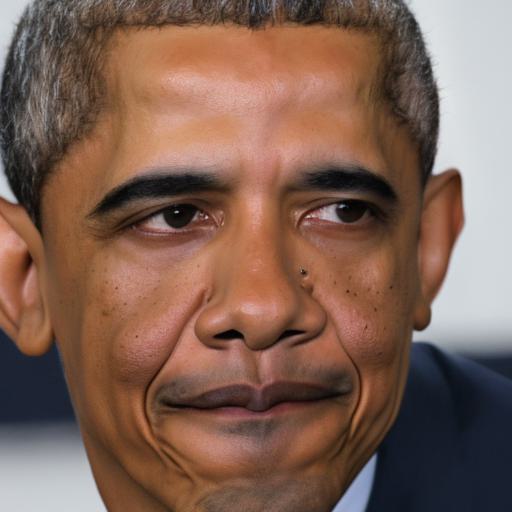} &
        \includegraphics[width=0.0714\textwidth]{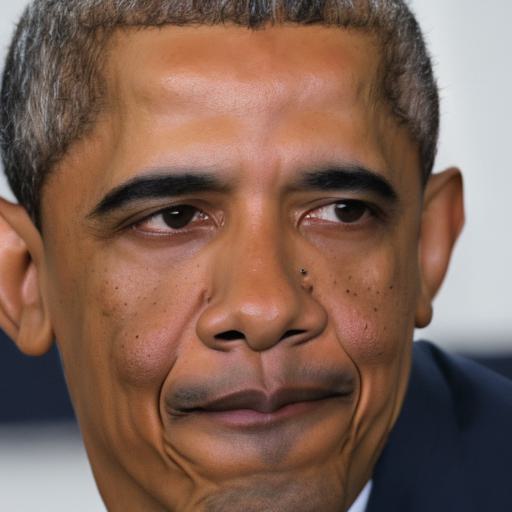} &
        \includegraphics[width=0.0714\textwidth]{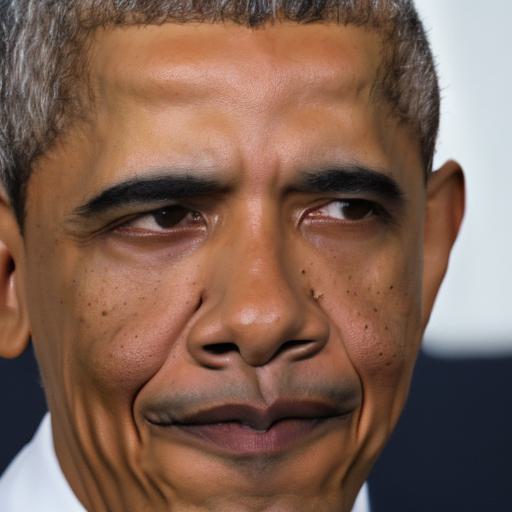} &
        \includegraphics[width=0.0714\textwidth]{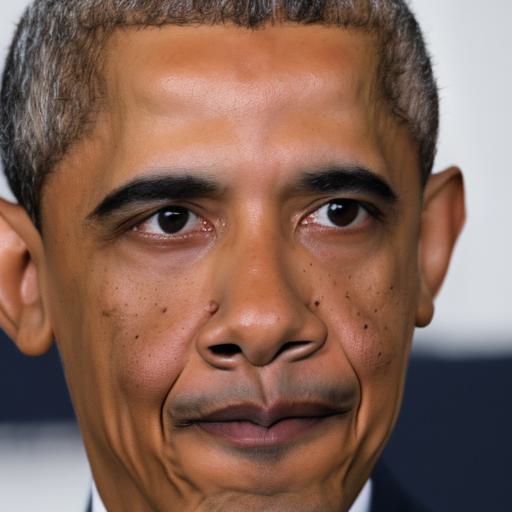} &
        \includegraphics[width=0.0714\textwidth]{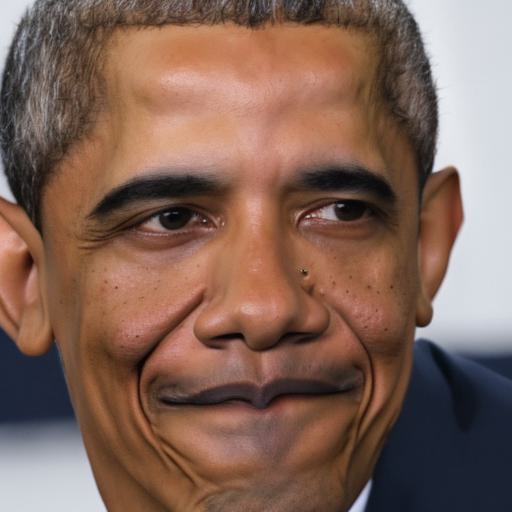} &
        \includegraphics[width=0.0714\textwidth]{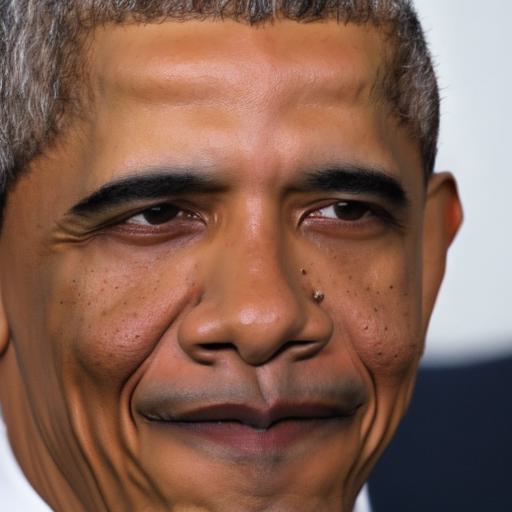} &
        \includegraphics[width=0.0714\textwidth]{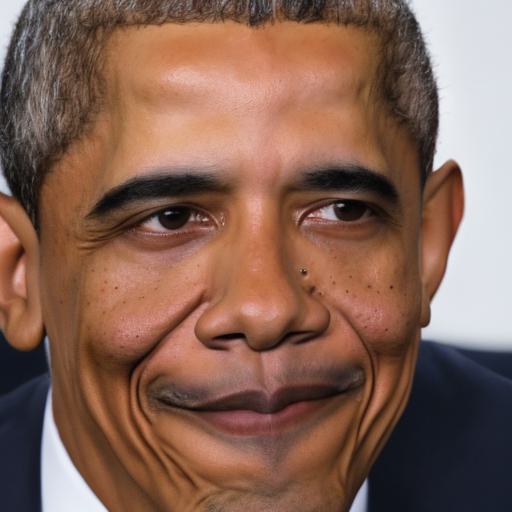} &
        \includegraphics[width=0.0714\textwidth]{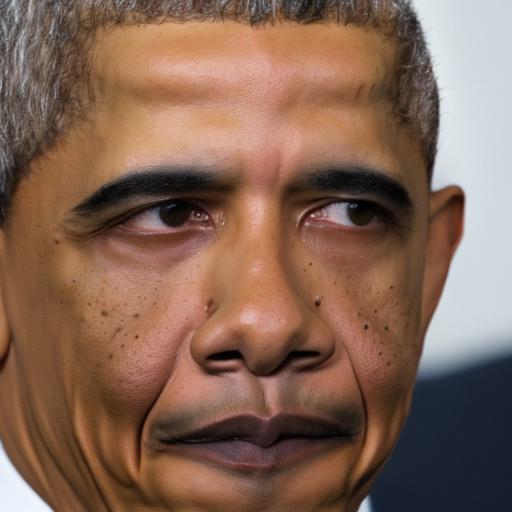} &
        \includegraphics[width=0.0714\textwidth]{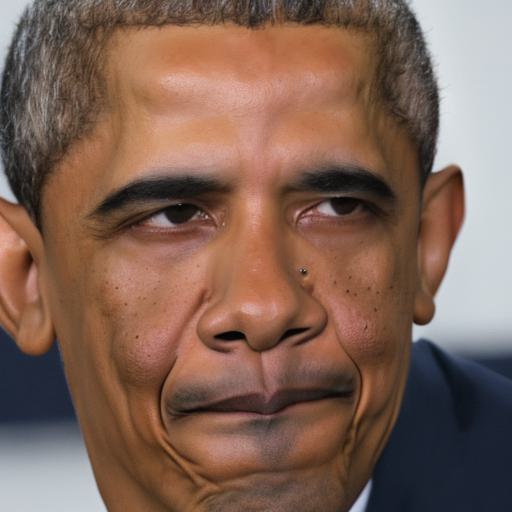} &
        \includegraphics[width=0.0714\textwidth]{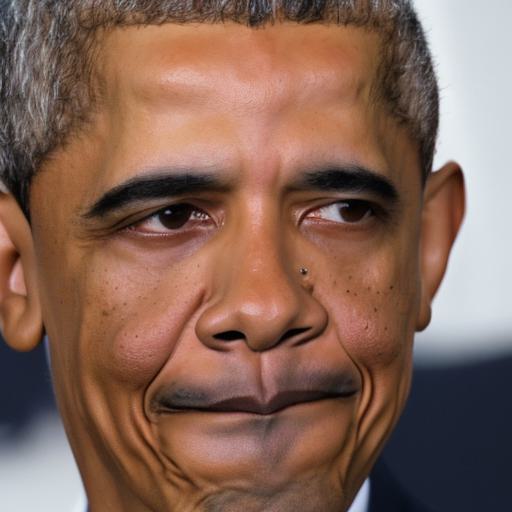} &
        \includegraphics[width=0.0714\textwidth]{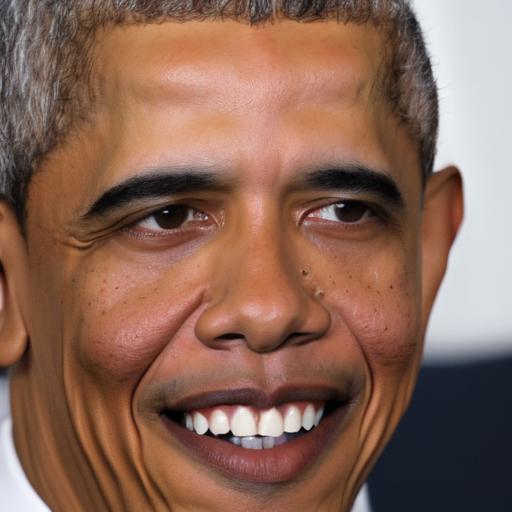} &
        \includegraphics[width=0.0714\textwidth]{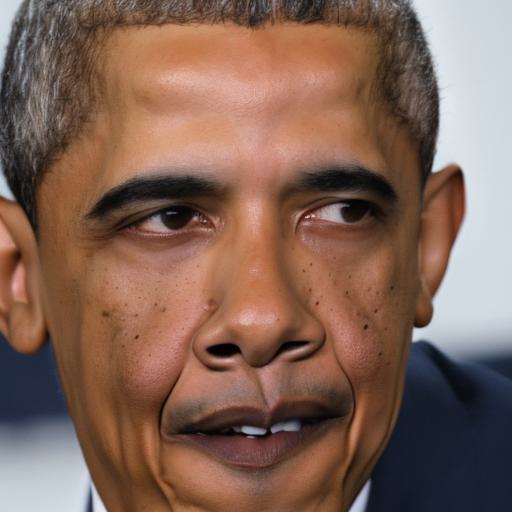} 
        \\
        \rotatebox{90}{\footnotesize{SD}} &
        \includegraphics[width=0.0714\textwidth]{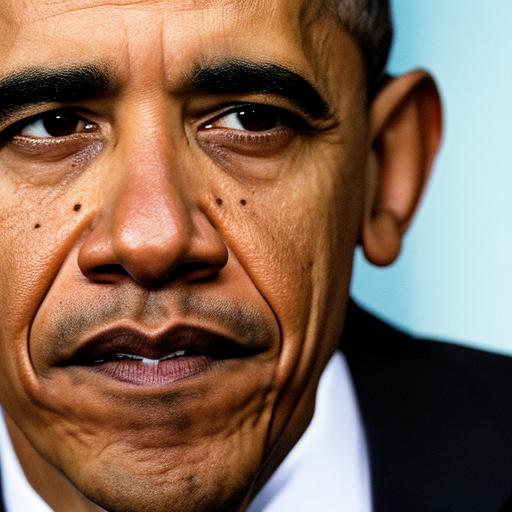} &
        \includegraphics[width=0.0714\textwidth]{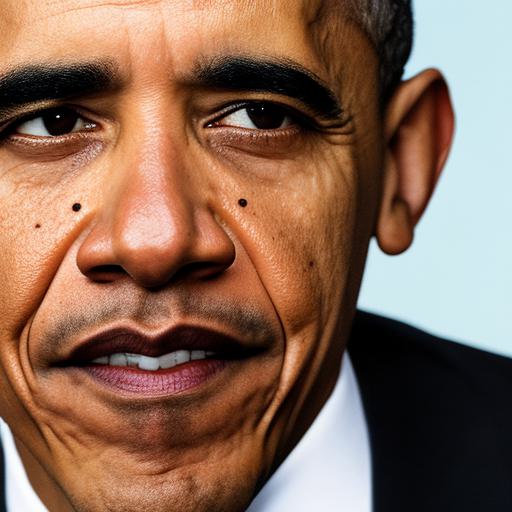} &
        \includegraphics[width=0.0714\textwidth]{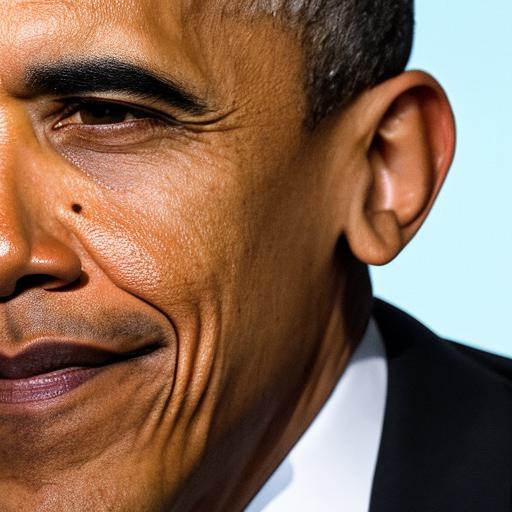} &
        \includegraphics[width=0.0714\textwidth]{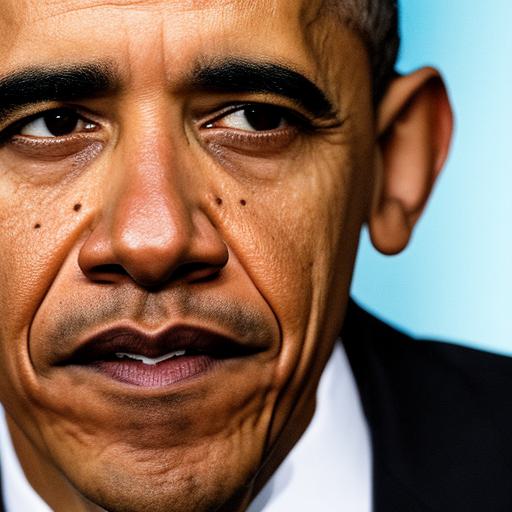} &
        \includegraphics[width=0.0714\textwidth]{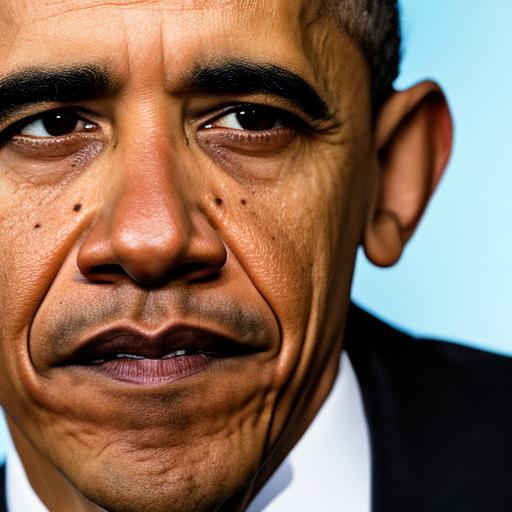} &
        \includegraphics[width=0.0714\textwidth]{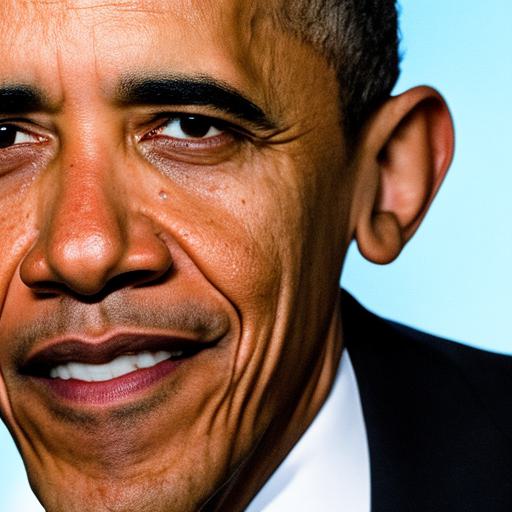} &
        \includegraphics[width=0.0714\textwidth]{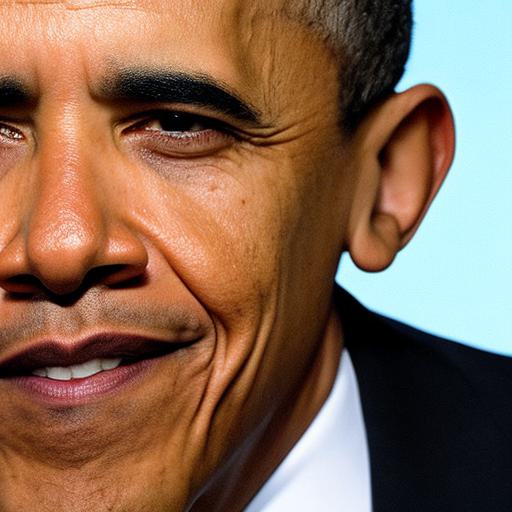} &
        \includegraphics[width=0.0714\textwidth]{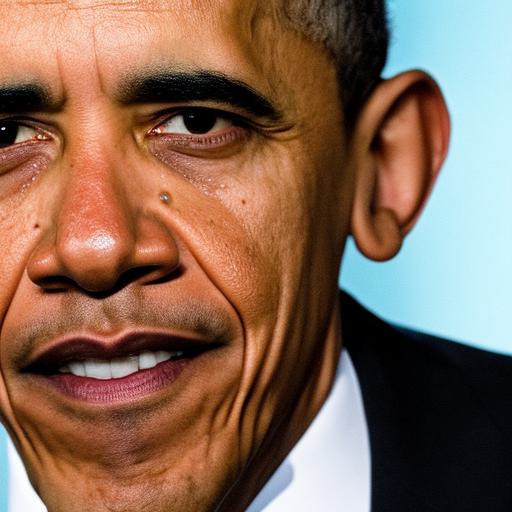} &
        \includegraphics[width=0.0714\textwidth]{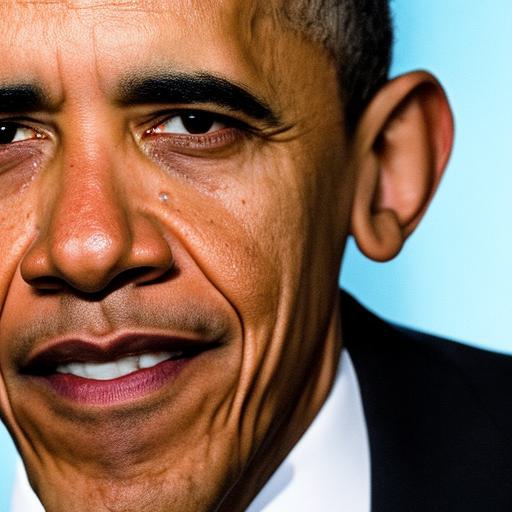} &
        \includegraphics[width=0.0714\textwidth]{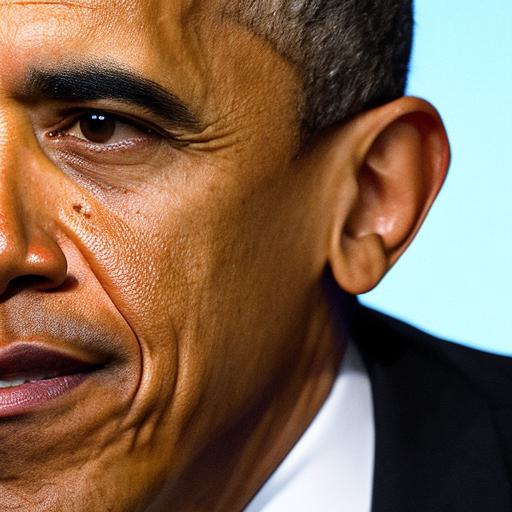} &
        \includegraphics[width=0.0714\textwidth]{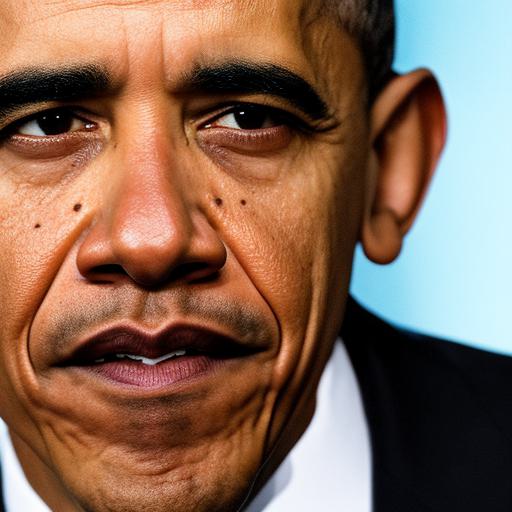} &
        \includegraphics[width=0.0714\textwidth]{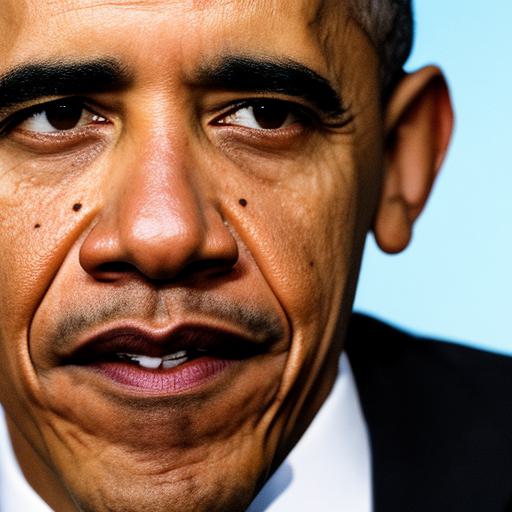} &
        \includegraphics[width=0.0714\textwidth]{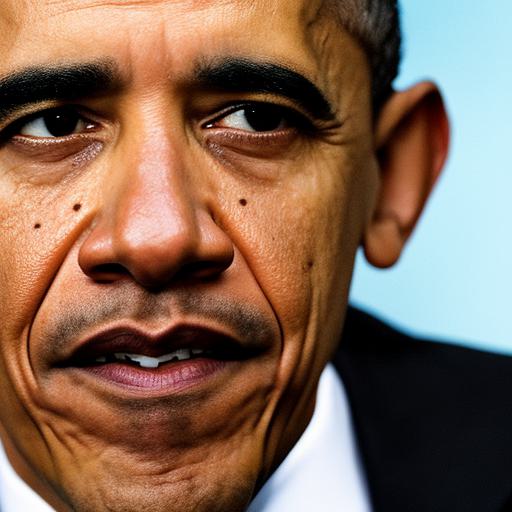} 
        \\
         & \footnotesize{Neutral} & \footnotesize{\qcr{AU1}} & \footnotesize{\qcr{AU2}} & \footnotesize{\qcr{AU4}} & \footnotesize{\qcr{AU5}} & \footnotesize{\qcr{AU6}} & \footnotesize{\qcr{AU9}} & \footnotesize{\qcr{AU12}} & \footnotesize{\qcr{AU15}} & \footnotesize{\qcr{AU17}} & \footnotesize{\qcr{AU20}} & \footnotesize{\qcr{AU25}} & \footnotesize{\qcr{AU26}} \\
        \vspace{-1cm}
        \end{tabular}
    \end{subfigure}%
\end{center}
\caption{
Comparison of different methods on 12 individual AUs with the prompt \textit{A close-up of Barack Obama}. See \cref{fig:aus} for the textual descriptions of AUs. 
}
\label{fig:au_comparison}
\vspace{-3mm}
\end{figure*}

%The learned embeddings of AUs should transition smoothly across varying intensity levels, while ensuring that combinations of individual AU values interact cohesively and produce accurate facial expressions.

%Assuming a moderate discretation step of 0.25 of the range $[0, 5]$ with vector length 12, there are over $7e^{15}$ possible values. Clearly there are not enough training samples to cover the whole distribution, even with discretation. Without access to a pre-trained model capable of mapping AUs to features ...
%The learned embeddings should be smooth across the different intensity values. 2. The embeddings need to be able to locally do individual AU edits. 3. Combining multiple AUs should face no problems and the intensities should be fully controllable even with multiple different AUs active.

\paragraph{AU Encoder}
\vspace{-2mm}
is responsible for transforming the raw AU vector into an embedding that can be passed to the adapter module. A simple one-layer (or multi-layer) MLP proves inadequate, as it tends to overfit to unique AU samples and fails to generalize to unseen cases. Experimental results show that using a raw AU vector without any encoding can effectively learn continuity, but it struggles with handling combinations of AUs. To address this, we combine both approaches by employing an MLP with a residual connection that incorporates the raw AU signal. This hybrid method leverages the strengths of both techniques, ensuring better generalization and effective learning of AU combinations.

% \begin{equation}
%     z_{au} = \mathbf{y} + \mlp(\mathbf{y})
% \end{equation}

%We experiment with several options for the AU encoder such as different MLPs and residual connections. See the experimental section for the best one.
%Based on empirical experiments we find that a simple MLP is not sufficient. It is able to represent local and global context but struggles with continuous intensity. By adding a residual connection directly from the AU vector enables the continuity.

\paragraph{Distribution Smoothing}
\label{sec:distribution_smoothing}
Continuous conditional GAN \cite{ccgan} employs vicinal risk estimation instead of the typical empirical risk estimation to improve performance on continuous labels. The underlying principle is that for a conditional distribution $p(x | y)$, a small perturbation to the label $y'$ results in a negligible change to the conditional distribution $p(x | y')$. This approach is particularly relevant in practical scenarios where AUs are annotated by humans and are likely to contain imperfections, particularly since the labels must be provided as integer values. We follow the suggestion by \cite{ccgan} to add a small perturbation to the labels during training.
%\begin{equation}
%    y' = y + \epsilon, \epsilon \sim \mathcal{N}(\mu,\,\sigma^{2}).
%\end{equation}

\subsection{Dataset Construction}
\label{sec:dataset}
Due to the limitations of existing datasets with AU labels, further processing is required. DISFA \cite{disfa} contains ground truth AU frame-by-frame annotations for twelve different AUs with intensity range of only 27 subjects in a laboratory environment with a facial resolution of around $250 \times 250$. Directly training on this partition not only lacks text prompts, but also leads to overfitting to the laboratory background and to the different individuals due to the sparsity of AU labels. To increase the number of subjects and high-resolution samples, AffectNet \cite{affectnet} is utilized. The dataset is first filtered to remove non-photorealistic facial images by using BLIP-2 \cite{blip-2} and low-quality images. After the filtering, LibreFace \cite{libreface} is used to automatically annotate AUs for each image. To ensure T2I model compatibility, images from AffectNet and DISFA are captioned with BLIP-2 \cite{blip-2}. DISFA \cite{disfa} provides 90,000 samples with accurate manual labels, compensating for the inaccuracy of automatic annotations, while AffectNet \cite{affectnet} contributes another 90,000 high-resolution samples featuring diverse backgrounds and a wide range of subjects. Further details can be found in the supplementary material.

\section{Experiments}

\subsection{Experiment details}
In all our experiments, we use the Stable Diffusion 2-1-base \footnote{\url{https://huggingface.co/stabilityai/stable-diffusion-2-1-base}} as the base diffusion model, operating at a $512 \times 512$ resolution. For methods with LoRA \cite{lora} we use rank 32. Guidance \cite{cfg} is applied by setting all AUs to 0s. For full details of the experiments see the supplementary.

\paragraph{Testing details}
\vspace{-2mm}
For the quantitative analysis, we generate two different sets of results. 1) Individual AUs with varying intensity and 2) Combination AUs. For both of the groups we use 15 different prompts, both handcrafted and sampled from the training dataset, to ensure in- and out-of-domain performance. For individual AUs there are 12 AUs, each with 5 different intensities. For the combinations 50 different sets are used for each prompt. This brings the total number of generated samples to $15 \times 12 \times 5 + 15 \times 50 = 1650$.

\paragraph{Metrics}
\vspace{-2mm}
The proposed metrics should consider two factors. 1) The correctness of the generated samples in accordance to the AU prompt and 2) the ability to retain the prompt and character consistency. The first is measured with an AU classifier from \cite{libreface}. By measuring difference
\begin{equation}
    \text{AU}_{\text{MSE}} = \frac{1}{n_{\text{AU}}}\sum_{au=0}^{n_{\text{AU}}} ||Y_{\text{au}} - \phi(\epsilon_{\theta}(p, Y_{\text{au}}))||_2
\end{equation}
between the prompted ground truth AU and the predicted result $\phi(\cdot)$ from the generated sample $\epsilon_\theta(\cdot)$ given the condition $Y_{\text{au}}$ and prompt $p$.

%We however note again as in \cref{sec:dataset} that the classifier's performance is very reliable. Nonetheless it gives indication to the models' performance.

To measure the prompt adherence and character consistency, we compute the CLIP similarity 
\begin{equation}
    \text{CLIP-I} = \text{CLIP-sim}(\epsilon_\theta(Y_0), \epsilon_\theta(Y_{\text{au}}))
\end{equation}
between a sample generated with no AU condition and a sample generated with the AU condition. In reality, the similarity should be moderate. Perfect similarity would mean no changes due to the AU condition, while high dissimilarity would break character consistency.

\begin{figure*}
\begin{center}
    \setlength{\tabcolsep}{1pt}
    \begin{subfigure}{0.95\textwidth}
    \hspace{-6mm}
    \begin{tabular}{*{5}{c}@{\hskip 8pt} *{5}{c}}
        \multicolumn{5}{c}{\qcr{AU6+12+25}} & \multicolumn{5}{c}{\qcr{AU4+15+17}}  \\
        \includegraphics[width=0.1\textwidth]{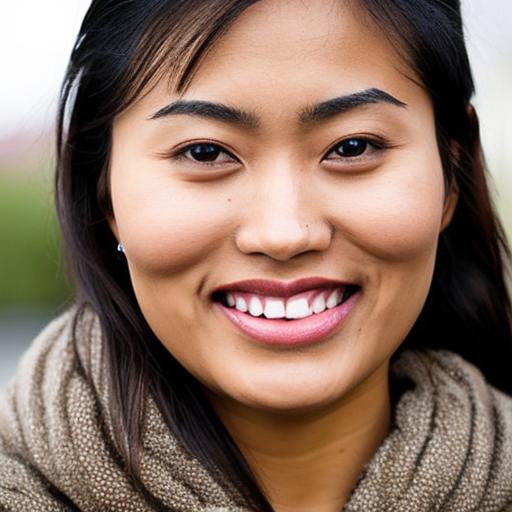} &
        \includegraphics[width=0.1\textwidth]{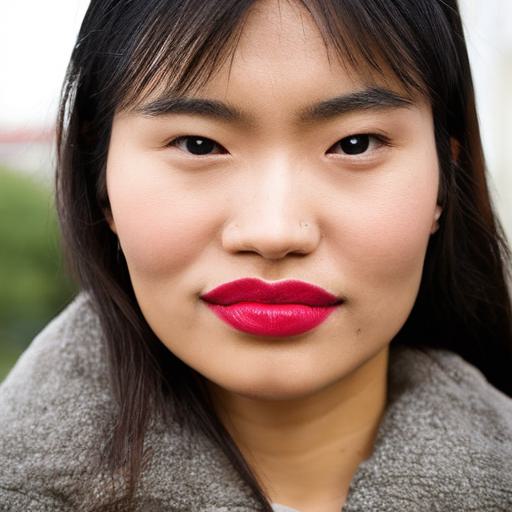} &
        \includegraphics[width=0.1\textwidth]{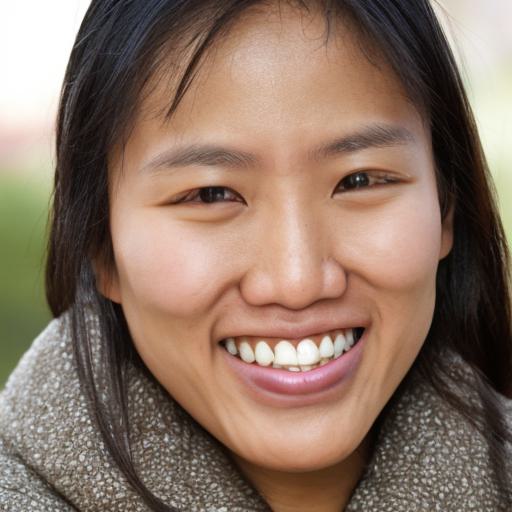} &
        \includegraphics[width=0.1\textwidth]{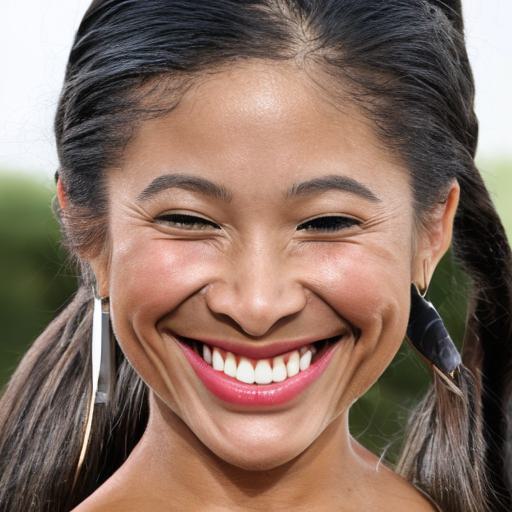} &
        \includegraphics[width=0.1\textwidth]{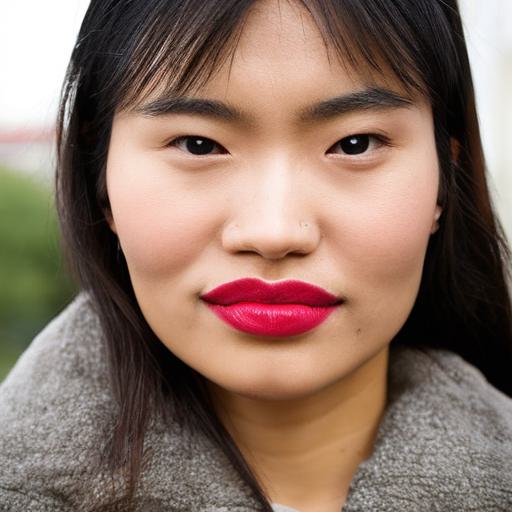} &
        \includegraphics[width=0.1\textwidth]{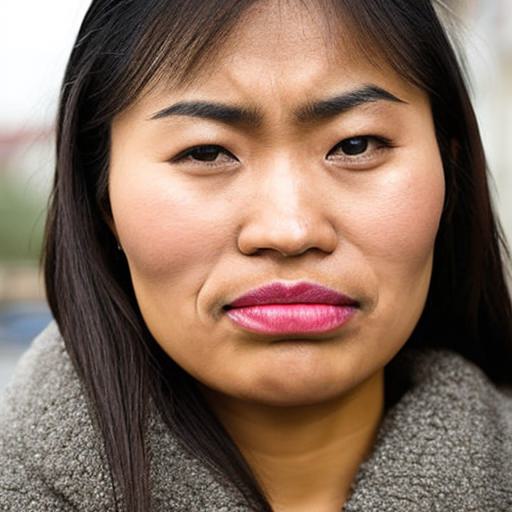} &
        \includegraphics[width=0.1\textwidth]{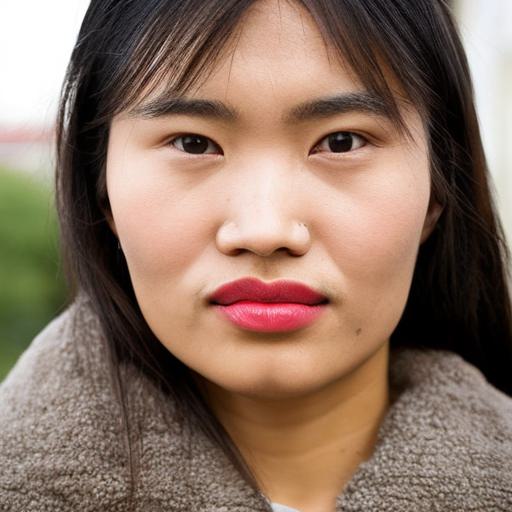} &
        \includegraphics[width=0.1\textwidth]{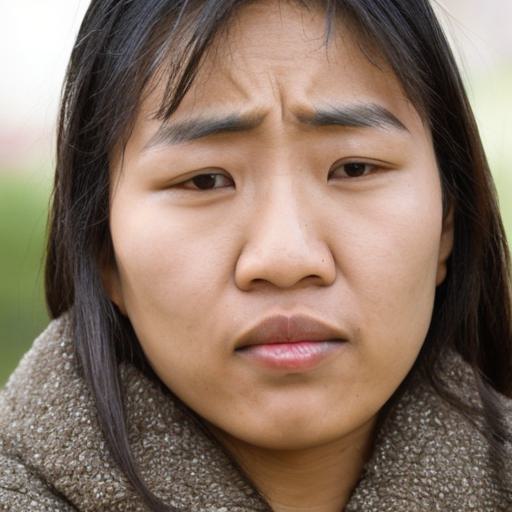} &
        \includegraphics[width=0.1\textwidth]{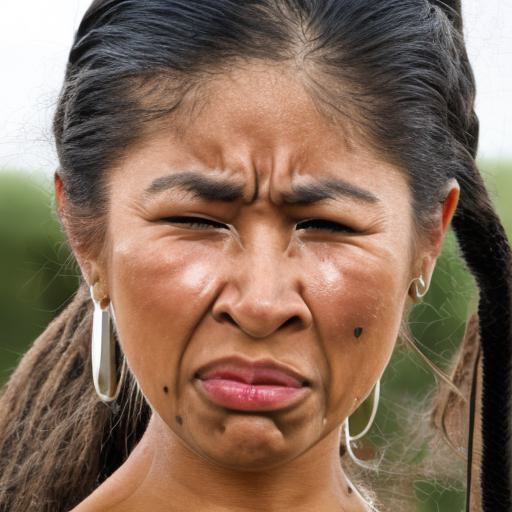} &
        \includegraphics[width=0.1\textwidth]{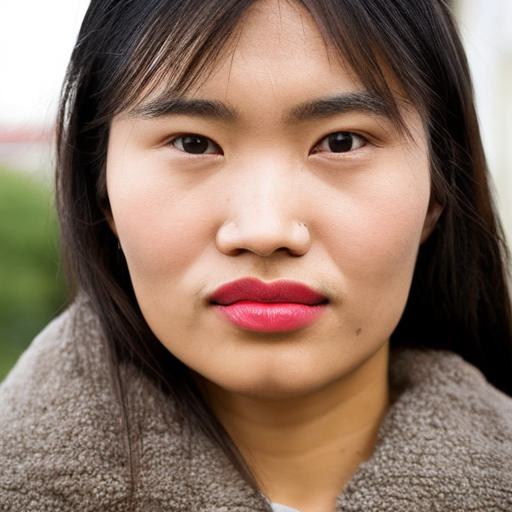}
        \\
        \multicolumn{5}{c}{\qcr{AU1+2+5+25+26}} & \multicolumn{5}{c}{\qcr{AU4+6+17+20}}  \\
        \includegraphics[width=0.1\textwidth]{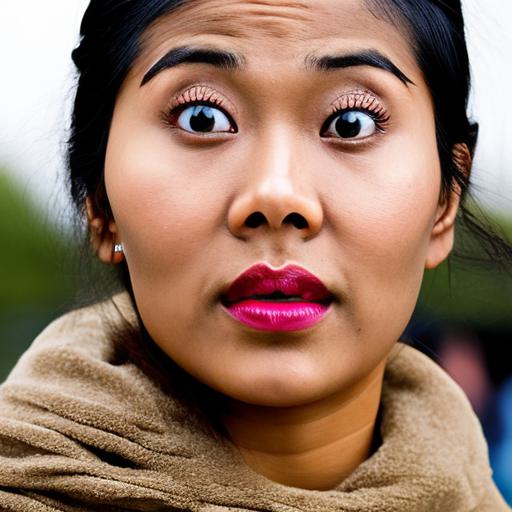} &
        \includegraphics[width=0.1\textwidth]{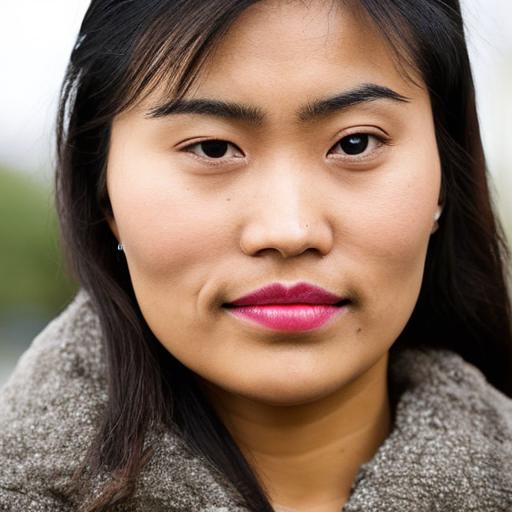} &
        \includegraphics[width=0.1\textwidth]{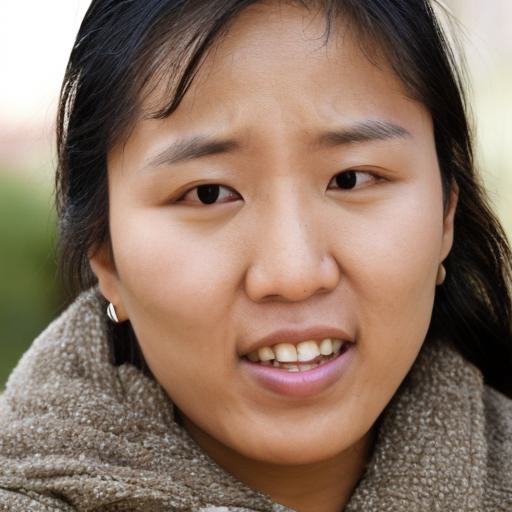} &
        \includegraphics[width=0.1\textwidth]{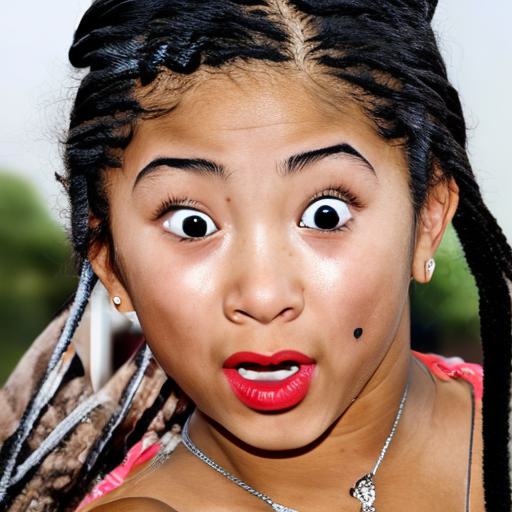} &
        \includegraphics[width=0.1\textwidth]{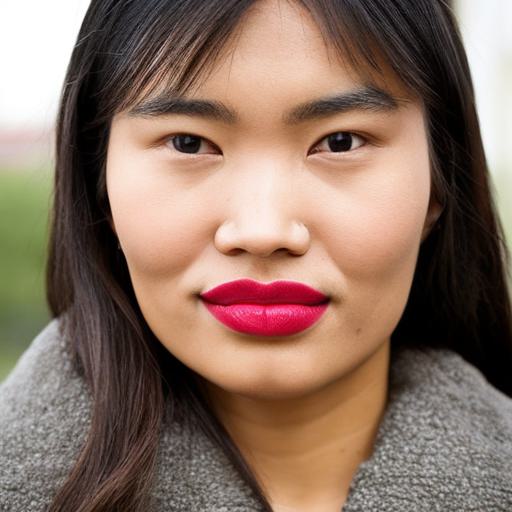} &
        \includegraphics[width=0.1\textwidth]{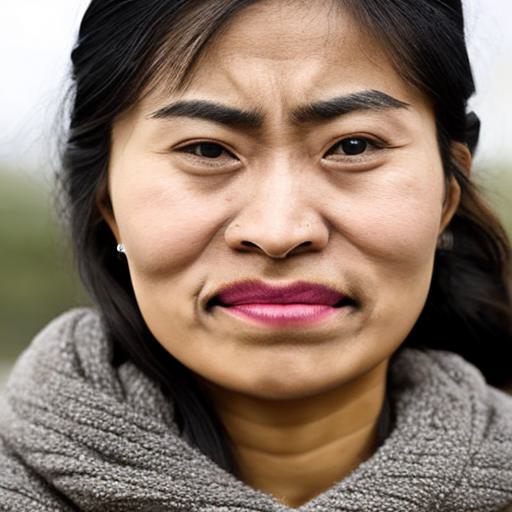} &
        \includegraphics[width=0.1\textwidth]{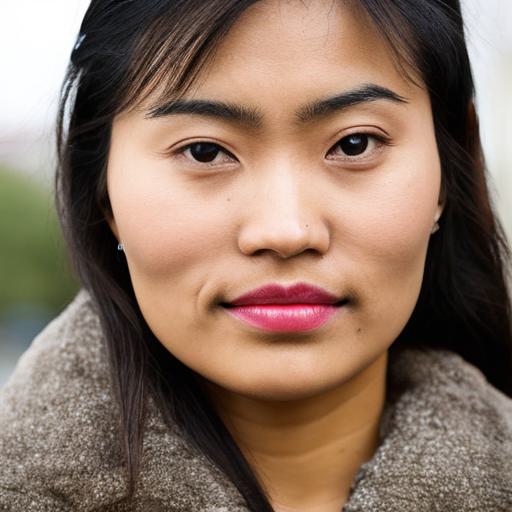} &
        \includegraphics[width=0.1\textwidth]{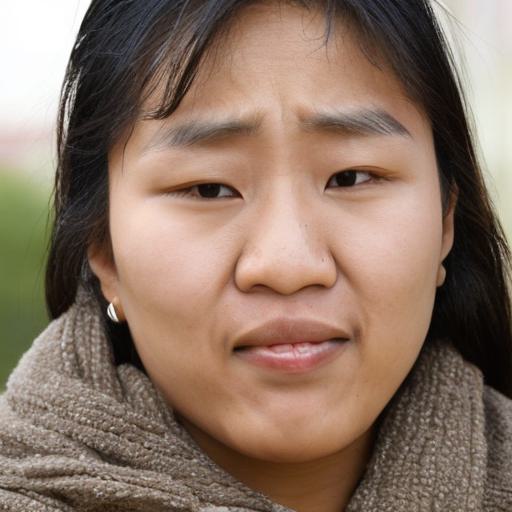} &
        \includegraphics[width=0.1\textwidth]{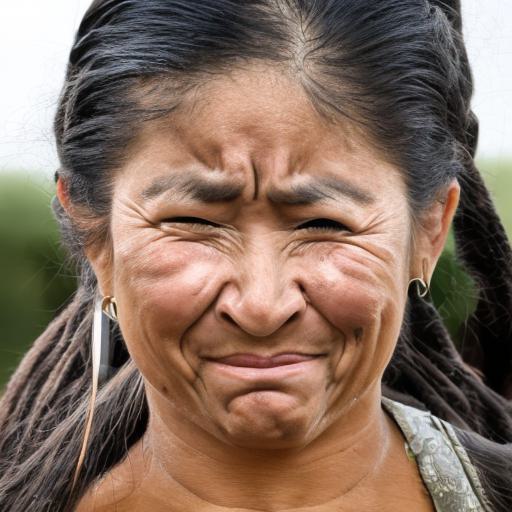} &
        \includegraphics[width=0.1\textwidth]{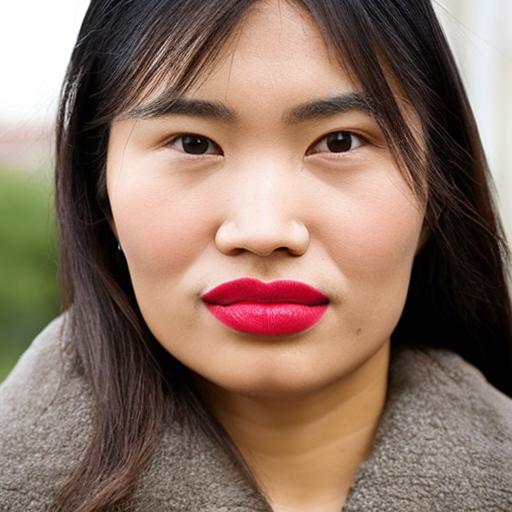}
        \\
        \multicolumn{5}{c}{\qcr{AU1+2+4+9}} & \multicolumn{5}{c}{\qcr{AU6+12}}  \\
        \includegraphics[width=0.1\textwidth]{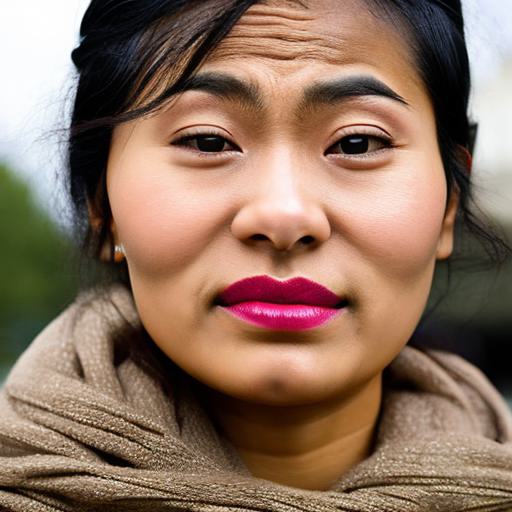} &
        \includegraphics[width=0.1\textwidth]{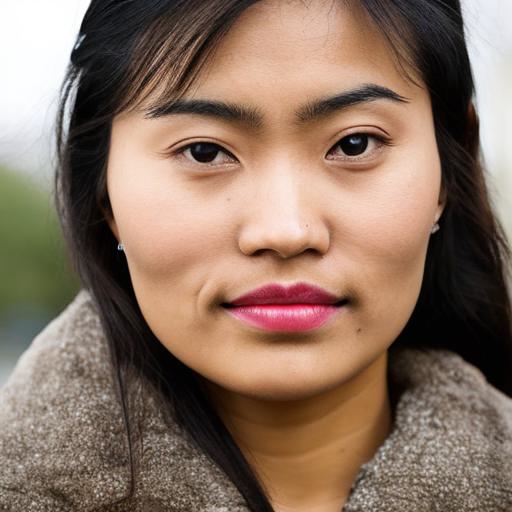} &
        \includegraphics[width=0.1\textwidth]{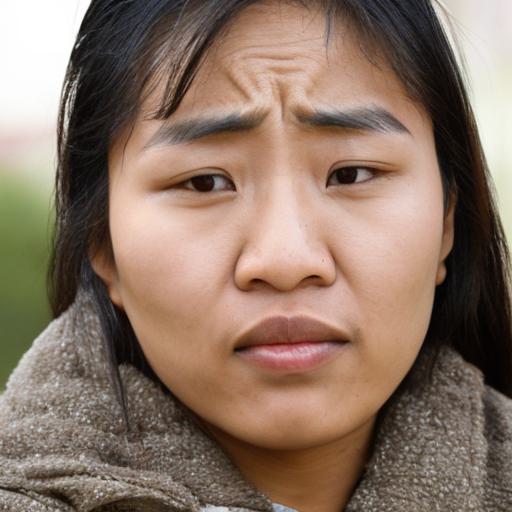} &
        \includegraphics[width=0.1\textwidth]{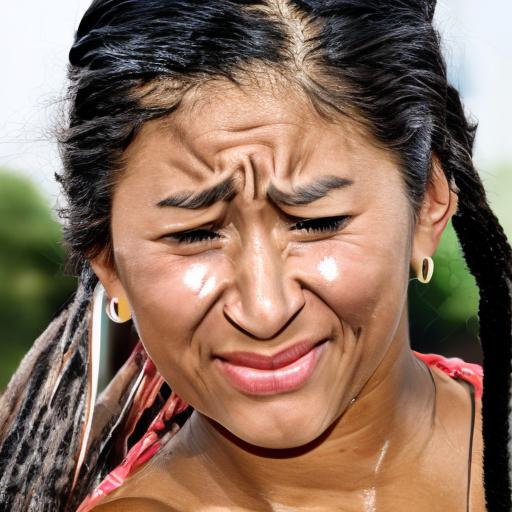} &
        \includegraphics[width=0.1\textwidth]{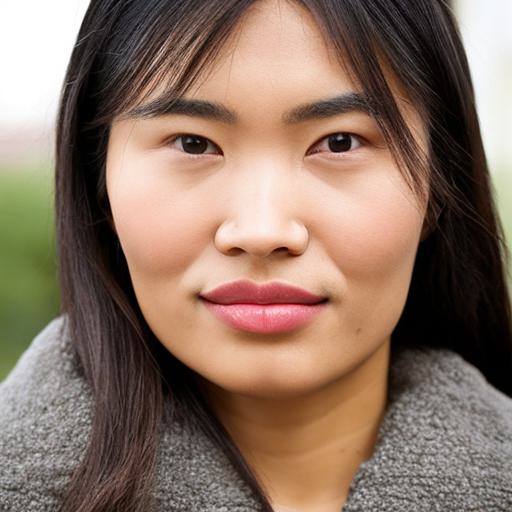} &
        \includegraphics[width=0.1\textwidth]{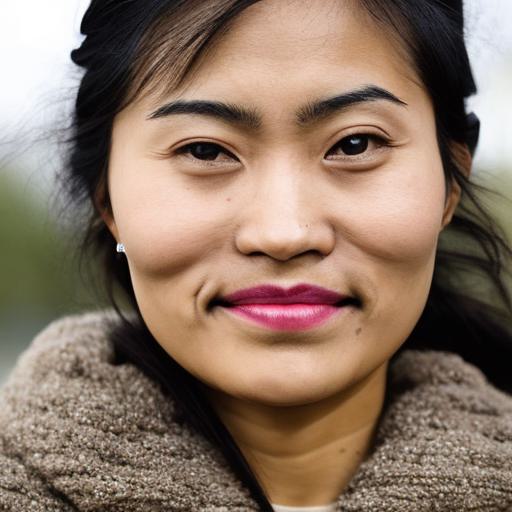} &
        \includegraphics[width=0.1\textwidth]{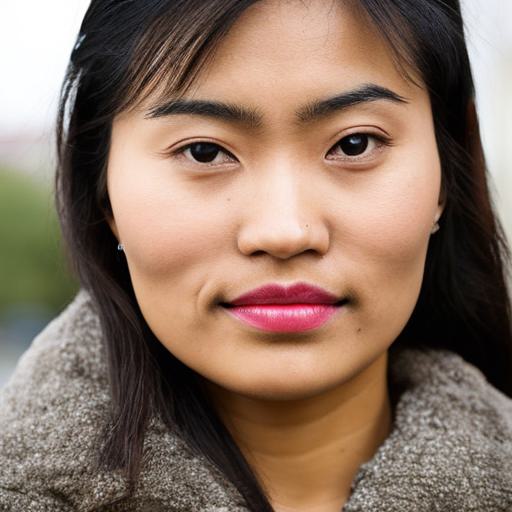} &
        \includegraphics[width=0.1\textwidth]{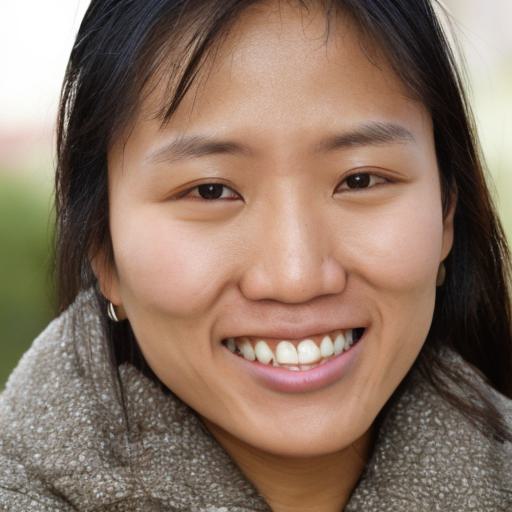} &
        \includegraphics[width=0.1\textwidth]{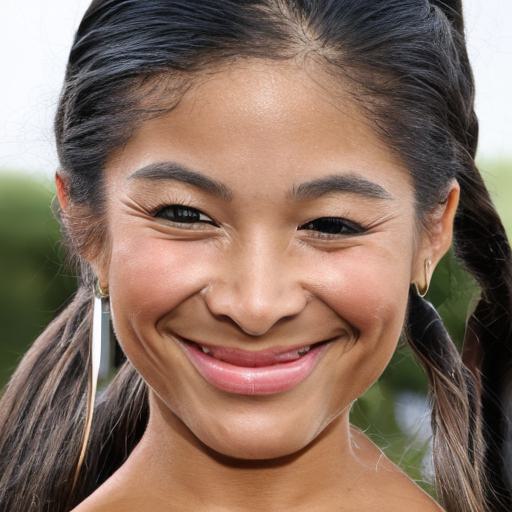} &
        \includegraphics[width=0.1\textwidth]{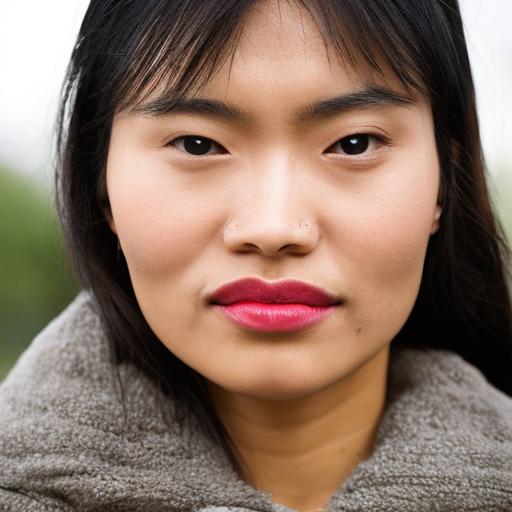}
        \\
        
         \footnotesize{FineFace (Ours)} & \footnotesize{LoRA-AU} & \footnotesize{LoRA-T} & \footnotesize{DB} & \footnotesize{SD} & \footnotesize{FineFace (Ours)} & \footnotesize{LoRA-AU} & \footnotesize{LoRA-T} & \footnotesize{DB} & \footnotesize{SD} \\
        \vspace{-1cm}
        \end{tabular}
    \end{subfigure}%
\end{center}
\caption{
Comparison of methods on combination AUs with the prompt \textit{An Asian woman in the park}.
}
\label{fig:au_combinations}
\vspace{-3mm}
\end{figure*}

%The quantitative metrics are considered from a few points. To measure the performance of AU condition, the MSE is computed between the AU prompt condition and the predicted AUs from a AU intensity estimator \cite{libreface}. To ensure the prompt agreement, CLIP-T is employed, which computes the average cosine similarity between the text prompt and image CLIP embeddings. Consistent character is an important factor in content creation. By measuring the CLIP similarity between the neutral and AU conditioned image embeddings, the dissimilarity of consistent character can be measured. We note this as CLIP-I.

\paragraph{Baseline methods}
\vspace{-2mm}
Since there are no previous works on generating facial images with AU conditions, we establish several baselines. The simplest baseline is Stable Diffusion (\textit{SD}) without any fine-tuning, where the AU condition is injected by first transforming the AU vector into text form and then adding it to the text prompt. For the DreamBooth (\textit{DB}) baseline \cite{dreambooth}, SD is fine-tuned using a prior-preservation loss \cite{dreambooth}. Next, instead of fine-tuning the entire SD, only the added LoRA layers \cite{lora} are trained, referred to as \textit{LoRA-T}. Finally, instead of injecting AU information with text prompts, the AU vector is projected to the clip space with a learnable AU encoder, similar to \cite{unifiedface}, referred to as \textit{LoRA-AU}.

\subsection{Results}
We compare the proposed method against the baselines noted above both qualitatively and quantitatively. For better understanding the qualitative results, we recommend the readers to see \cref{fig:aus}.

\subsubsection{Qualitative results}
%Qualitative results can be seen from \cref{fig:teaser,fig:au_comparison,fig:au_combinations,fig:au_intensity}. 
\paragraph{Individual AUs}
In \cref{fig:au_comparison} we show a comparison with 12 AUs conditioned individually. As we can see plain SD is unable to follow the conditions accurately, with only minor changes to facial expressions with AU12 that seems reasonable. DB is able to follow the AU conditions in a fair manner, but fails with most of the upper face AUs (1, 2, 4, 5). Furthermore DB overfits to the training data and fails to be consistent with the prompt, compared to the original SD. LoRA-AU is unable to change the results and retains the same facial expression, as the method focuses on following the text prompt. The proposed method is able to retain the original prompt, while still changing the facial expressions for most of the AUs. The method struggles to change the facial expression for AUs such as 6 and 9, as these AUs are rarely seen individually.

\paragraph{Combinations}
\vspace{-2mm}
\Cref{fig:au_combinations} showcases the results from combining multiple AUs together to the condition. We can again observe that SD and LoRA-AU are unable to follow the AU conditions accurately and only make minor changes. Although DB can perform most AU combinations reasonably well, it tends to exaggerate the results, which is expected as DB fine-tunes the entire UNet. LoRA-T is able to follow the AU conditions well for most cases, but struggles with the more complex \qcr{AU1+2+5+25+26}. Furthermore, the prompt is not followed as well as the character is changed compared to the original from SD. The proposed method accurately follows the AU conditions, producing natural-looking results. However, minor deviations from the character can occasionally be observed.

\begin{figure}
\begin{center}
    \setlength{\tabcolsep}{1pt}
    \begin{subfigure}{0.81\textwidth}
    \hspace{-3mm}
    \begin{tabular}{*{7}{c}}
        \multicolumn{7}{c}{\qcr{AU4}}  \\
        \rotatebox{90}{\footnotesize{\makecell{FineFace \\(Ours)}}} &
        \includegraphics[width=0.0833\textwidth]{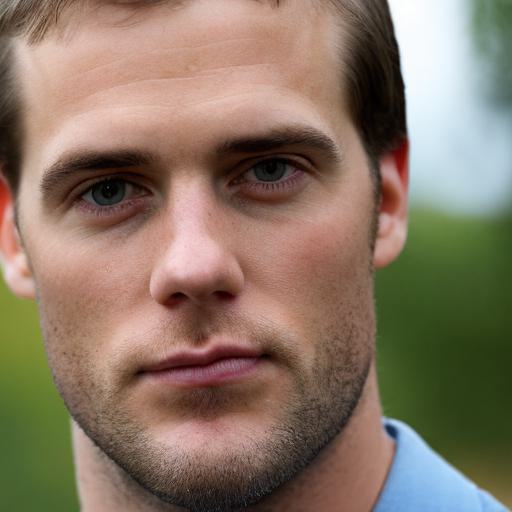} &
        \includegraphics[width=0.0833\textwidth]{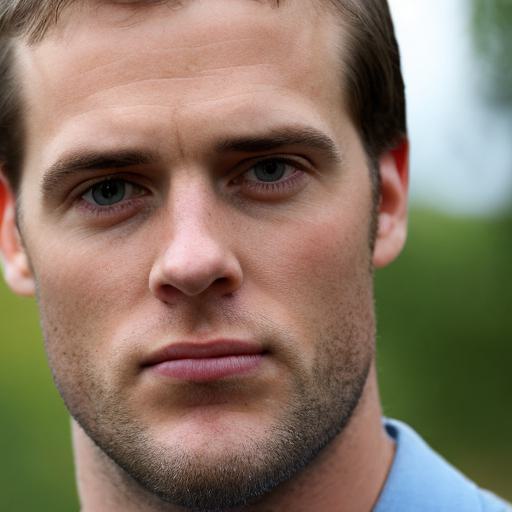} &
        \includegraphics[width=0.0833\textwidth]{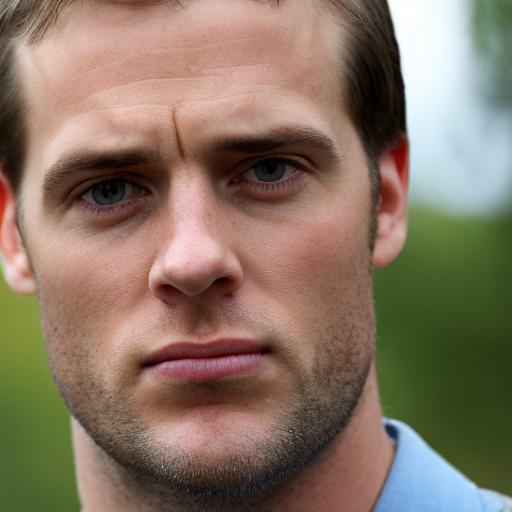} &
        \includegraphics[width=0.0833\textwidth]{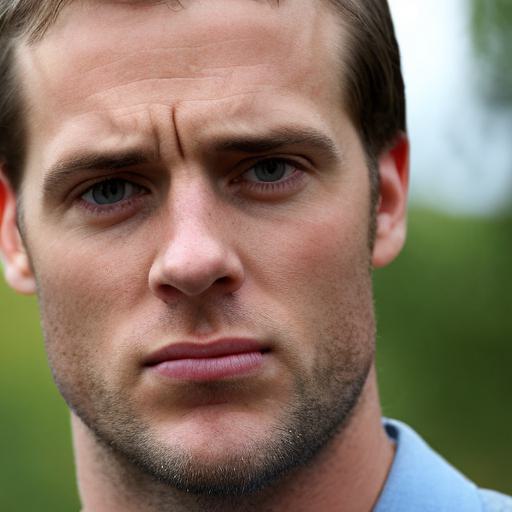} &
        \includegraphics[width=0.0833\textwidth]{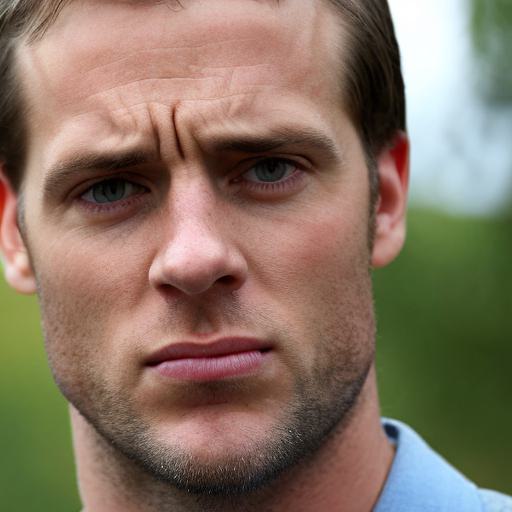} &
        \includegraphics[width=0.0833\textwidth]{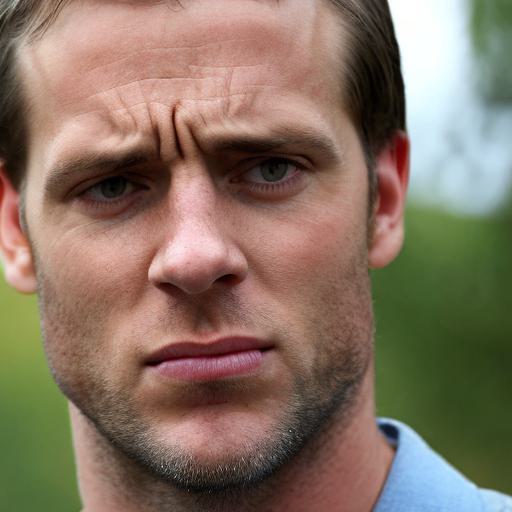}
        \\
        \rotatebox{90}{\footnotesize{LoRA-T}} &
        \includegraphics[width=0.0833\textwidth]{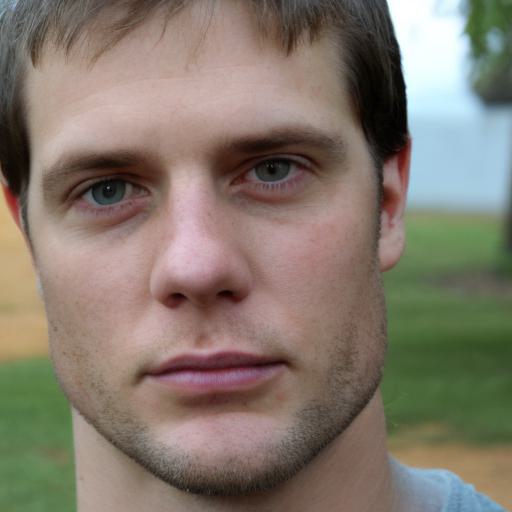} &
        \includegraphics[width=0.0833\textwidth]{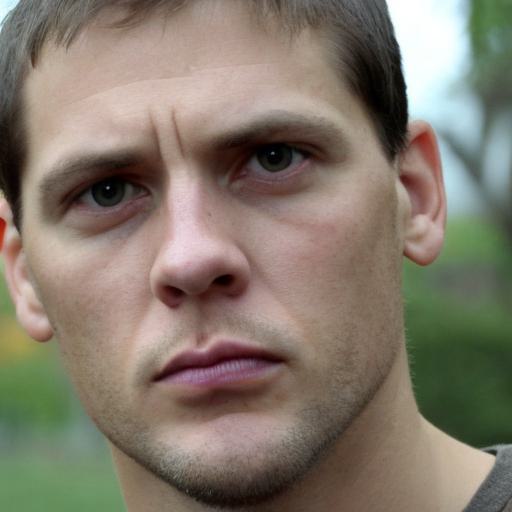} &
        \includegraphics[width=0.0833\textwidth]{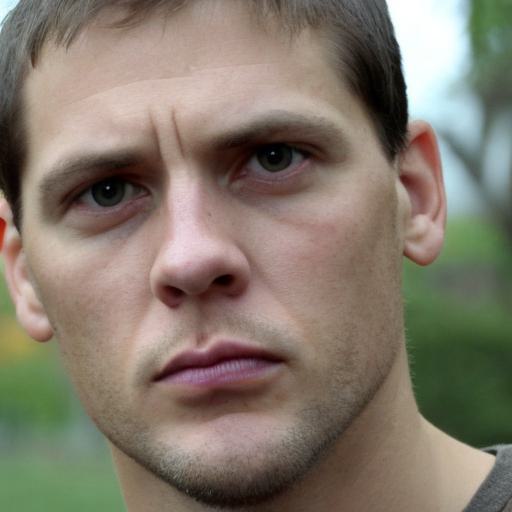} &
        \includegraphics[width=0.0833\textwidth]{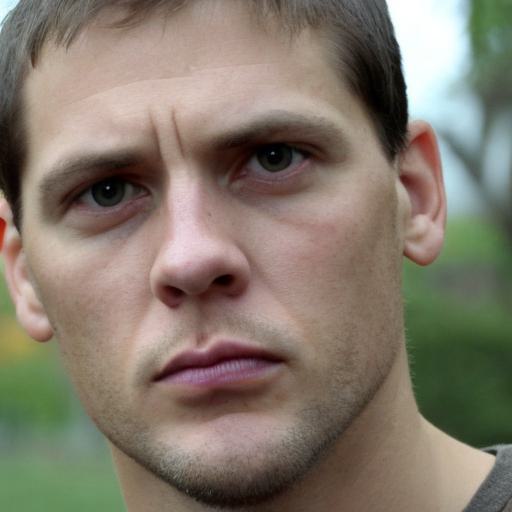} &
        \includegraphics[width=0.0833\textwidth]{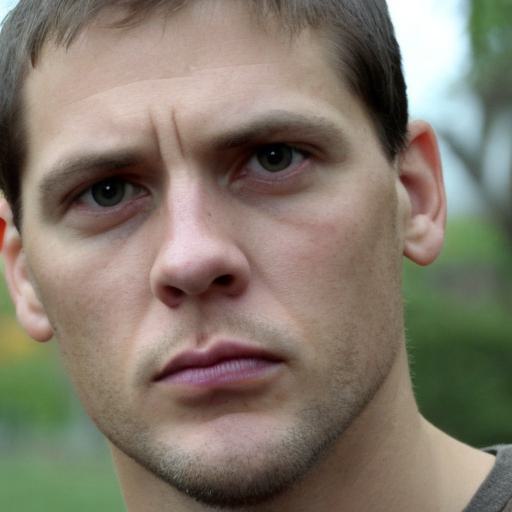} &
        \includegraphics[width=0.0833\textwidth]{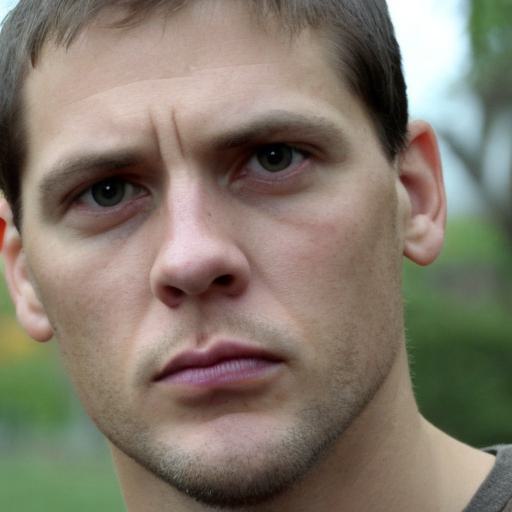}
        \\
        \multicolumn{7}{c}{\qcr{AU12}}  \\
        \rotatebox{90}{\footnotesize{\makecell{FineFace \\(Ours)}}} &
        \includegraphics[width=0.0833\textwidth]{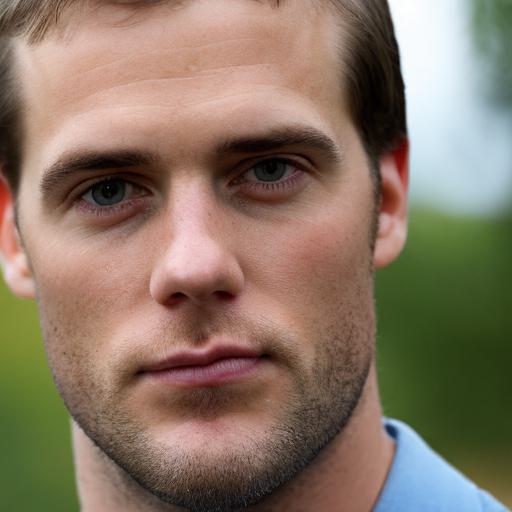} &
        \includegraphics[width=0.0833\textwidth]{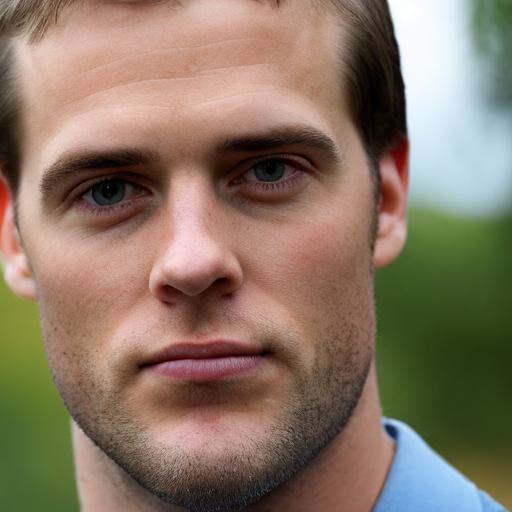} &
        \includegraphics[width=0.0833\textwidth]{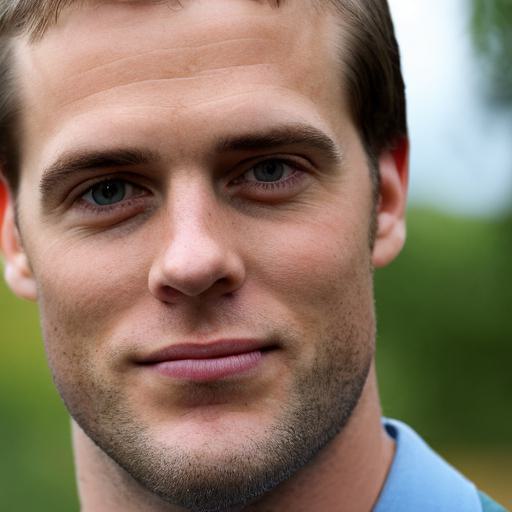} &
        \includegraphics[width=0.0833\textwidth]{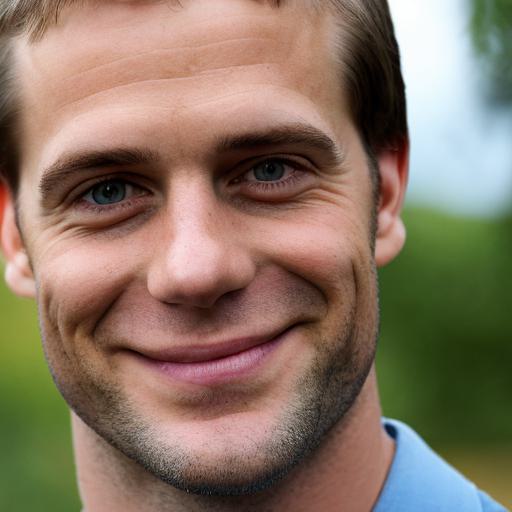} &
        \includegraphics[width=0.0833\textwidth]{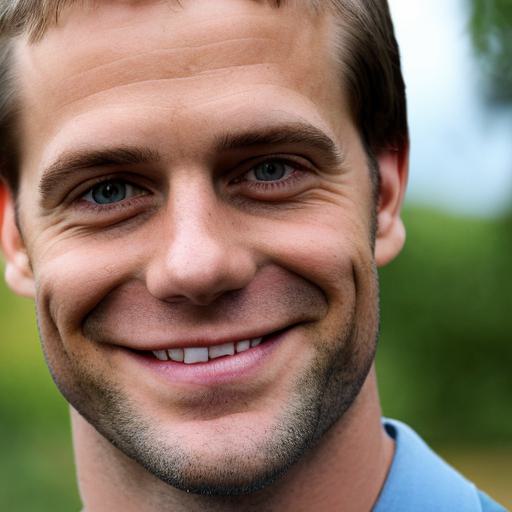} &
        \includegraphics[width=0.0833\textwidth]{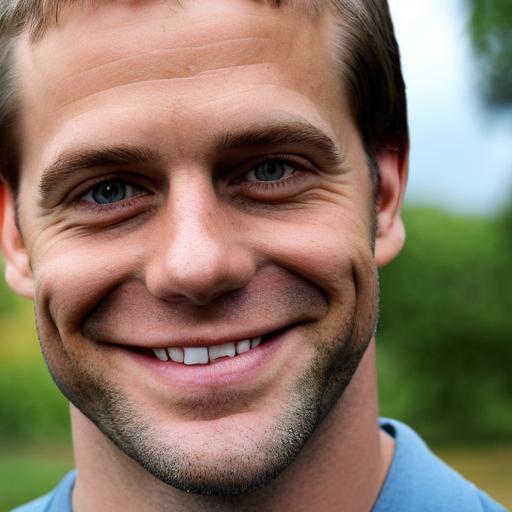}
        \\
        \rotatebox{90}{\footnotesize{LoRA-T}} &
        \includegraphics[width=0.0833\textwidth]{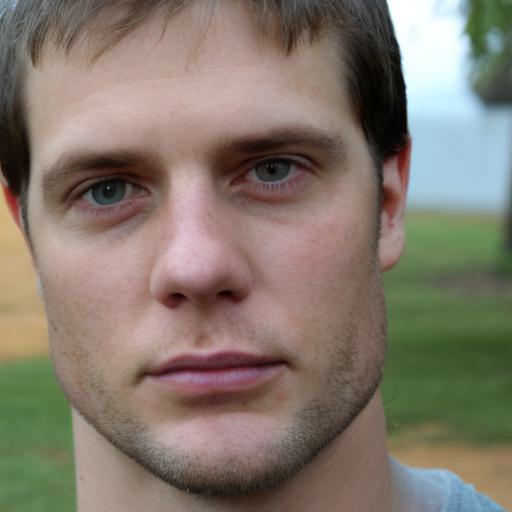} &
        \includegraphics[width=0.0833\textwidth]{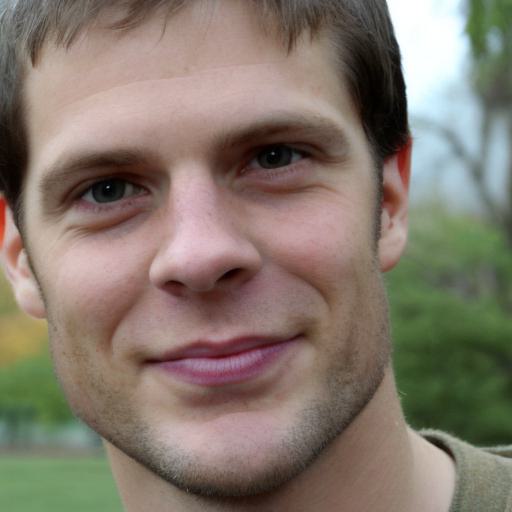} &
        \includegraphics[width=0.0833\textwidth]{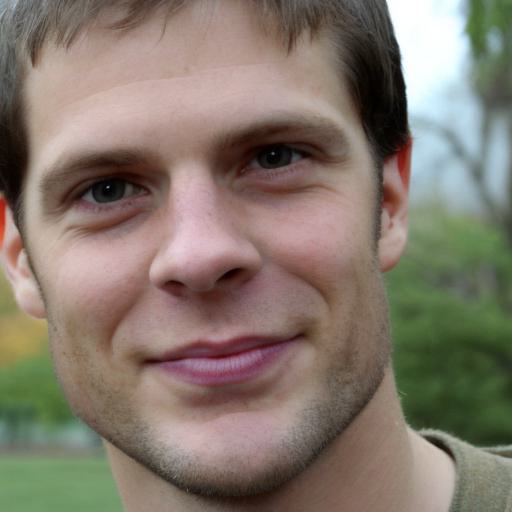} &
        \includegraphics[width=0.0833\textwidth]{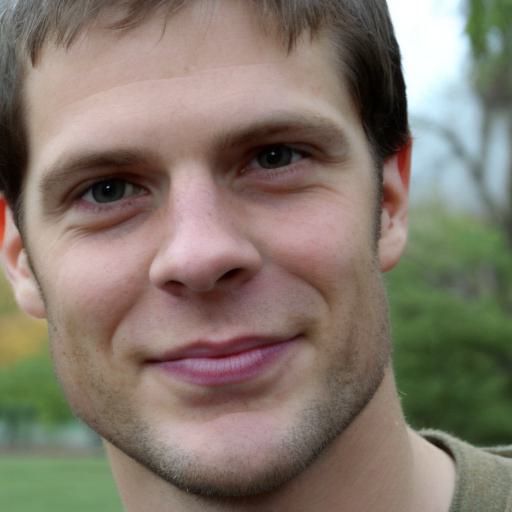} &
        \includegraphics[width=0.0833\textwidth]{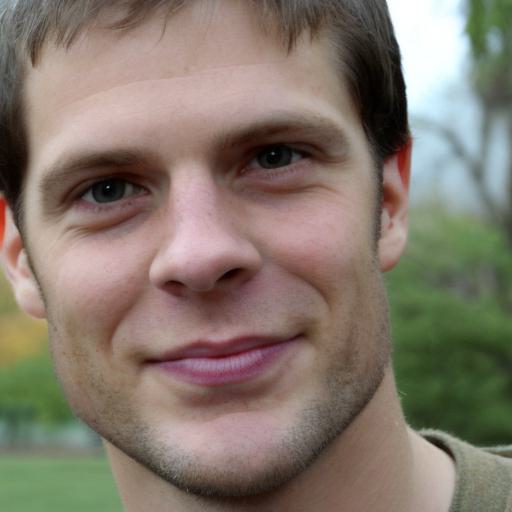} &
        \includegraphics[width=0.0833\textwidth]{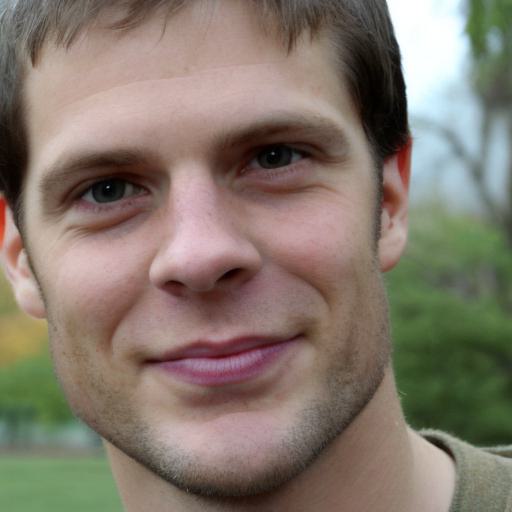}
        \\
        
         \footnotesize{Intensity} & \footnotesize{0} & \footnotesize{1} & \footnotesize{2} & \footnotesize{3} & \footnotesize{4} & \footnotesize{5} \\
        \vspace{-1cm}
        \end{tabular}
    \end{subfigure}%
\end{center}
\caption{
A comparison between LoRA-T and the proposed method of continuity of the AU intensity scale. FineFace is able to smoothly change the intensity level, while LoRA-T is stuck to a single intensity level due to the text input.
}
\label{fig:au_intensity}
\vspace{-3mm}
\end{figure}

\paragraph{Intensity}
\vspace{-2mm}
Intensity control of AUs is crucial for creating appropriate reactions to different scenarios. The text based inputs are unable to individually control the AUs, the only way to control the strength is by changing the guidance scale \cite{cfg}, but this also affects the text prompt. \Cref{fig:au_intensity} showcases the proposed model's ability to smoothly control intensity. It should be noted that the used range is nonlinear, see \cref{fig:aus}.

%tensor([[6.9813, 7.1918, 7.4596, 6.8247],
%        [0.2312, 0.2274, 0.2175, 0.2179],
%        [0.8099, 0.8740, 0.9165, 0.9801]])

%tensor([[8.9006, 8.7896, 9.0658, 8.7482],
%        [0.2176, 0.2253, 0.2183, 0.2112],
%        [0.7156, 0.8051, 0.8603, 0.9196]])

\begin{table}[]
\centering
\vspace{-2mm}
\caption{Quantitative results with different baselines. Best results in \textbf{bold}, second best in \underline{underline}.}
\resizebox{0.48\textwidth}{!}{%
\begin{tabular}{|c|cc|cc|}
    \hline
    \multirow{2}{4em}{Method} & \multicolumn{2}{|c|}{Individual} & \multicolumn{2}{|c|}{Combination} \\
     & AU MSE $\downarrow$ & CLIP-I $\uparrow$ & AU MSE $\downarrow$ & CLIP-I $\uparrow$ \\
    \hline
    \hline
    SD & \underline{6.98} & 0.80 & 8.90 & 0.71 \\
    \hline
    DB & 7.19 & 0.87 & \underline{8.78} & 0.80 \\
    \hline
    LoRA-T & 7.45 & \underline{0.91} & 9.06 & \textbf{0.86} \\
    \hline
    LoRA-AU  & \color{gray}{6.82*} & \color{gray}{0.98$^*$} & \color{gray}{8.74*} & \color{gray}{0.91$^*$} \\
    \hline
    FineFace (Ours) & \textbf{4.71} & \textbf{0.92} & \textbf{7.54} & \underline{0.83} \\
    \hline
\end{tabular}}
\label{tab:baselines}
\scriptsize{\\$^*$ Limited change in facial expressions.}
\vspace{-7mm}
\end{table}

\subsubsection{Quantitative results}
To evaluate the effectiveness of the AU condition and retaining consistency numerical values are shown in \cref{tab:baselines} for the different baseline methods. The best performance in terms of AU MSE is achieved by the proposed method, especially in the individual AU cases. The highest prompt adherence, \ie, CLIP-I, is achieved by LoRA-AU, however as can be seen from the qualitative results, the facial expressions are hardly changed at all. Disregarding this result, the proposed method achieves the best CLIP-I of individual AUs and LoRA-T for the combinations. This result corresponds with the qualitative results: in \cref{fig:au_comparison} the pose is changing for LoRA-T, while for the proposed method it remains consistent.

\subsection{Additional Results}

\begin{figure}
\begin{center}
    \setlength{\tabcolsep}{1pt}
    \begin{subfigure}{0.40\textwidth}
    \hspace{-4mm}
    \begin{tabular}{@{\hskip -8pt}p{0.25\textwidth}@{\hskip 12pt}  p{0.25\textwidth}@{\hskip 12pt}  p{0.25\textwidth}@{\hskip 12pt}  p{0.25\textwidth}@{\hskip 12pt}}
        \includegraphics[width=0.25\textwidth]{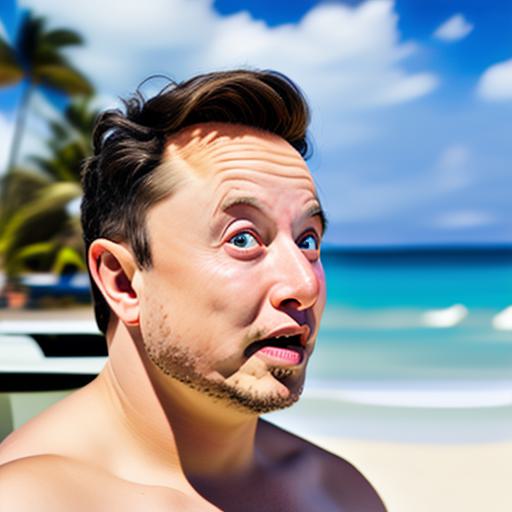} &
        \includegraphics[width=0.25\textwidth]{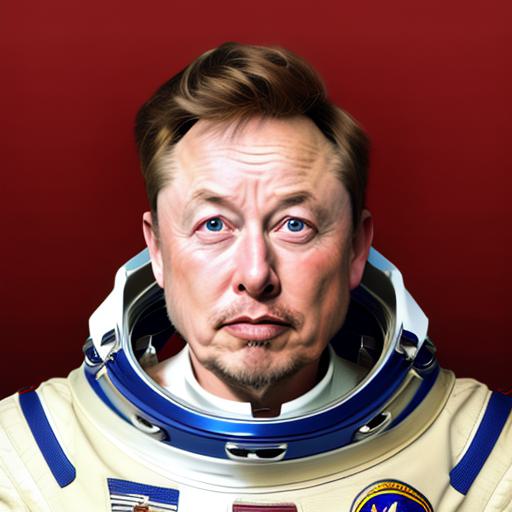} &
        \includegraphics[width=0.25\textwidth]{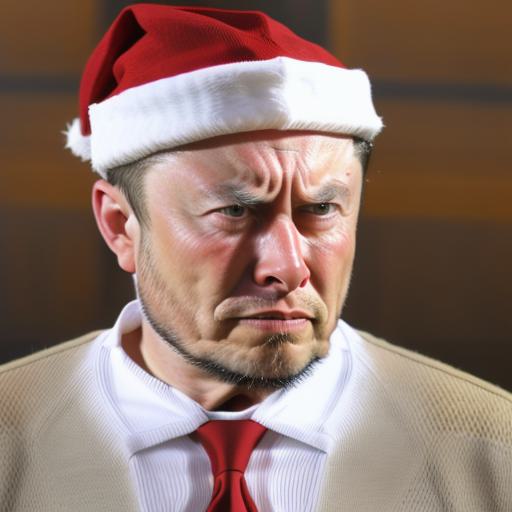} &
        \includegraphics[width=0.25\textwidth]{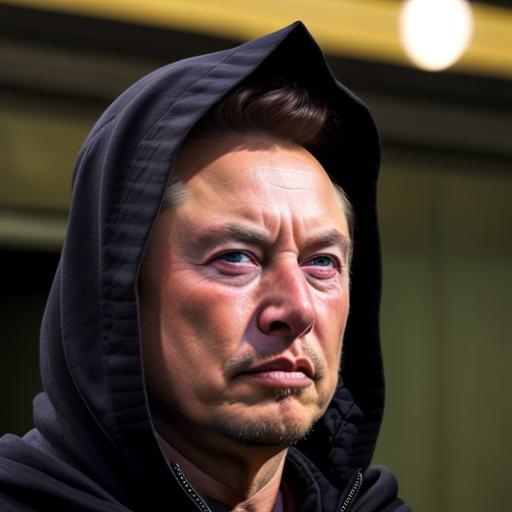}
        \\
        \footnotesize{A \{\textit{man}\} at the beach. \qcr{AU1+2+5+ 25+26}} & \footnotesize{A \{\textit{man}\} in a spacesuit \qcr{AU1+4+5}} & \footnotesize{A \{\textit{man}\} in a santa hat \qcr{AU4+17}} & \footnotesize{A \{\textit{man}\} as a dark hooded emperor \qcr{AU4+6}} \\
        \end{tabular}
    \end{subfigure}%
\end{center}
\vspace{-6mm}
\caption{
Results by combining FineFace with IP-AdapterFace. The method takes in a text prompt, AU condition and an input image. \{\textit{man}\} refers to the input image used.
}
\label{fig:ip-adapter}
\vspace{-3mm}
\end{figure}

\paragraph{Combination with Image Prompts}
Due to the architecture of the method it can be combined with IP-AdapterFace\footnote{\url{https://huggingface.co/h94/IP-Adapter-FaceID}} \cite{ipadapter}, enabling the combination of text, AUs and facial images as conditions. \Cref{fig:teaser,fig:ip-adapter} showcase results with additional image prompts. The use of three prompts—text, image, and AU—enables broader applications by providing precise and nuanced control over facial expressions as well as the identity. It should be noted that the image adapter and AU adapter are trained separately and combined during inference.%The use of three different inputs poses a challenge for the model and we can observe some degradation in both the AU conditions as well as the identity consistency.

\begin{figure}
\begin{center}
    \setlength{\tabcolsep}{1pt}
    \begin{subfigure}{0.46\textwidth}
    \hspace{-0.3cm}
    \begin{tabular}{*5c}
        \includegraphics[width=0.2\textwidth]{figures/aus/resmlp64/obama/AU1/0.jpg} &
        \includegraphics[width=0.2\textwidth]{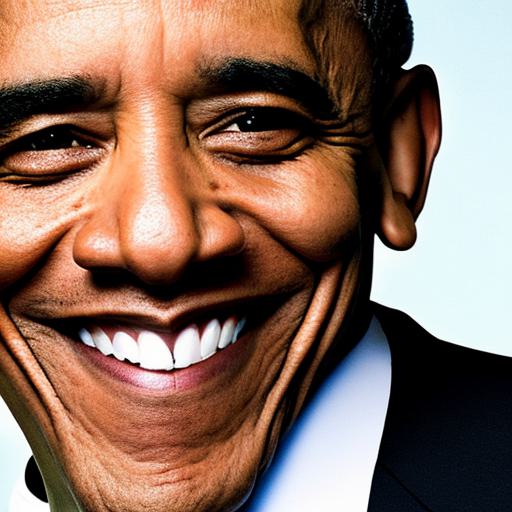} &
        \includegraphics[width=0.2\textwidth]{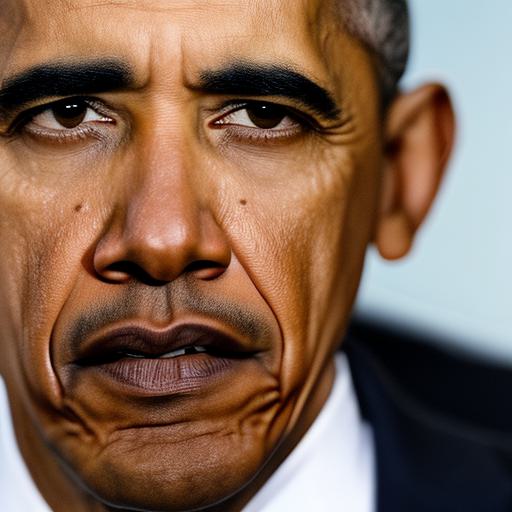} &
        \includegraphics[width=0.2\textwidth]{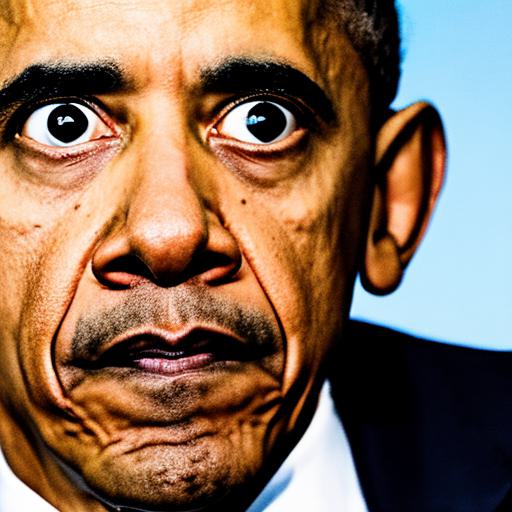} &
        \includegraphics[width=0.2\textwidth]{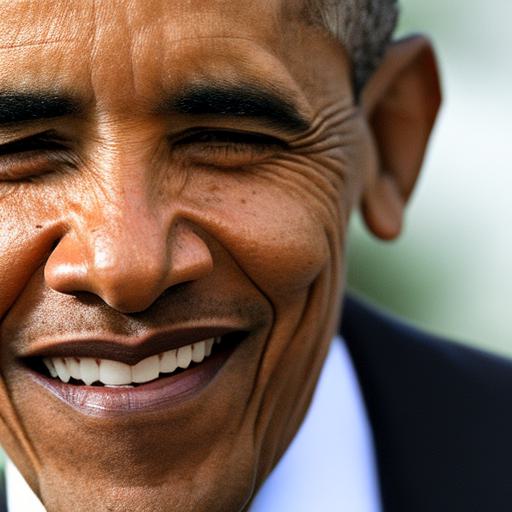}
        \\
        \footnotesize{Neutral} & \footnotesize{\qcr{AU12[10]}} & \footnotesize{\qcr{AU12[-10]}} & \footnotesize{\qcr{AU5[10]}} & \footnotesize{\qcr{AU5[-10]}} \\
        \vspace{-1cm}
        \end{tabular}
    \end{subfigure}%
\end{center}
\caption{
Going beyond the learned $[0, 5]$ scale. Negative \qcr{AU12} (Lip Corner Puller) resembles \qcr{AU15} (Lip Corner Depressor). Negative \qcr{AU5} (Upper Lid Raiser) results in expression resembling closed eyes (Upper Lid Closer). The results are directly comparable to \cref{fig:au_comparison} FineFace.
}
\label{fig:extremes}
\vspace{-3mm}
\end{figure}

\paragraph{Continuity}
\vspace{-2mm}
We find that the model is able to perform generation out the distribution of the original $[0, 5]$ range in which it was trained. Values over the range make the actions stronger, while negative values perform the opposite of the action. See \cref{fig:extremes}. This further showcases the model's ability to disentangle AUs and learn a semantically meaningful continuous distribution.

%The model is also able to perform some AUs on animals showing a strong and robust understanding of faces in general. See the supplementary.

%An important property of the AU encoder is to learn a continuous representation. We test this by using extreme values of certain AUs. By taking the negative of AU, upper lid raiser, one would expect it to be upper lid lowerer. For AU12, one could expect the negative to be AU15.

\subsection{Model Analysis}

\paragraph{AU Encoder}
\Cref{tab:au_encoders} presents the results for different AU encoders. The encoders vary from no encoding at all to individually encoding each AU to the CLIP space. A simple lifting from the AU vector to the CLIP space by using a MLP (row 2) tends to overfit due to the sparse AU label space, which can be seen from the low CLIP-I. Not encoding (row 1) the AUs at all results in the best CLIP-I values, but slightly lower AU MSE scores compared to Res + MLP64 (ours), in which a small MLP with 64 output size is used to encode more complex relationships in addition to the raw AU residual connection (res). This is particularly evident in the combination results, where encoding alone ranks only fourth. Hence we choose Res + MLP64 as the optimal method (ours). Further details and qualitative results of the different AU encoders are in the supplementary.

\begin{table}[]
\centering
\caption{Quantitative results with different AU encoders.}
\vspace{-2mm}
\resizebox{0.48\textwidth}{!}{%
\begin{tabular}{|c|cc|cc|}
    \hline
    \multirow{2}{4em}{AU Encoder} & \multicolumn{2}{|c|}{Individual} & \multicolumn{2}{|c|}{Combination} \\
     & AU MSE $\downarrow$ & CLIP-I $\uparrow$ & AU MSE $\downarrow$ & CLIP-I $\uparrow$ \\
    \hline
    \hline
    No encoding & \underline{4.72} & \textbf{0.94} & 7.64 & \textbf{0.86} \\
    \hline
    MLP & 4.79 & 0.86 & \underline{7.60} & 0.76 \\
    \hline
    Res + MLP & 4.77 & 0.84 & 7.61 & 0.74 \\
    \hline
    \textbf{Res + MLP64 (Ours)} & \textbf{4.71} & \underline{0.92} & \textbf{7.54} & \underline{0.83} \\
    \hline
    Res + 3MLP & 4.97 & 0.91 & 7.70 & \underline{0.83} \\
    \hline
    Individual Encoding + MLP & 5.48 & 0.85 & 7.97 & 0.70 \\
    \hline
\end{tabular}}
\label{tab:au_encoders}
\end{table}

\paragraph{Distribution Smoothing}
The impact of the distribution smoothing presented in \cref{sec:distribution_smoothing} is shown in \cref{tab:distribution_smoothing}. Large improvements in the CLIP-I can be observed. Improvements in prompt adherence are particularly noticeable in out-of-distribution cases in qualitative results, which the model has not seen during training.

\begin{table}[]
\centering
\caption{Ablation study on aplying distribution smoothing.}
\vspace{-2mm}
\resizebox{0.48\textwidth}{!}{%
\begin{tabular}{|c|cc|cc|}
    \hline
    \multirow{2}{4em}{Method} & \multicolumn{2}{|c|}{Individual} & \multicolumn{2}{|c|}{Combination} \\
     & AU MSE $\downarrow$ & CLIP-I $\uparrow$ & AU MSE $\downarrow$ & CLIP-I $\uparrow$ \\
    \hline
    \hline
    Normal & 4.81 & 0.81 & 7.64 & 0.72 \\
    \hline
    \textbf{Distribution Smoothing (ours)} & \textbf{4.71} & \textbf{0.92} & \textbf{7.54} & \textbf{0.83} \\
    \hline
\end{tabular}}
\label{tab:distribution_smoothing}
\vspace{-5mm}
\end{table} 

\vspace{-2mm}
\paragraph{Limitations}
As the model is based on Stable Diffusion \cite{stablediffusion} it inherits its limitations and biases. The used data limits the model's capabilities as only 12 different AUs are labeled and the labels are coded symmetrically. Prompt adherence and character consistency is still a common problem with diffusion based models \cite{ipadapter, instantid, consistentid}.

\section{Summary}
In this work we introduce the use of AUs as conditioning signals for controlling facial expressions in generated content with T2I diffusion models. The work lays groundwork for the future, as we propose techniques for handling the problems associated with continuity and multi-label nature of AUs. FineFace, a robust method with an AU adapter is capable of retaining the base diffusion model's capabilities and is compatible with image prompt adapters. The proposed method's abilities are shown through qualitative and quantitative studies. In future works, we aim to develop improved solutions for the issues with continuous multi-label AUs and expand to highly controlled facial image editing.

%%%%%%%%% REFERENCES
{\small
\bibliographystyle{ieee_fullname}
\bibliography{main}
}

\clearpage
\twocolumn[{
\renewcommand\twocolumn[1][]{#1}
\centering
\Large
\textbf{Towards Localized Fine-Grained Control for Facial Expression Generation}\\
\vspace{1.5em}Supplementary Material \\
\vspace{1.0em}
}]

\section*{Experimental Settings}
The Diffusers library \footnote{\url{https://github.com/huggingface/diffusers}} is used for training and extracting models. Batch size of 16 is used with a learning rate of $1e^{-4}$. For LoRA \cite{lora} based methods, $\alpha$ is set to default. For the implemented AU-Adapter the $\lambda_{AU}$ from \cref{eq:ip_adapter} is set to 1.0 for both training and inference. The small perturbation added to the labels in distribution smoothing \cref{sec:distribution_smoothing} is drawn from a Gaussian distribution with $\mu = 0$ and $\sigma^2 = 0.2$. After the perturbation, the values are clipped to the original $[0, 5]$ range. To ensure good performance during inference, in which typically integer values are used, the values are randomly quantized to integers with a 20\% probability.

For results which use the IP-Adapter \cite{ipadapter} the \cref{eq:ip_adapter} is modified to
\begin{align}
    \mathbf{Z} = & \attention(Q_{noise}, K_{text}, V_{text}) \nonumber\\ 
               & + \lambda_{img} \cdot \attention(Q_{noise}, K_{img}, V_{img}) \nonumber\\  & + \lambda_{AU} \cdot \attention(Q_{noise}, K_{AU}, V_{AU}),
\label{eq:image_prompt}
\end{align}
where both $\lambda_{img}$ and $\lambda_{AU}$ are set to 1. The adapters are trained separately and the \cref{eq:image_prompt} is used only during inference.

\section*{Dataset Construction}
This section provides further details on the dataset processing. AffectNet \cite{affectnet} is first filtered from low-quality images by keeping only images with both height and width being at least 512. Next cartoon and non-photorealistic images such as drawings are removed by using a VQA model, BLIP-2 \cite{blip-2}, with prompts such as "Is this a cartoon?" and "Is this photo real?". Next, the images are annotated with an automatic AU intensity method LibreFace \cite{libreface}. It should be noted that the performance of the annotation tool is not fully reliable as AU intensity prediction is still an on-going research problem. Therefore further filtering and corrections are provided to the AUs manually.

Samples with a total intensity from all AUs combined having less than 0.2 are filtered. Next corrections are made to \qcr{AU1} and \qcr{AU4} by scaling them down as the prediction intensities are too high based on visual analysis and the distribution of intensities is skewed to higher intensity values. The scaling down is achieved by utilizing a power transform $x_{transformed} = x^\gamma$, where $\gamma$ is the scaling factor and is set to 1.8. To ensure the $[0, 5]$ scaling, min-max normalization is used. The opposite observations are made for \qcr{AU15}, \qcr{AU17} and \qcr{AU20}, where the distribution is more skewed towards the small values. Hence the same operation but with a scaling factor of 0.8 is applied. Finally, due to the noisiness of the model and the inherent difficult of observing low intensity AUs from static images, all AUs with intensity values less than 1 are scaled down to 0. Therefore, the model is relying on the DISFA dataset's annotation of low intensity values, which are based on videos and manual annotations, making them much more reliable. For DISFA no such processing is done for AUs, except filtering most of the samples with no AUs at all. Around 10\% of the no AUs are left for learning guidance \cite{cfg}.

\begin{figure}
\begin{center}
    \setlength{\tabcolsep}{1pt}
    \begin{subfigure}{0.47\textwidth}
    \hspace{-2mm}
    \begin{tabular}{*5c}
        \multicolumn{5}{c}{\qcr{AU4}}  \\
        \includegraphics[width=0.2\textwidth]{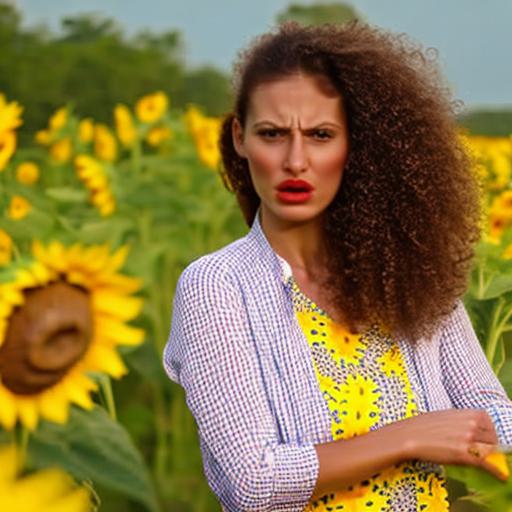} &
        \includegraphics[width=0.2\textwidth]{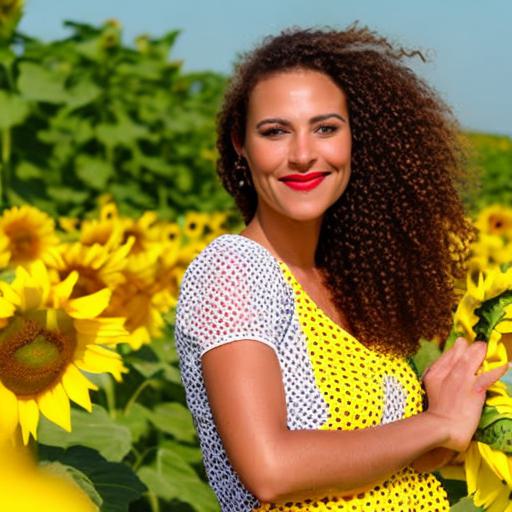} &
        \includegraphics[width=0.2\textwidth]{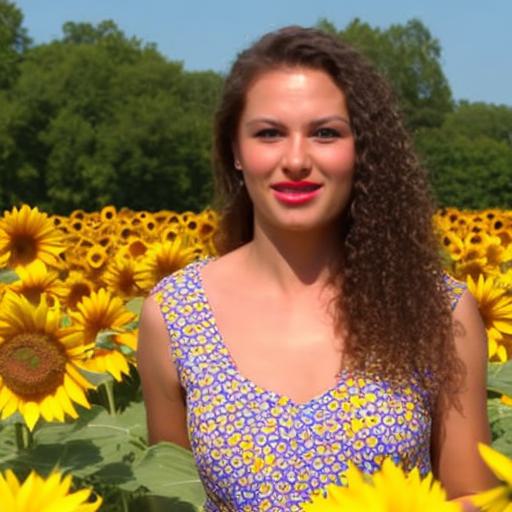} &
        \includegraphics[width=0.2\textwidth]{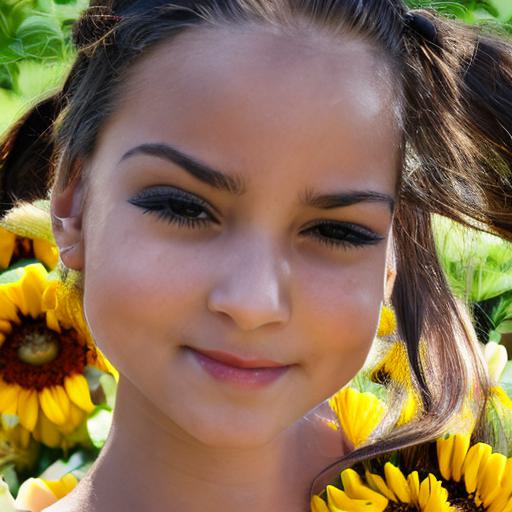} & 
        \includegraphics[width=0.2\textwidth]{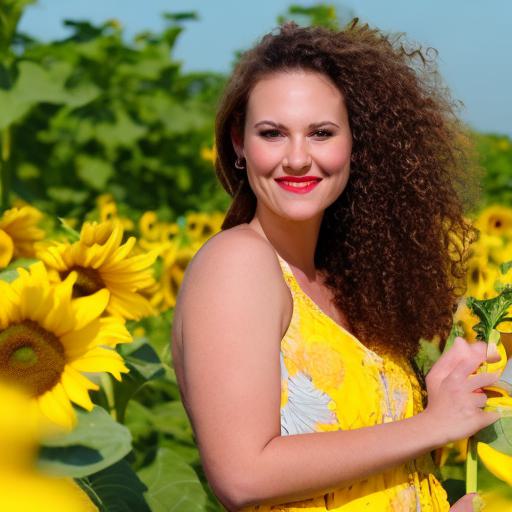}
        \\
        \footnotesize{FineFace} & \footnotesize{LoRA-AU} & \footnotesize{LoRA-T} & \footnotesize{DB} & \footnotesize{SD} \\
        \end{tabular}
    \end{subfigure}%
\end{center}
\vspace{-6mm}
\caption{
\qcr{AU4} with the prompt \textit{A girl wearing a sundress in a sunflower field}. DB has overfit to the portrait training data.
}
\label{fig:flower}
\vspace{-3mm}
\end{figure}

\section*{AU Encoders}
The AU encoder plays an important part in transforming the raw AU vector to a representation that can be used by the $K$ and $V$ projection matrices in the adapter (see \cref{eq:ip_adapter}). In this section we go over the details of the different AU encoders showcased in \cref{tab:au_encoders}. 

The no-encoding approach takes in the raw AU vector (of length 12) and concatenates 1012 zeros after it to match the dimension of 1024 used by CLIP \cite{clip} space. This approach ensures the continuity of AUs, but does not encore the interactions of multiple AUs. On the second row, the MLP approach lifts the AU vector to the CLIP space, which ensures strong embeddings of multiple AUs but tends to overfit due to the sparsity of the available labels.

The proposed approach uses the next method, Res + MLP, but with a smaller MLP of 64 output dimension. Res + MLP uses a MLP encoder with a residual connection from the AU vector to gain benefits of both approaches above. In practice, concatenation of the 12 AU values and 1012 MLP output is used as the final result. In the Res + MLP64, the remaining values are padded with zeros as it was found the lifting from a 12-dimensional space to 1012 can be difficult to learn and lead to overfitting. Res + MLP3 uses a more complex three layer MLP with normalization and leaky ReLUs, in an attempt to better encoder the interactions between AUs. Finally, the individual encoding + MLP approach uses a 1024 dimensional MLP for each individual AU to encode the indidual AUs, and an additional MLP to cover the interactions.

\section*{Further Analysis}
Further analysis is provided on the prompt adherence and decision to use LoRAs instead of fine-tuning the entire network. The result can be seen from \cref{fig:flower}, where it can be seen that DB has overfit to the portrait style facial images present in the dataset. LoRA-T is unable to keep the original composure of the prompt, which can be seen from the SD result. LoRA-AU is unable to produce the AU condition, but better preserves the prompt.

\begin{figure}
\begin{center}
    \setlength{\tabcolsep}{1pt}
    \begin{subfigure}{0.47\textwidth}
    \hspace{-2mm}
    \begin{tabular}{*4c}
        \includegraphics[width=0.25\textwidth]{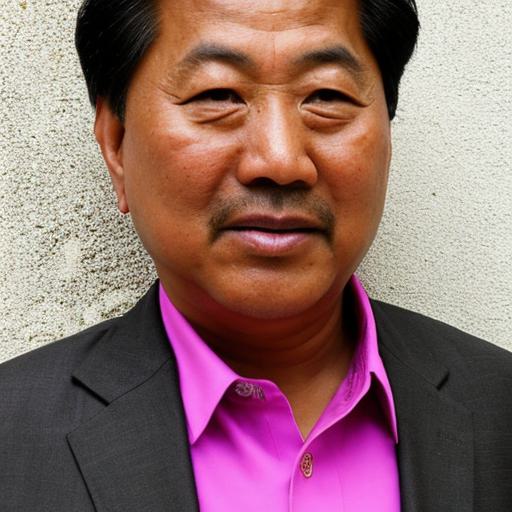} &
        \includegraphics[width=0.25\textwidth]{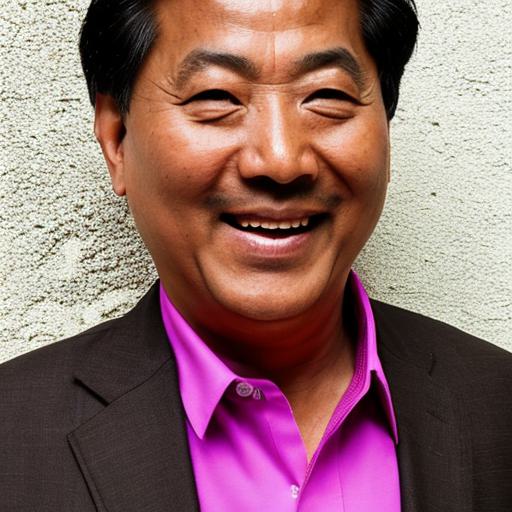} &
        \includegraphics[width=0.25\textwidth]{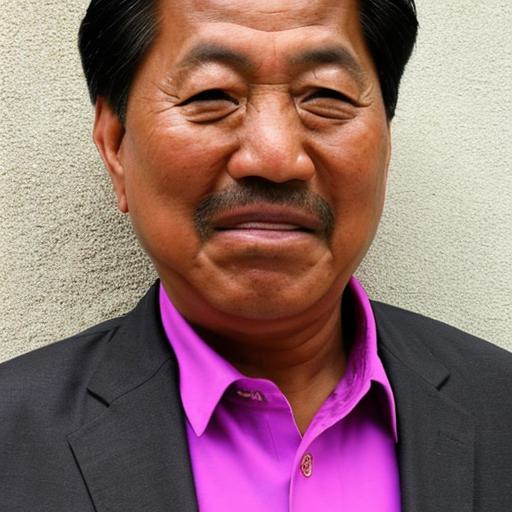} &
        \includegraphics[width=0.25\textwidth]{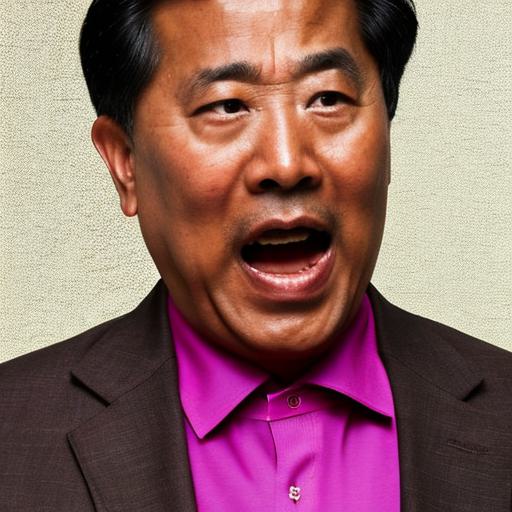}
        \\
        \multicolumn{4}{c}{Distribution Smoothing}  \\
        \includegraphics[width=0.25\textwidth]{figures/aus/resmlp64/pink_shirt/neutral.jpg} &
        \includegraphics[width=0.25\textwidth]{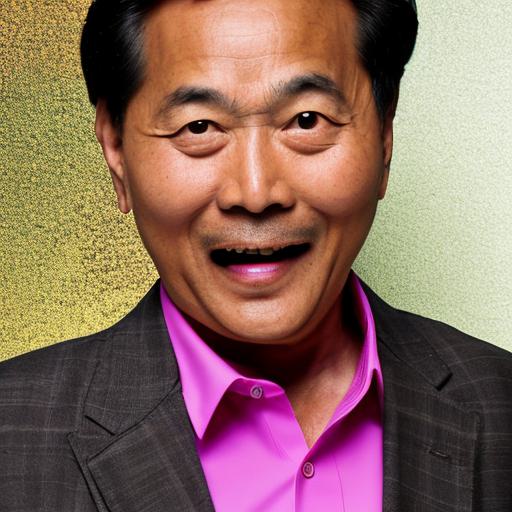} &
        \includegraphics[width=0.25\textwidth]{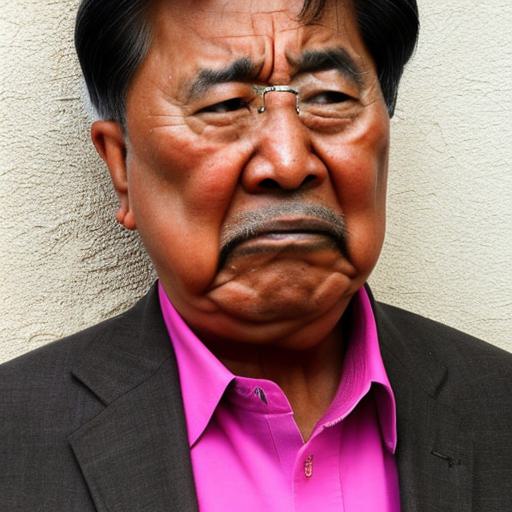} &
        \includegraphics[width=0.25\textwidth]{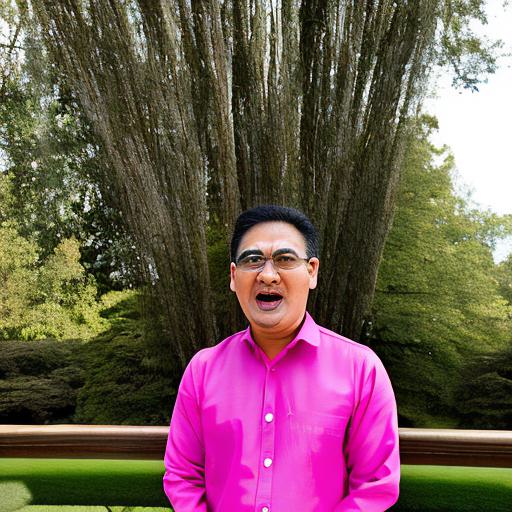}
        \\
        \multicolumn{4}{c}{No Distribution Smoothing}  \\
        \footnotesize{Neutral} & \footnotesize{\qcr{AU12+25+26}} & \footnotesize{\qcr{AU4+15+17}} & \footnotesize{\qcr{AU4+25+26}} \\
        \end{tabular}
    \end{subfigure}%
\end{center}
\vspace{-6mm}
\caption{
An ablation of using distribution smoothing, with the prompt \textit{An asian man in a pink shirt}. Without smoothing the AU distribution, the model can overfit to specific AUs and significantly diminish the prompt adherence.
}
\label{fig:smoothing_comparison}
\vspace{-3mm}
\end{figure}

\Cref{fig:smoothing_comparison} compares the results with using and not using distribution smoothing during training. Without smoothing, a given AU may significantly alter the intermediate features, causing the generated image to deviate from the prompt. This can be seen with the \qcr{AU4+25+26}, where the image drastically changes. 

\section*{AU Encoder Qualitative Results}
\begin{figure*}
\begin{center}
    \setlength{\tabcolsep}{1pt}
    \begin{subfigure}{1.02\textwidth}
    \hspace{-0.2cm}
\begin{tabular}{*{14}{c}}
        \rotatebox{90}{\footnotesize{No Enc}} &
        \includegraphics[width=0.0714\textwidth]{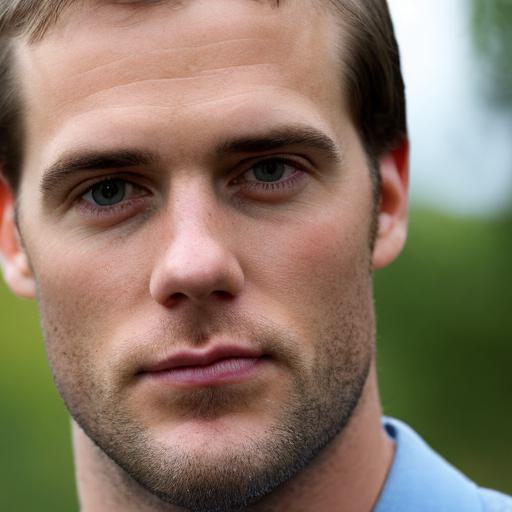} &
        \includegraphics[width=0.0714\textwidth]{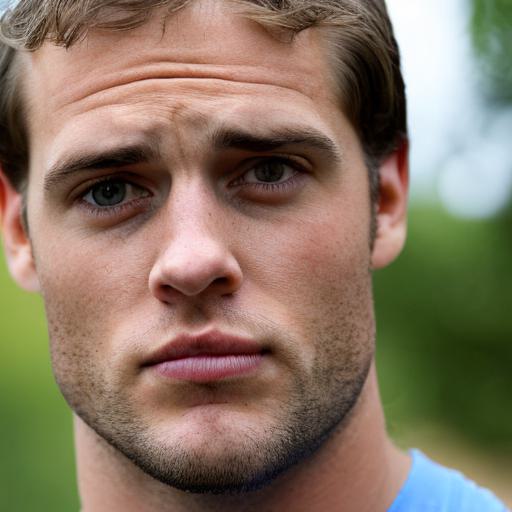} &
        \includegraphics[width=0.0714\textwidth]{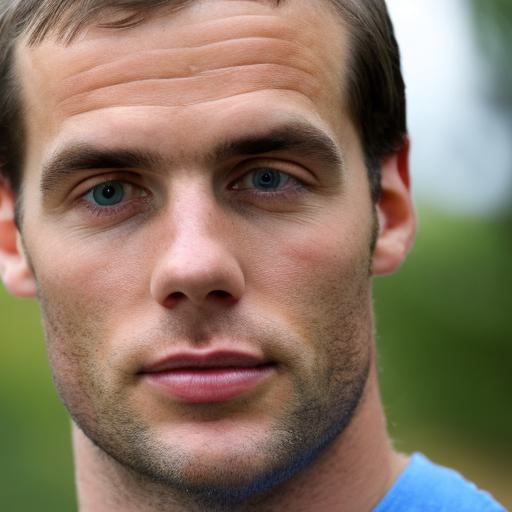} &
        \includegraphics[width=0.0714\textwidth]{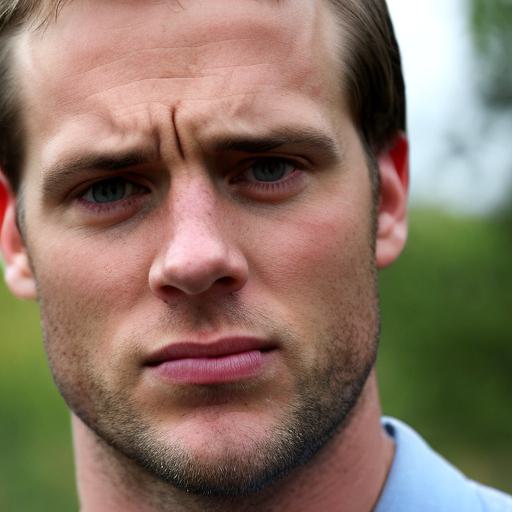} &
        \includegraphics[width=0.0714\textwidth]{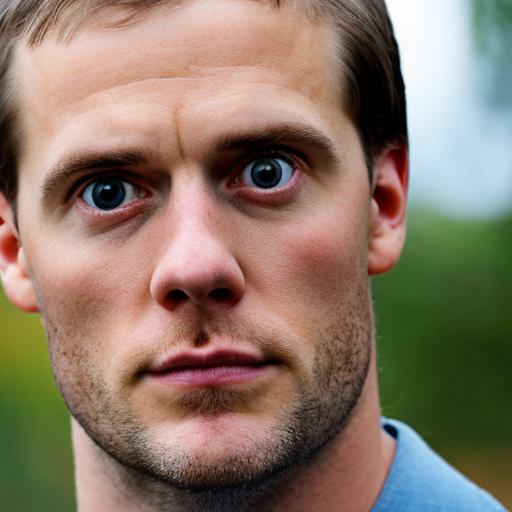} &
        \includegraphics[width=0.0714\textwidth]{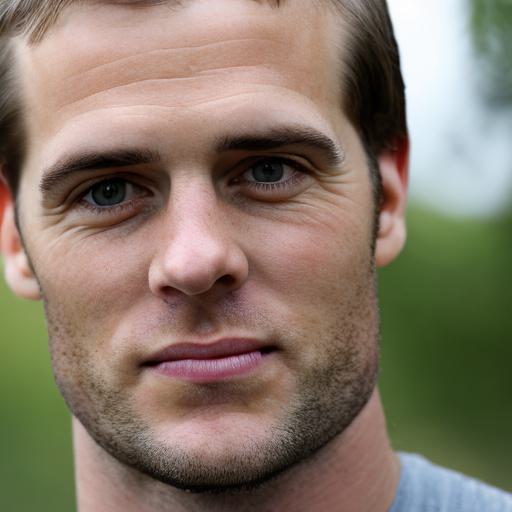} &
        \includegraphics[width=0.0714\textwidth]{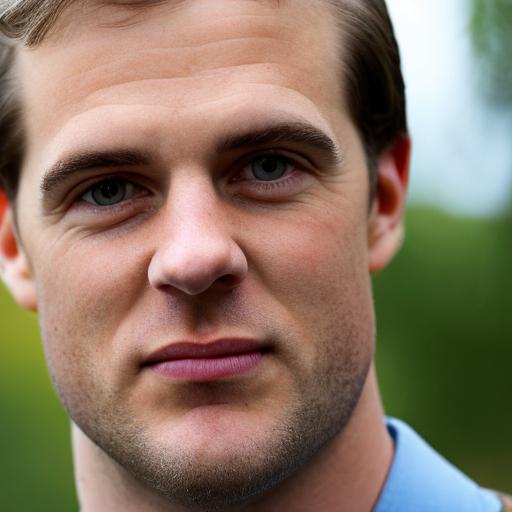} &
        \includegraphics[width=0.0714\textwidth]{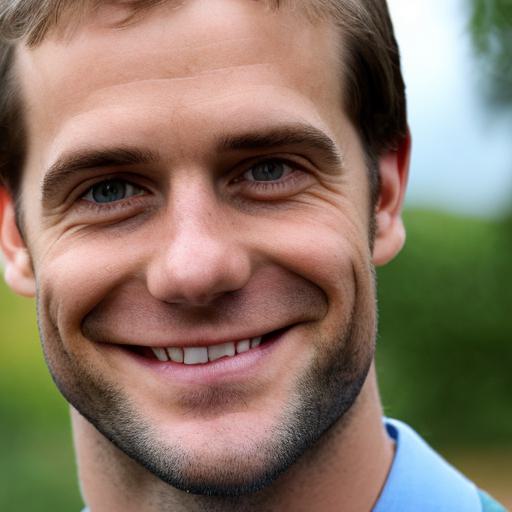} &
        \includegraphics[width=0.0714\textwidth]{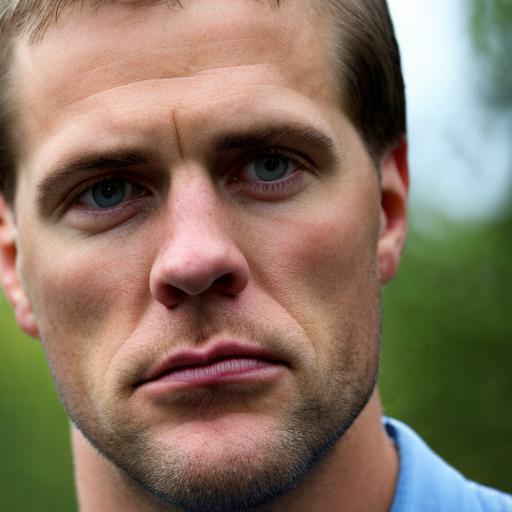} &
        \includegraphics[width=0.0714\textwidth]{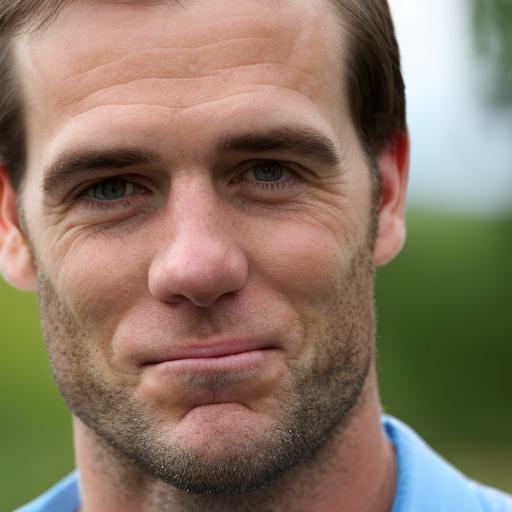} &
        \includegraphics[width=0.0714\textwidth]{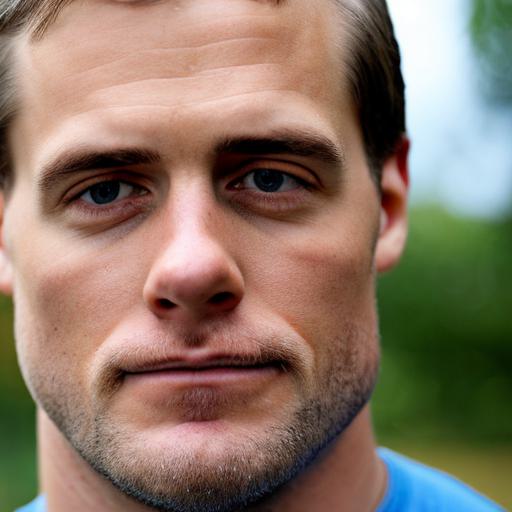} &
        \includegraphics[width=0.0714\textwidth]{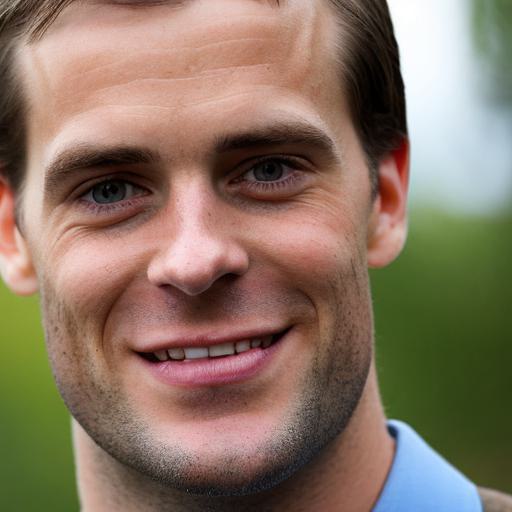} &
        \includegraphics[width=0.0714\textwidth]{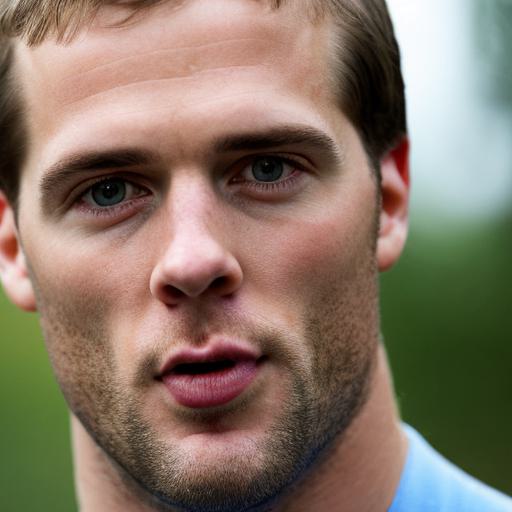} 
        \\
\end{tabular}
    \end{subfigure}%
\end{center}
\caption{
Comparison of different methods on 12 individual AUs with the prompt \textit{A young man in a park}. See \cref{fig:aus} for the textual descriptions of AUs. 
}
\label{fig:encoder_comparison}
\vspace{-3mm}
\end{figure*}

\Cref{fig:encoder_comparison} compares the different AU encoders in individual AU generation. \textit{No Enc} refers to the no encoding method, which performs exceptionally well with both the AU generation and retaining the character. It is possible to see even the difficult \qcr{AU9} working reasonably. The MLP approach struggles with both the AUs and retaining the character. The results are similar for Res+MLP. Res+MLP64 performs very similarly to No Enc, small differences can be seen in \qcr{AU25} where Res+MLP64 is better. In terms of AU accuracy Res+MLP3 also performs similarly, but achieves a better result in \qcr{AU6}. \textit{IC}, referring to individual encoding, achieves good results in terms of AUs, although they tend to be a bit exaggerated, however the character consistency is off.

\begin{figure*}
\begin{center}
    \setlength{\tabcolsep}{1pt}
    \begin{subfigure}{1.02\textwidth}
    \hspace{-0.2cm}
    \begin{tabular}{*{14}{c}}
        \rotatebox{90}{\footnotesize{No Enc}} &
        \includegraphics[width=0.0714\textwidth]{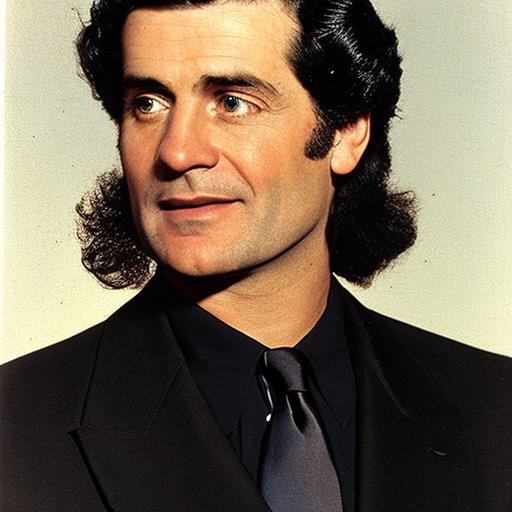} &
        \includegraphics[width=0.0714\textwidth]{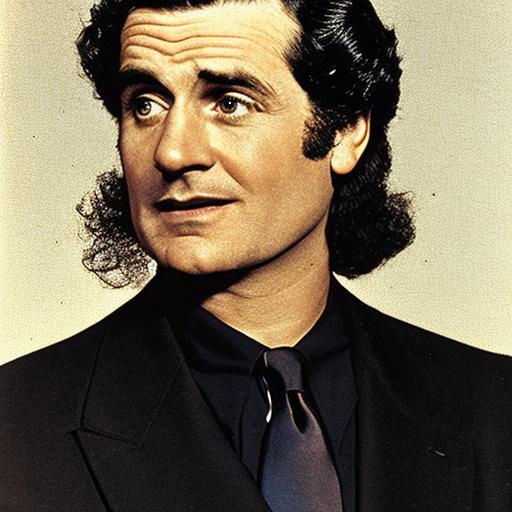} &
        \includegraphics[width=0.0714\textwidth]{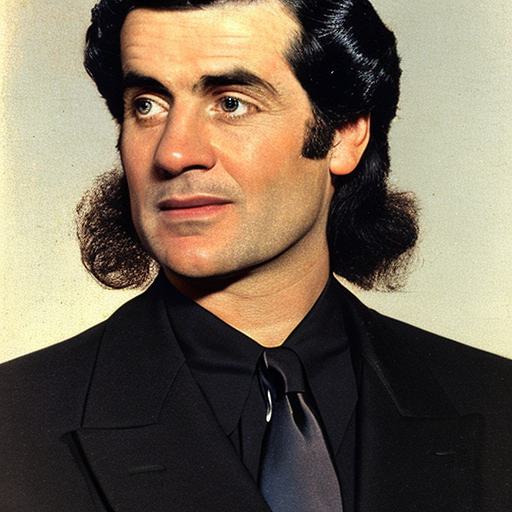} &
        \includegraphics[width=0.0714\textwidth]{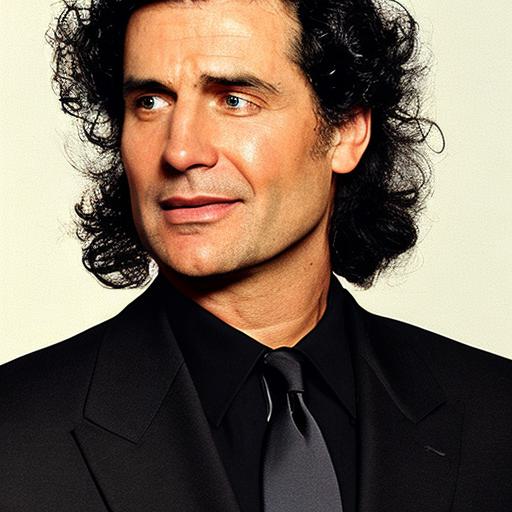} &
        \includegraphics[width=0.0714\textwidth]{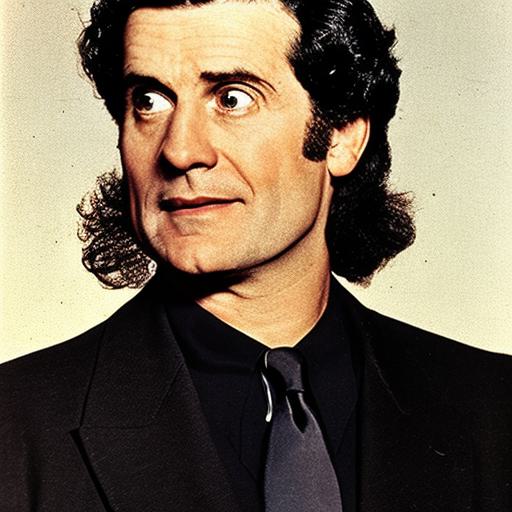} &
        \includegraphics[width=0.0714\textwidth]{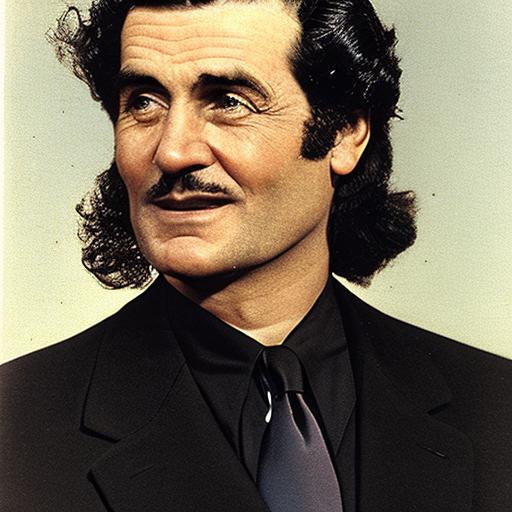} &
        \includegraphics[width=0.0714\textwidth]{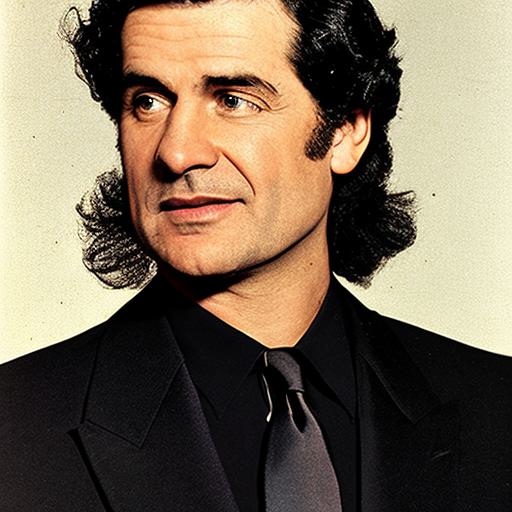} &
        \includegraphics[width=0.0714\textwidth]{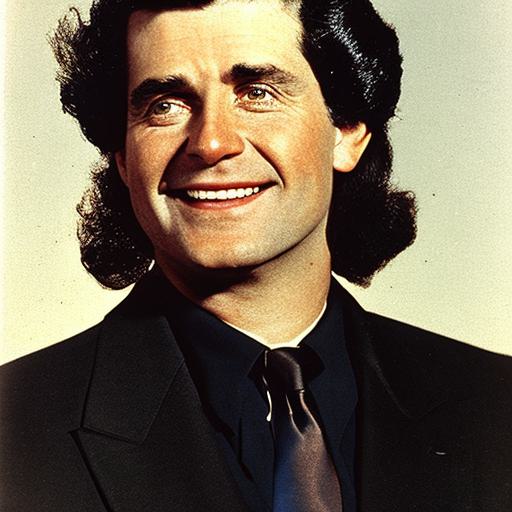} &
        \includegraphics[width=0.0714\textwidth]{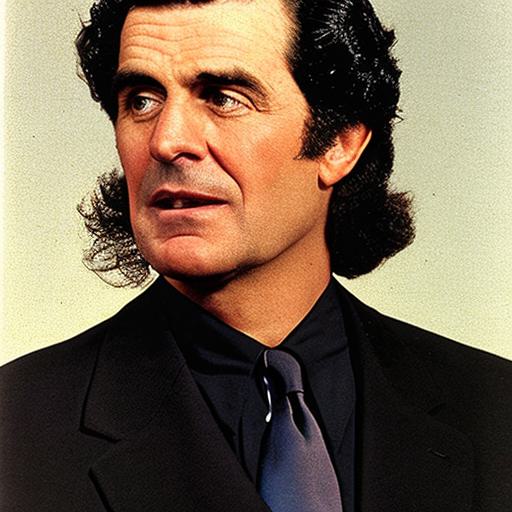} &
        \includegraphics[width=0.0714\textwidth]{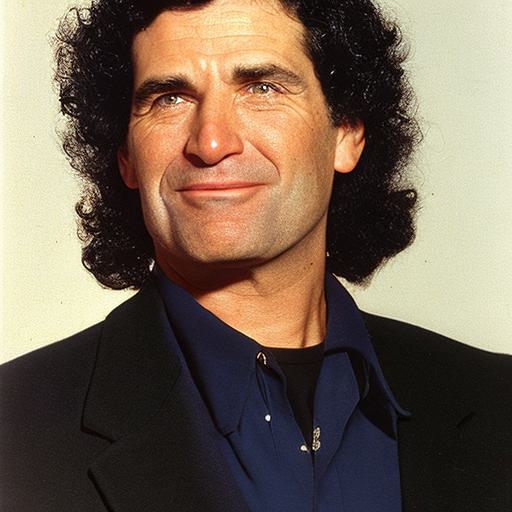} &
        \includegraphics[width=0.0714\textwidth]{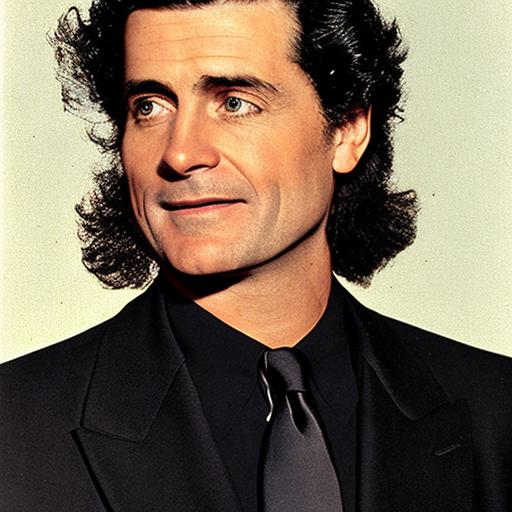} &
        \includegraphics[width=0.0714\textwidth]{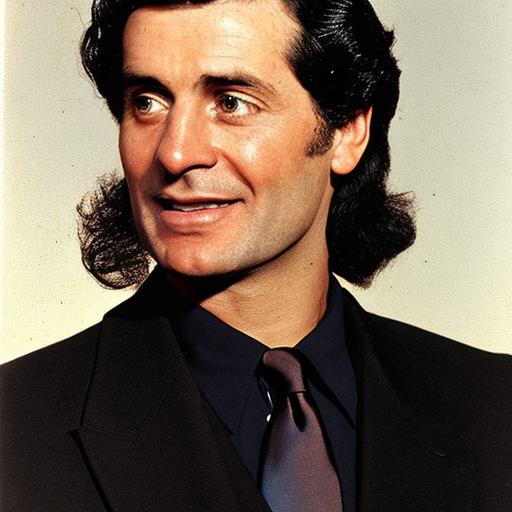} &
        \includegraphics[width=0.0714\textwidth]{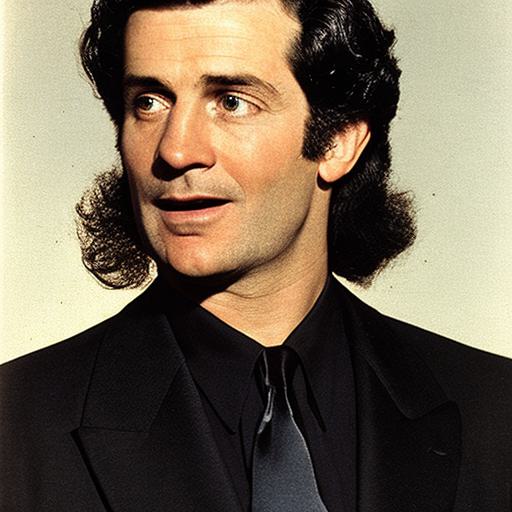} 
        \\
\end{tabular}
    \end{subfigure}%
\end{center}
\caption{
Comparison of different methods on 12 individual AUs with the prompt \textit{A man with black hair wearing a black jacket}. See \cref{fig:aus} for the textual descriptions of AUs. 
}
\label{fig:encoder_comparison2}
\vspace{-3mm}
\end{figure*}

Similar observations can be drawn from \cref{fig:encoder_comparison2}. A major observation overall is that the character consistency significantly breaks when the face is not taking most of the image and it is a side-view. This is likely caused by the training data including mostly frontal face images. The AU performance is also degraded compared to \cref{fig:encoder_comparison}, as \qcr{AU2} is not discernible using any of the methods.

\begin{figure*}
\begin{center}
    \setlength{\tabcolsep}{1pt}
    \begin{subfigure}{0.95\textwidth}
    \hspace{-6mm}
    \begin{tabular}{*{6}{c}}
        \multicolumn{6}{c}{\qcr{AU6+12+25}}  \\
        \includegraphics[width=0.166\textwidth]{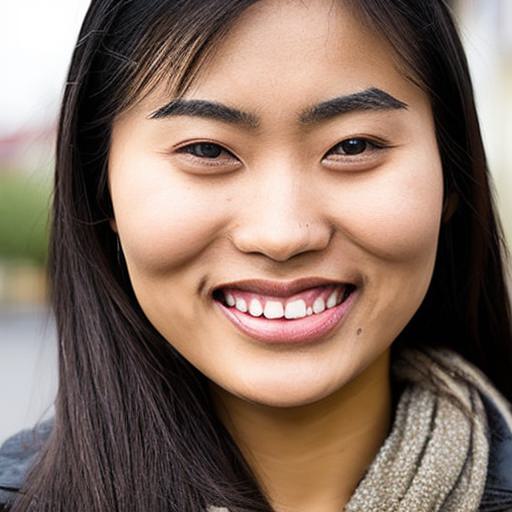} &
        \includegraphics[width=0.166\textwidth]{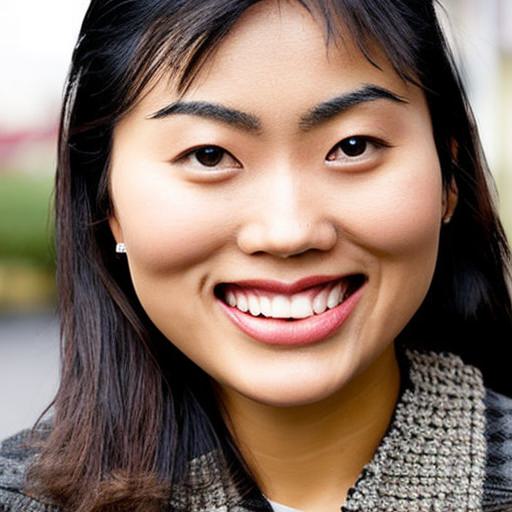} &
        \includegraphics[width=0.166\textwidth]{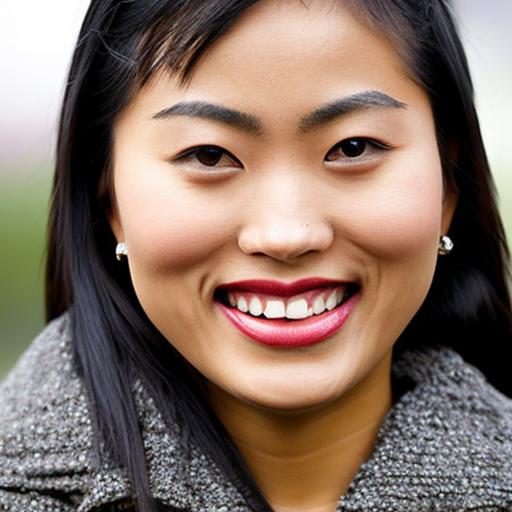} &
        \includegraphics[width=0.166\textwidth]{figures/aus/resmlp64/asian_woman/AU6+AU12+AU25+1,4,4.jpg} &
        \includegraphics[width=0.166\textwidth]{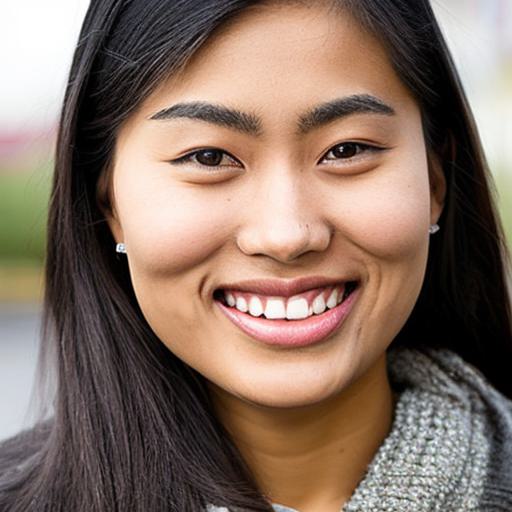} &
        \includegraphics[width=0.166\textwidth]{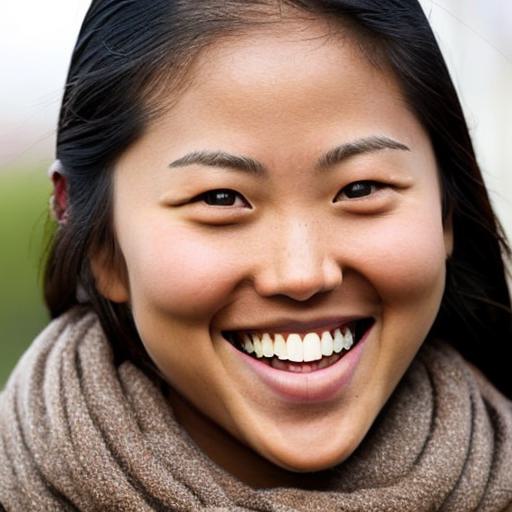}
        \\
        \multicolumn{6}{c}{\qcr{AU4+AU15+AU17}}  \\
        \includegraphics[width=0.166\textwidth]{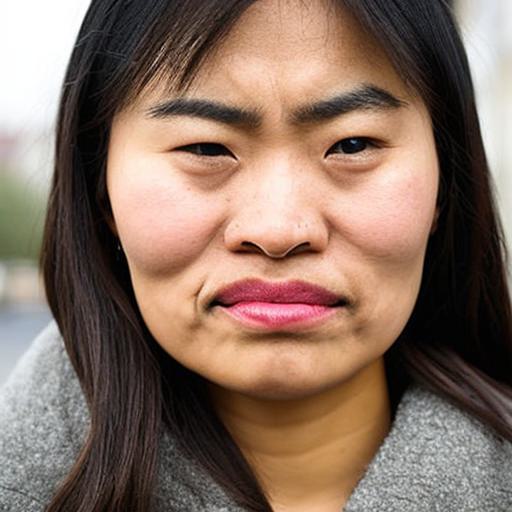} &
        \includegraphics[width=0.166\textwidth]{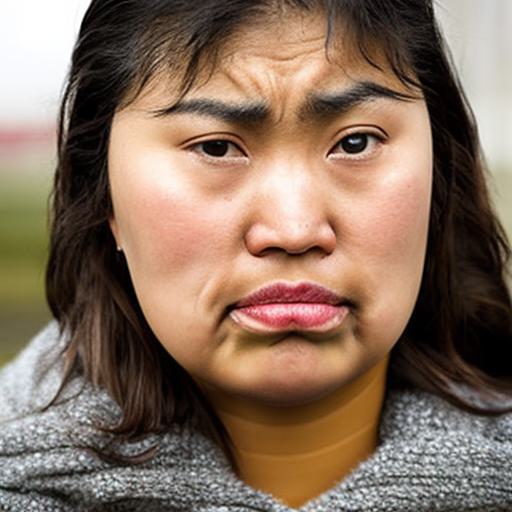} &
        \includegraphics[width=0.166\textwidth]{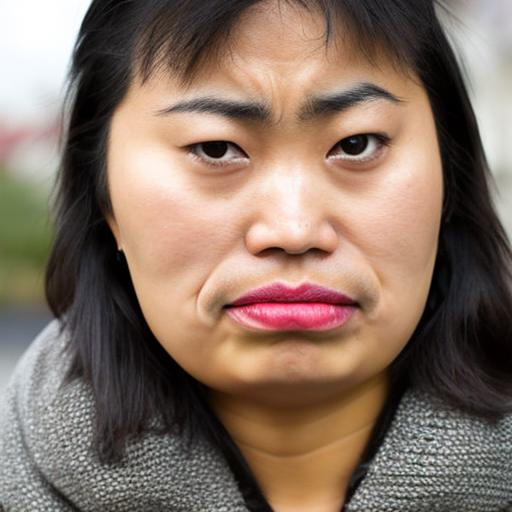} &
        \includegraphics[width=0.166\textwidth]{figures/aus/resmlp64/asian_woman/AU4+AU15+AU17+2,3,4.jpg} &
        \includegraphics[width=0.166\textwidth]{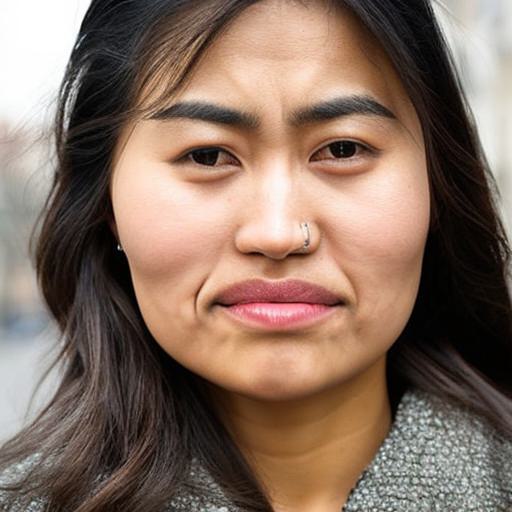} &
        \includegraphics[width=0.166\textwidth]{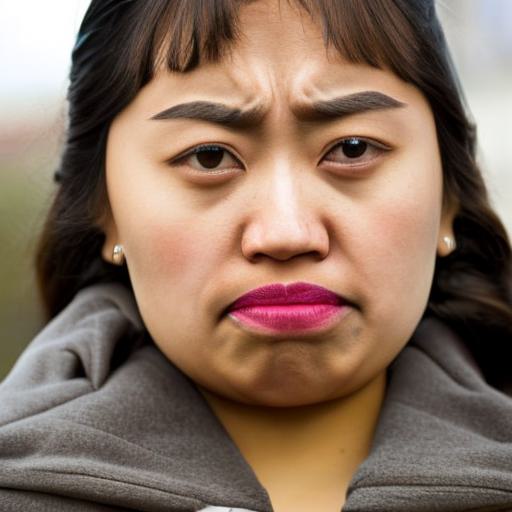}
        \\
        \multicolumn{6}{c}{\qcr{AU1+AU2+AU5+AU25+AU26}}  \\
\end{tabular}
    \end{subfigure}%
\end{center}
\caption{
Comparison of methods on combination AUs with the prompt \textit{An Asian woman in the park}.
}
\label{fig:encoder_combinations}
\vspace{-3mm}
\end{figure*}

\Cref{fig:encoder_combinations} showcases results of AU combinations with the AU encoders. Once again, MLP, Res+MLP and IC tends to have a poor character consistency. \textit{No Enc} generally performs well but struggles with the complex combination of \qcr{AU1+AU2+AU5+AU25+AU26}. It fails to generate \qcr{AU25} and \qcr{AU26} effectively and only partially captures \qcr{AU5}, while also introducing an artifact on the forehead. For \qcr{AU4+AU6+AU17+AU20} Res+MLP3 only creates a trace of \qcr{AU4} and struggles to retain the character for \qcr{AU1+AU2+AU4+AU9}.

\section*{Additional Qualitative Results}
\begin{figure*}
\begin{center}
    \setlength{\tabcolsep}{1pt}
    \begin{subfigure}{1.02\textwidth}
    \hspace{-0.2cm}
    \begin{tabular}{*{14}{c}}
        \rotatebox{90}{\footnotesize{\makecell{FineFace \\(Ours)}}} &
        \includegraphics[width=0.0714\textwidth]{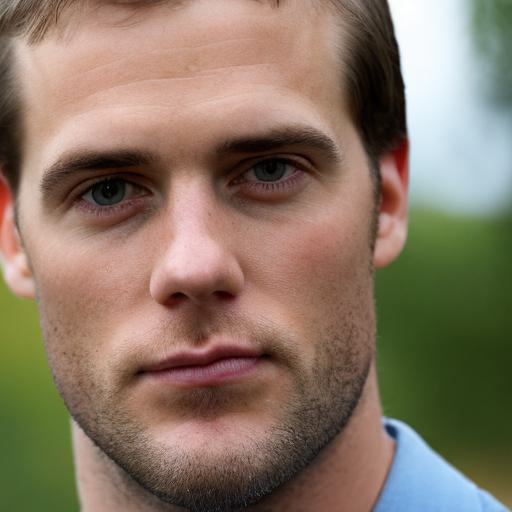} &
        \includegraphics[width=0.0714\textwidth]{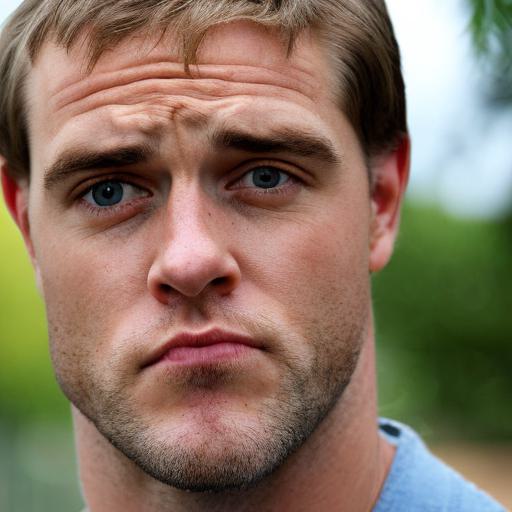} &
        \includegraphics[width=0.0714\textwidth]{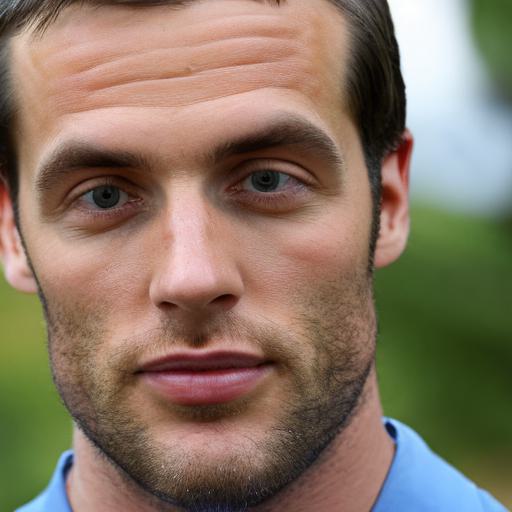} &
        \includegraphics[width=0.0714\textwidth]{figures/aus/resmlp64/man_park/AU4/5.jpg} &
        \includegraphics[width=0.0714\textwidth]{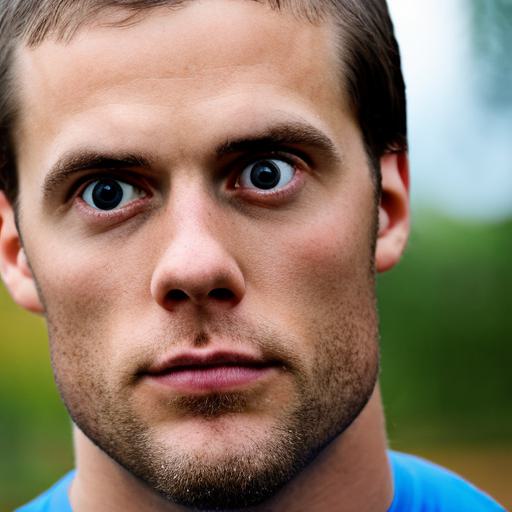} &
        \includegraphics[width=0.0714\textwidth]{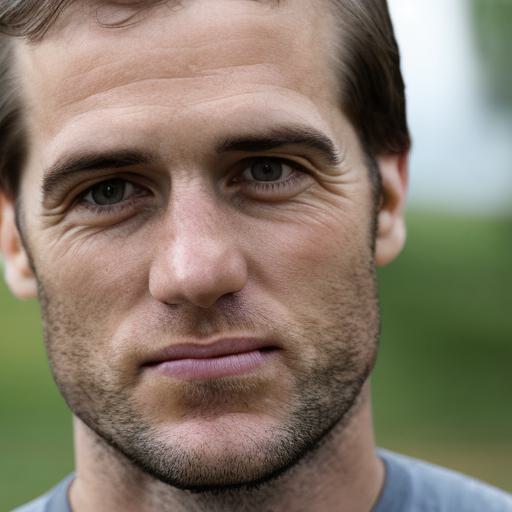} &
        \includegraphics[width=0.0714\textwidth]{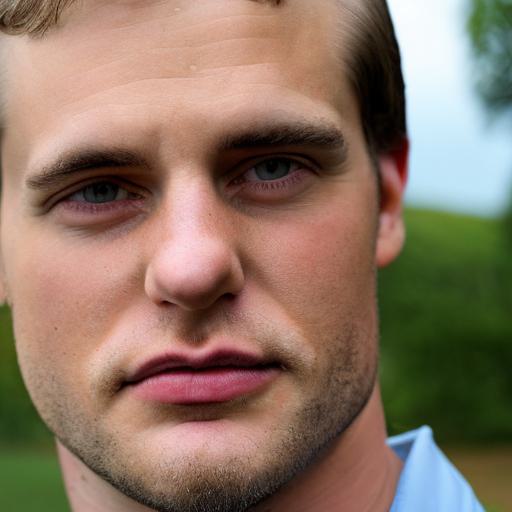} &
        \includegraphics[width=0.0714\textwidth]{figures/aus/resmlp64/man_park/AU12/5.jpg} &
        \includegraphics[width=0.0714\textwidth]{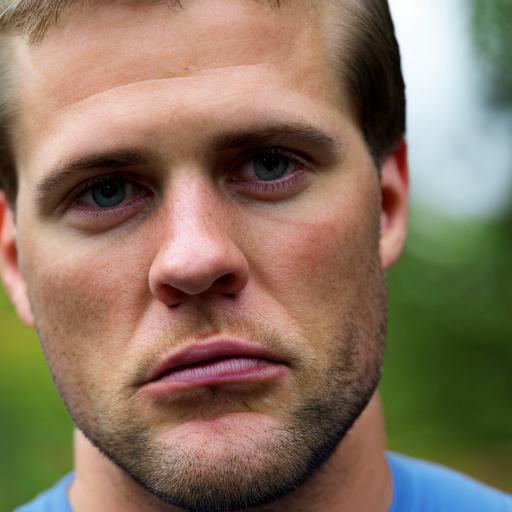} &
        \includegraphics[width=0.0714\textwidth]{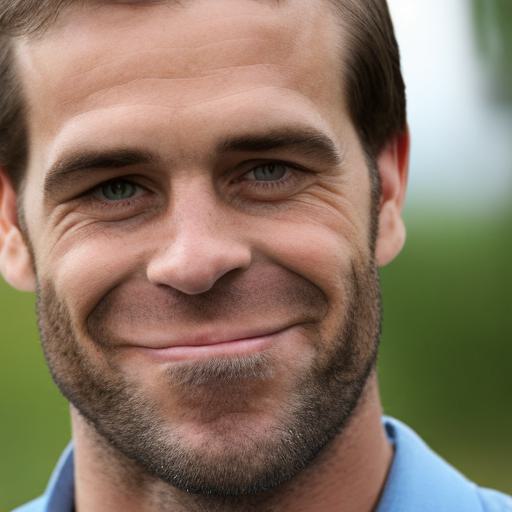} &
        \includegraphics[width=0.0714\textwidth]{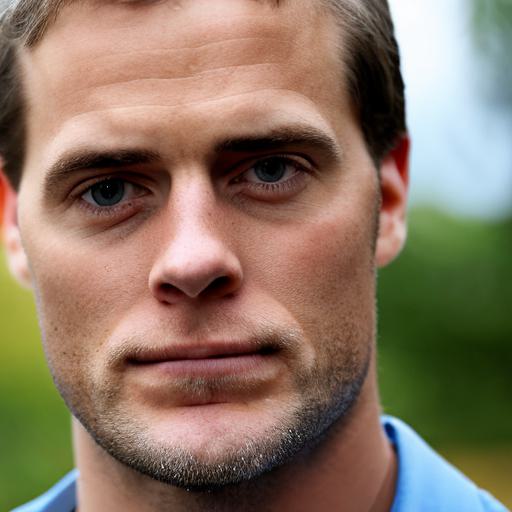} &
        \includegraphics[width=0.0714\textwidth]{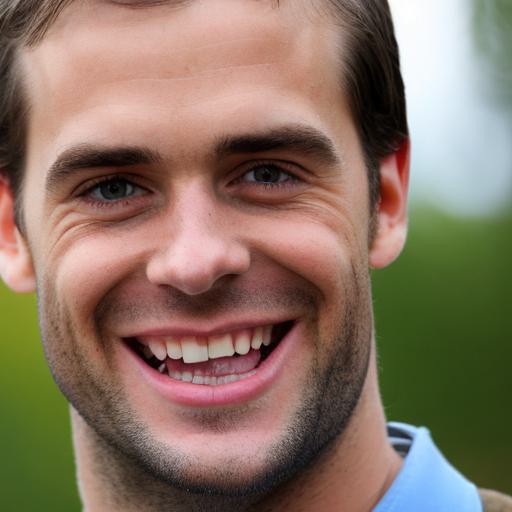} &
        \includegraphics[width=0.0714\textwidth]{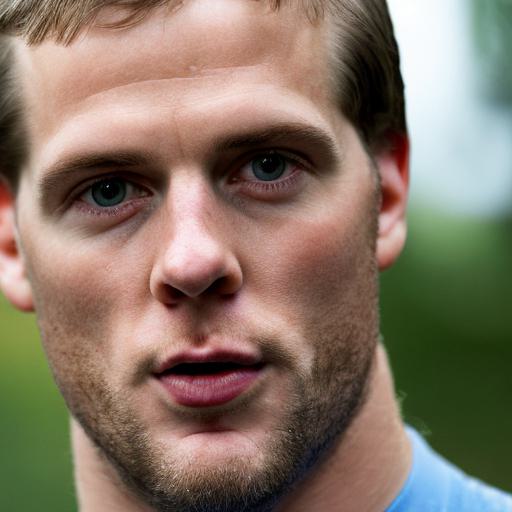} 
        \\
\end{tabular}
    \end{subfigure}%
\end{center}
\caption{
Comparison of different methods on 12 individual AUs with the prompt \textit{A young man in a park}. See \cref{fig:aus} for the textual descriptions of AUs. 
}
\label{fig:au_comparison2}
\vspace{-3mm}
\end{figure*}

\begin{figure*}
\begin{center}
    \setlength{\tabcolsep}{1pt}
    \begin{subfigure}{1.02\textwidth}
    \hspace{-0.2cm}
    \begin{tabular}{*{14}{c}}
        \rotatebox{90}{\footnotesize{\makecell{FineFace \\(Ours)}}} &
        \includegraphics[width=0.0714\textwidth]{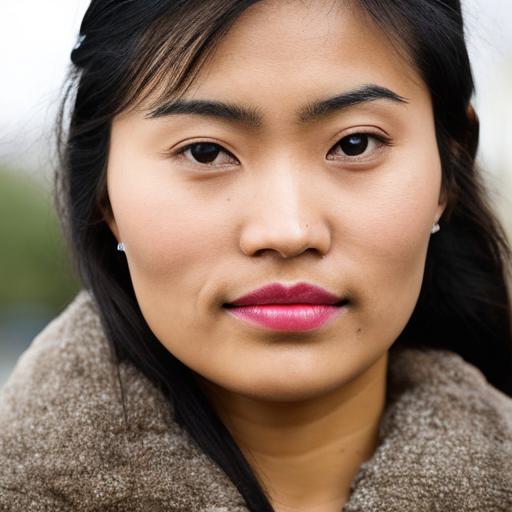} &
        \includegraphics[width=0.0714\textwidth]{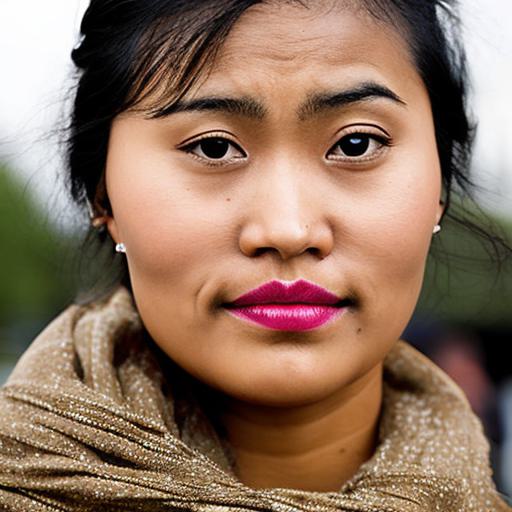} &
        \includegraphics[width=0.0714\textwidth]{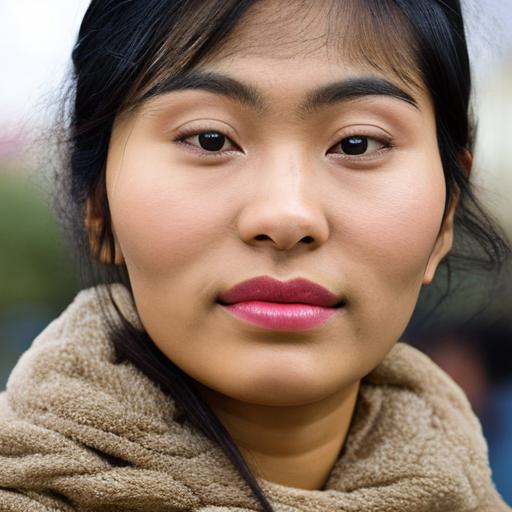} &
        \includegraphics[width=0.0714\textwidth]{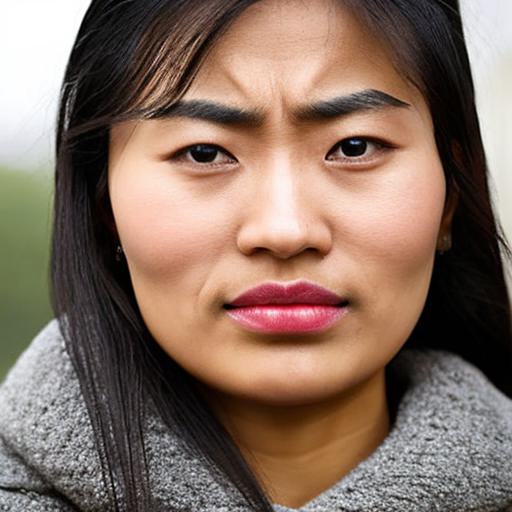} &
        \includegraphics[width=0.0714\textwidth]{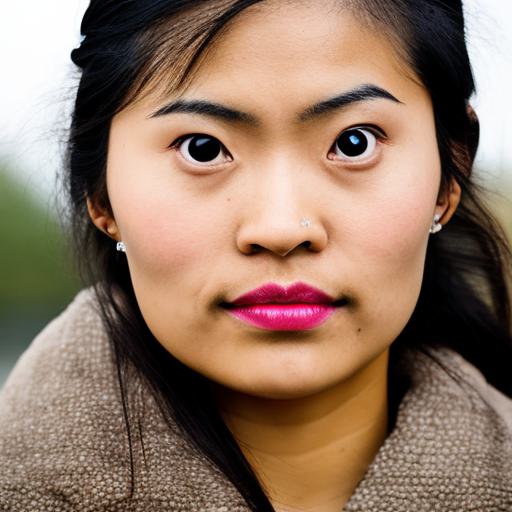} &
        \includegraphics[width=0.0714\textwidth]{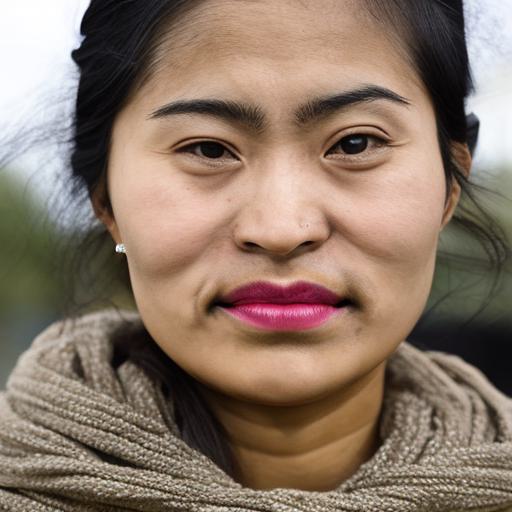} &
        \includegraphics[width=0.0714\textwidth]{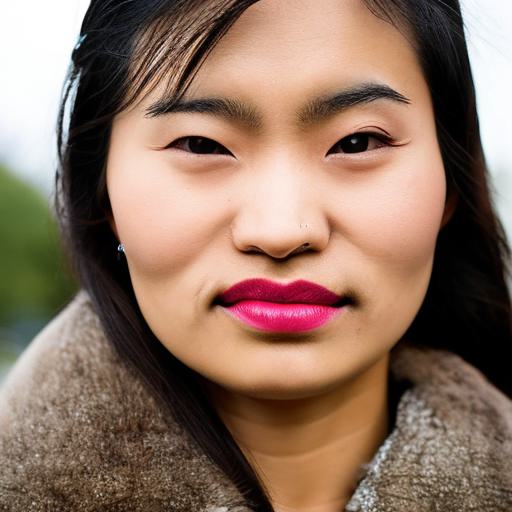} &
        \includegraphics[width=0.0714\textwidth]{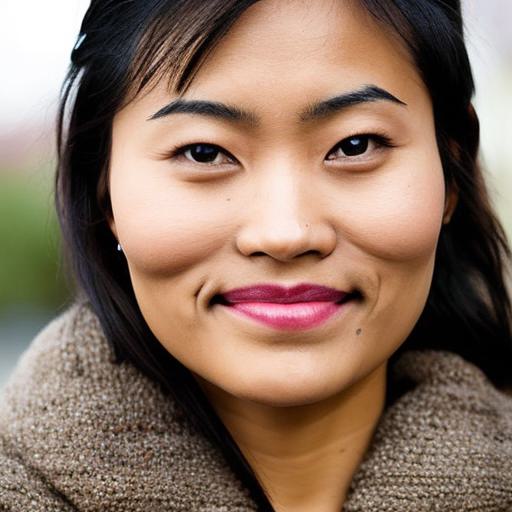} &
        \includegraphics[width=0.0714\textwidth]{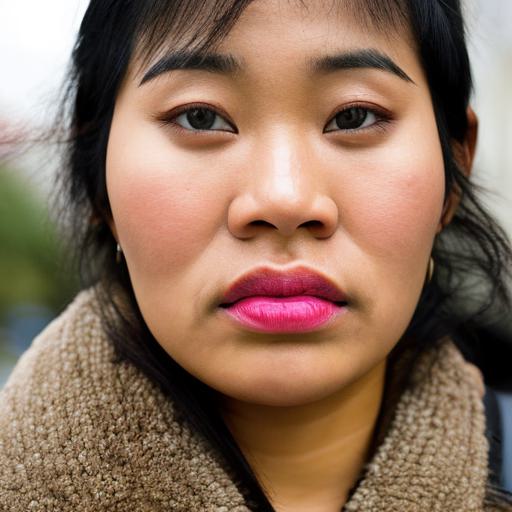} &
        \includegraphics[width=0.0714\textwidth]{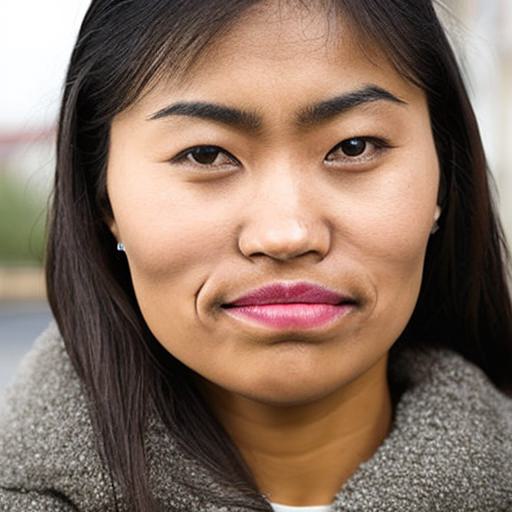} &
        \includegraphics[width=0.0714\textwidth]{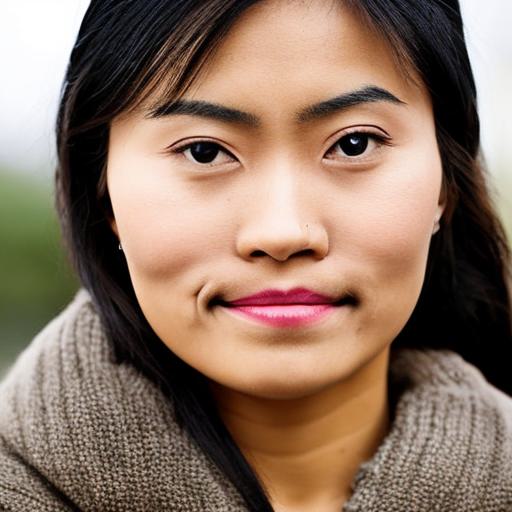} &
        \includegraphics[width=0.0714\textwidth]{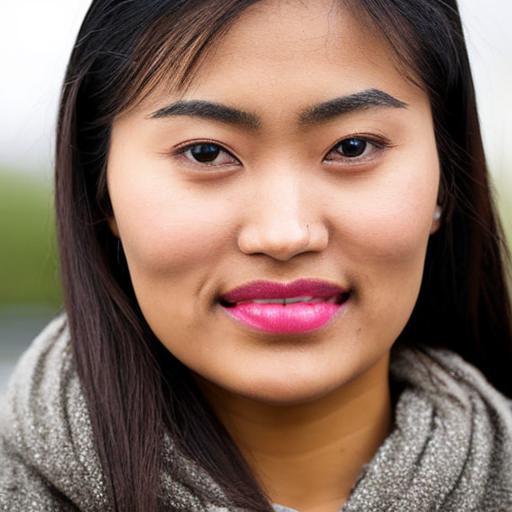} &
        \includegraphics[width=0.0714\textwidth]{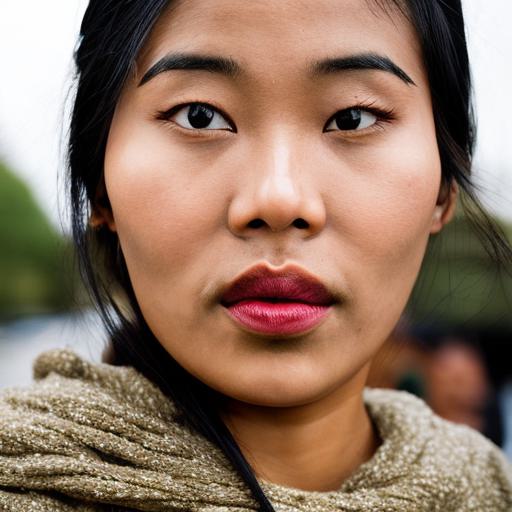} 
        \\
\end{tabular}
    \end{subfigure}%
\end{center}
\caption{
Comparison of different methods on 12 individual AUs with the prompt \textit{An Asian woman in the park.}.
}
\label{fig:au_comparison3}
\vspace{-3mm}
\end{figure*}

\begin{figure*}
\begin{center}
    \setlength{\tabcolsep}{1pt}
    \begin{subfigure}{0.95\textwidth}
    \hspace{-6mm}
    \begin{tabular}{*{5}{c}@{\hskip 8pt} *{5}{c}}
        \multicolumn{5}{c}{\qcr{AU6+12+25}} & \multicolumn{5}{c}{\qcr{AU4+15+17}}  \\
        \includegraphics[width=0.1\textwidth]{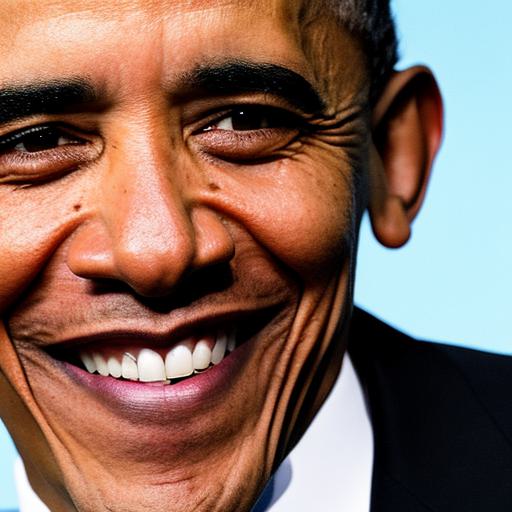} &
        \includegraphics[width=0.1\textwidth]{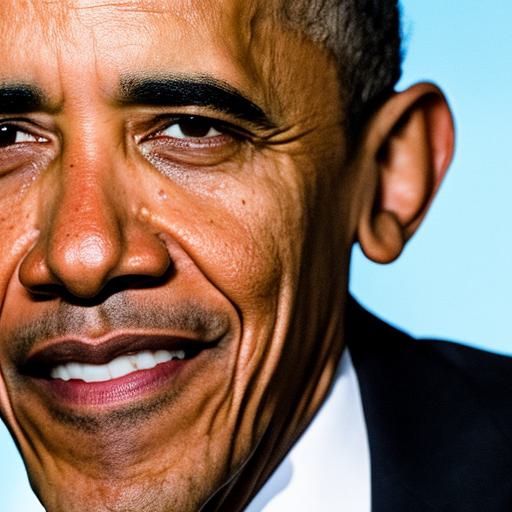} &
        \includegraphics[width=0.1\textwidth]{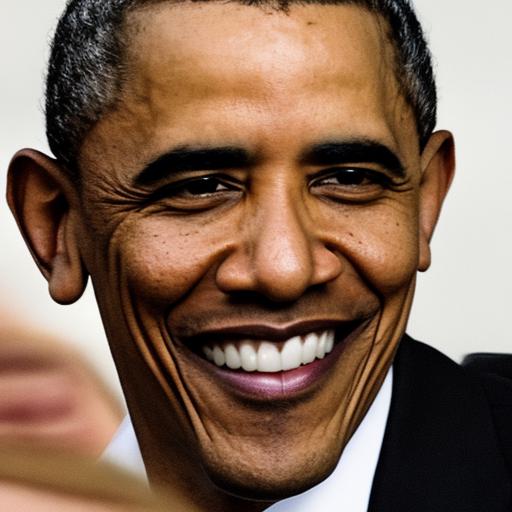} &
        \includegraphics[width=0.1\textwidth]{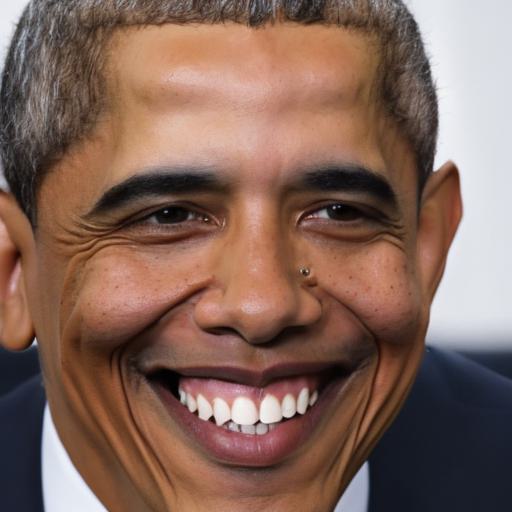} &
        \includegraphics[width=0.1\textwidth]{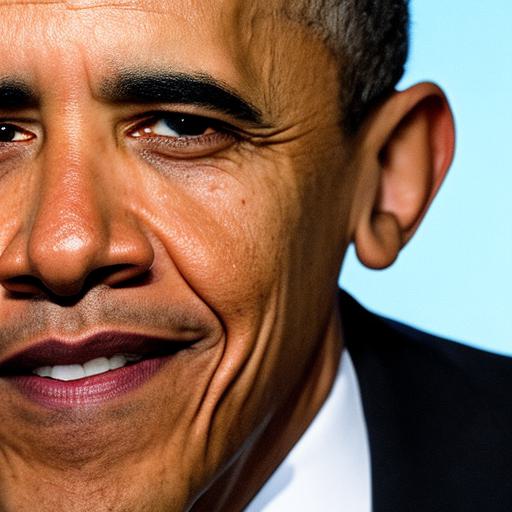} &
        \includegraphics[width=0.1\textwidth]{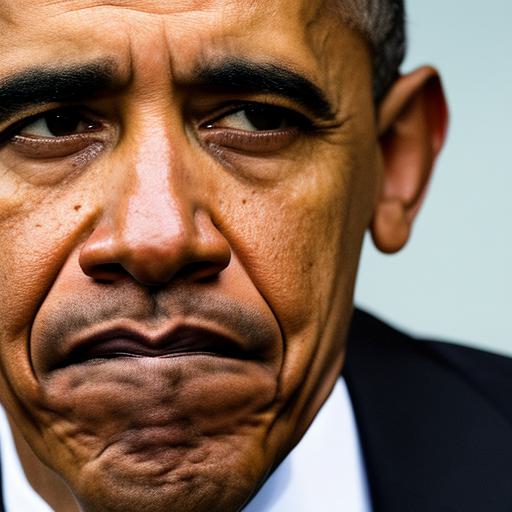} &
        \includegraphics[width=0.1\textwidth]{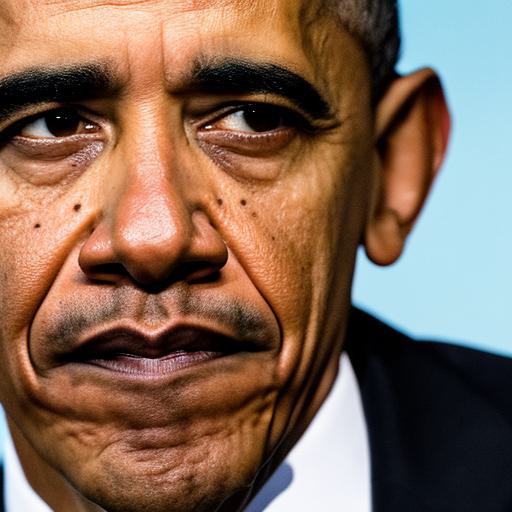} &
        \includegraphics[width=0.1\textwidth]{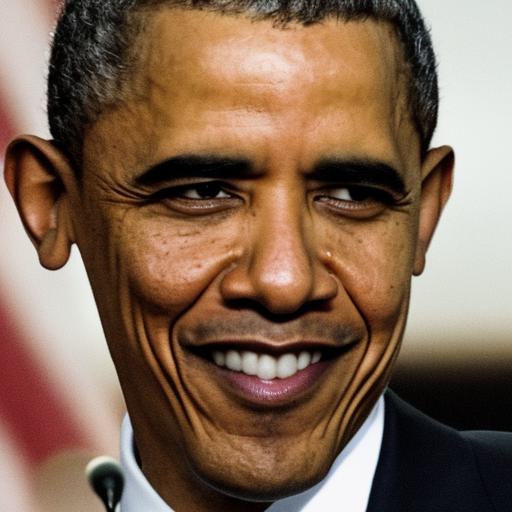} &
        \includegraphics[width=0.1\textwidth]{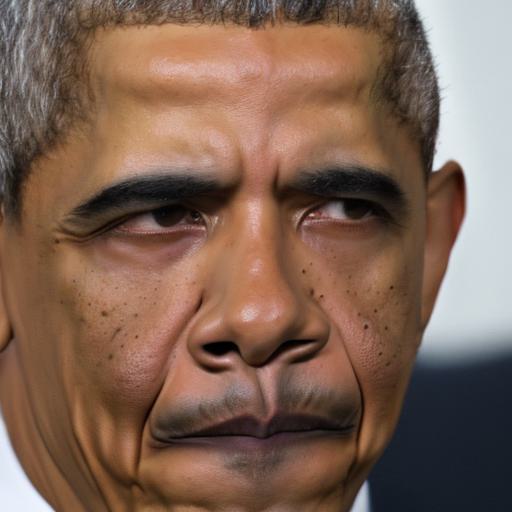} &
        \includegraphics[width=0.1\textwidth]{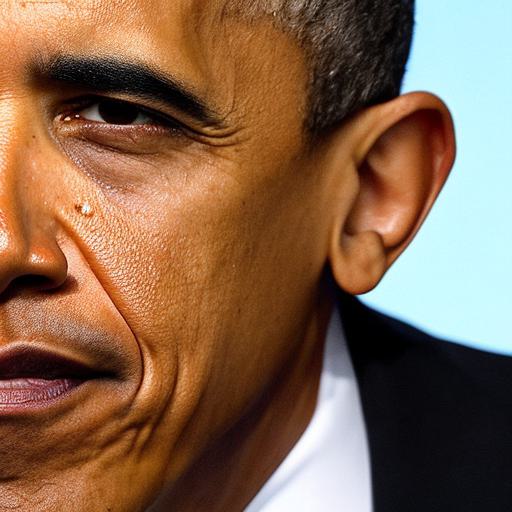}
        \\
        \multicolumn{5}{c}{\qcr{AU1+2+5+25+26}} & \multicolumn{5}{c}{\qcr{AU4+6+17+20}}  \\
\end{tabular}
    \end{subfigure}%
\end{center}
\caption{
Comparison of methods on combination AUs with the prompt \textit{A close-up of Barack Obama}.
}
\label{fig:au_combinations2}
\vspace{-3mm}
\end{figure*}

\begin{figure*}
\begin{center}
    \setlength{\tabcolsep}{1pt}
    \begin{subfigure}{0.95\textwidth}
    \hspace{-6mm}
    \begin{tabular}{*{5}{c}@{\hskip 8pt} *{5}{c}}
        \multicolumn{5}{c}{\qcr{AU6+12+25}} & \multicolumn{5}{c}{\qcr{AU4+15+17}}  \\
        \includegraphics[width=0.1\textwidth]{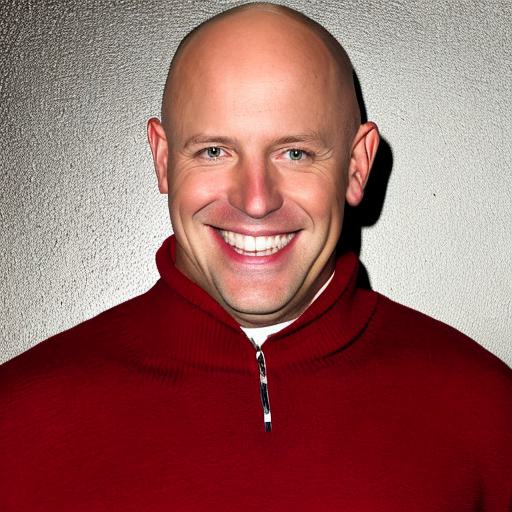} &
        \includegraphics[width=0.1\textwidth]{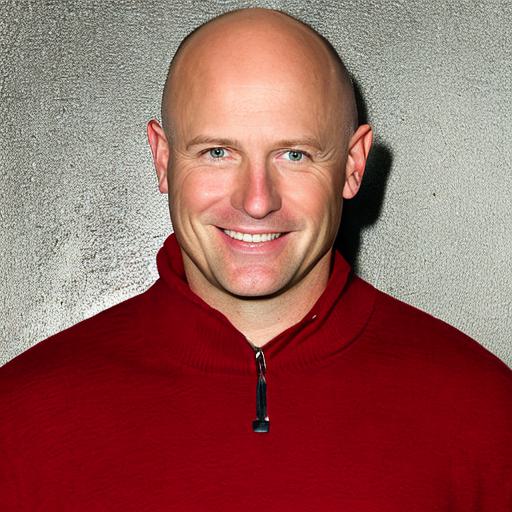} &
        \includegraphics[width=0.1\textwidth]{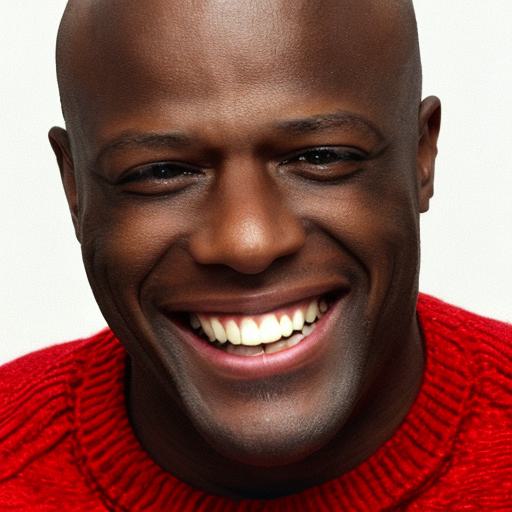} &
        \includegraphics[width=0.1\textwidth]{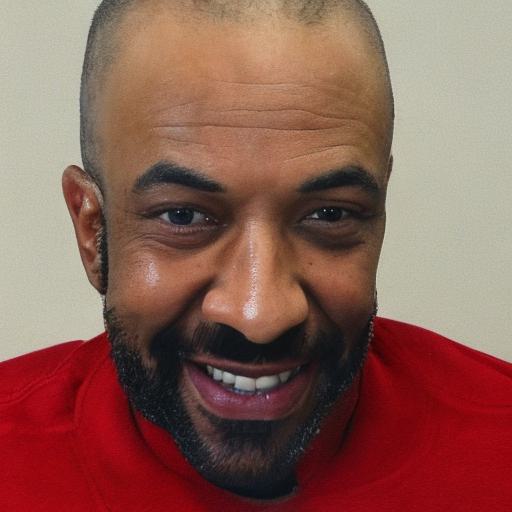} &
        \includegraphics[width=0.1\textwidth]{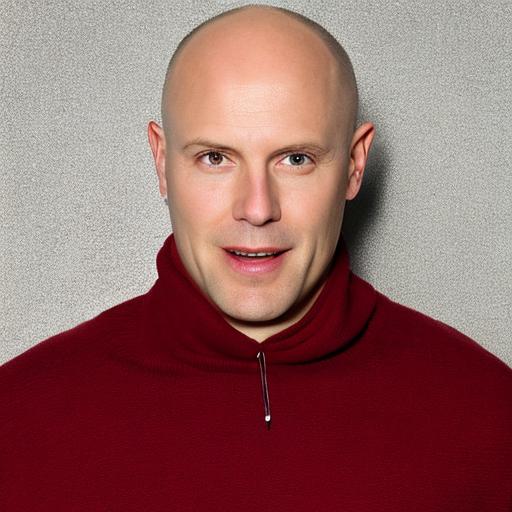} &
        \includegraphics[width=0.1\textwidth]{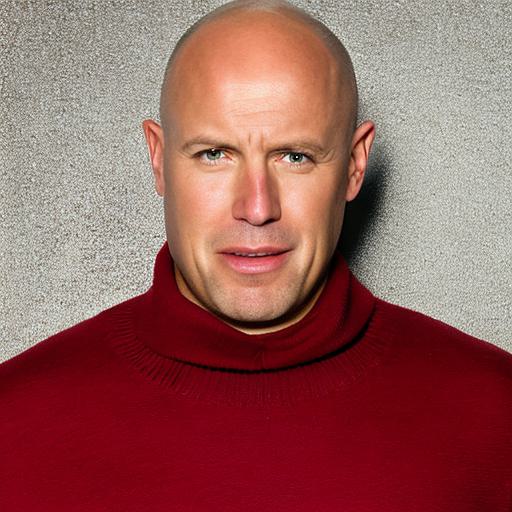} &
        \includegraphics[width=0.1\textwidth]{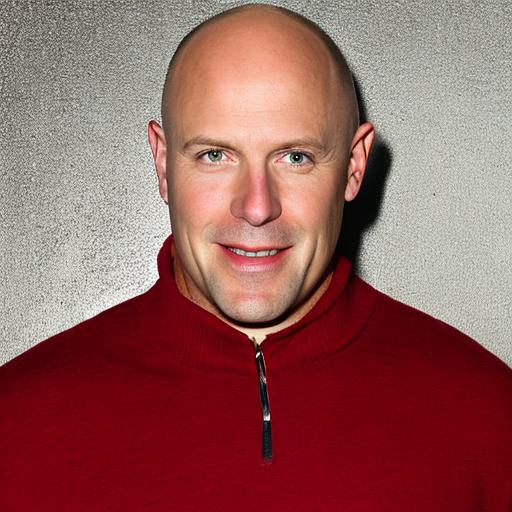} &
        \includegraphics[width=0.1\textwidth]{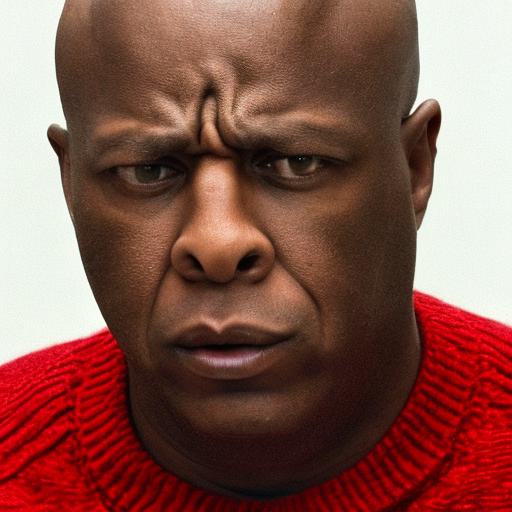} &
        \includegraphics[width=0.1\textwidth]{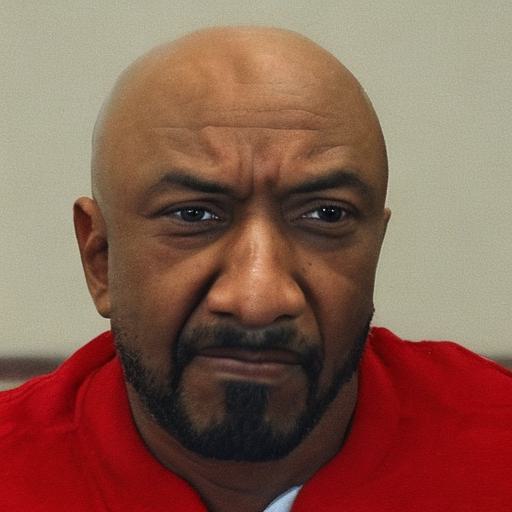} &
        \includegraphics[width=0.1\textwidth]{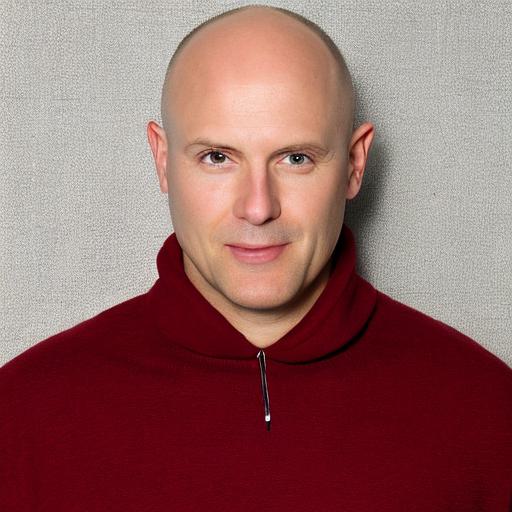}
        \\
        \multicolumn{5}{c}{\qcr{AU1+2+5+25+26}} & \multicolumn{5}{c}{\qcr{AU4+6+17+20}}  \\
\end{tabular}
    \end{subfigure}%
\end{center}
\caption{
Comparison of methods on combination AUs with the prompt \textit{A caucasian man with a bald head wearing a red sweater}.
}
\label{fig:au_combinations3}
\vspace{-3mm}
\end{figure*}

\begin{figure*}
\begin{center}
    \setlength{\tabcolsep}{1pt}
    \begin{subfigure}{0.91\textwidth}
    \hspace{-2mm}
    \begin{tabular}{*{6}{c}}
        \multicolumn{6}{c}{\qcr{AU1}}  \\
        \\
        \includegraphics[width=0.1666\textwidth]{figures/aus/resmlp64/obama/AU1/0.jpg} &
        \includegraphics[width=0.1666\textwidth]{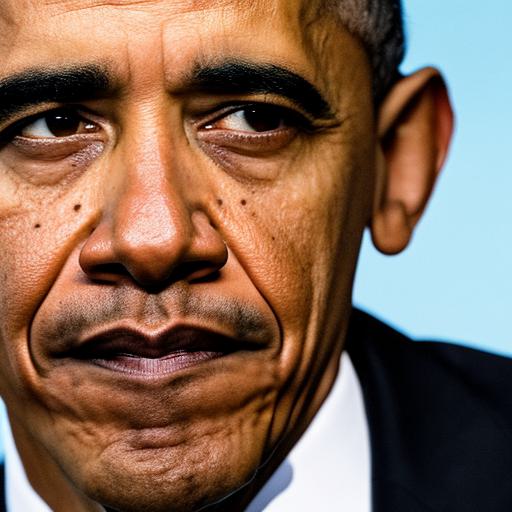} &
        \includegraphics[width=0.1666\textwidth]{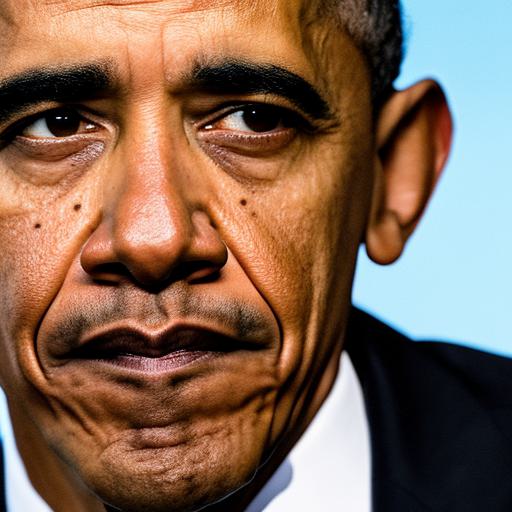} &
        \includegraphics[width=0.1666\textwidth]{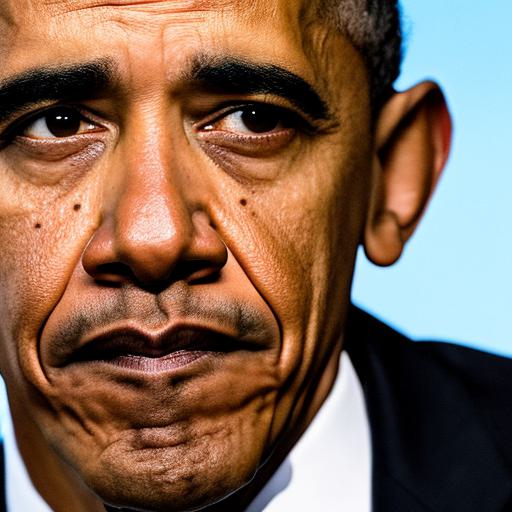} &
        \includegraphics[width=0.1666\textwidth]{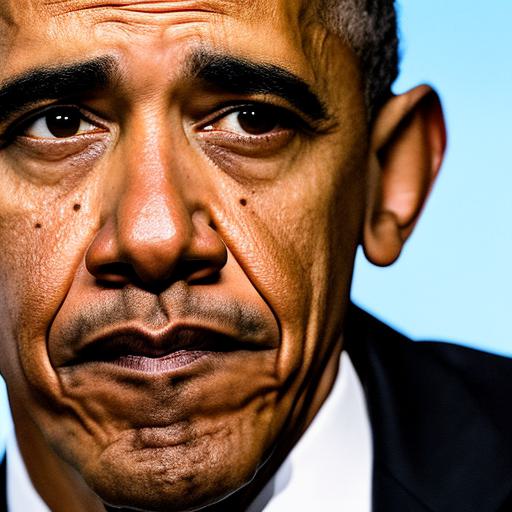} &
        \includegraphics[width=0.1666\textwidth]{figures/aus/resmlp64/obama/AU1/5.jpg}
        \\
        \multicolumn{6}{c}{\qcr{AU2}}  \\
        \\
        \includegraphics[width=0.1666\textwidth]{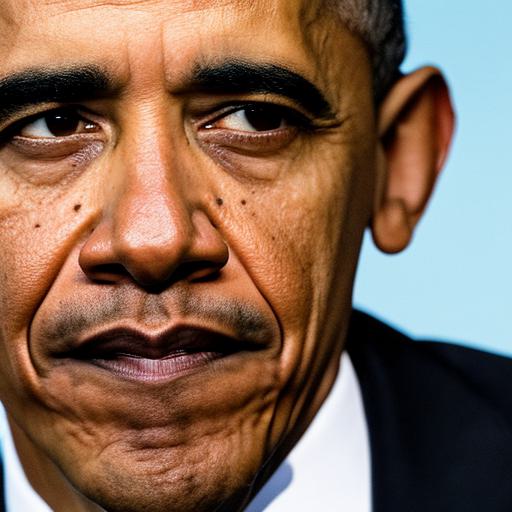} &
        \includegraphics[width=0.1666\textwidth]{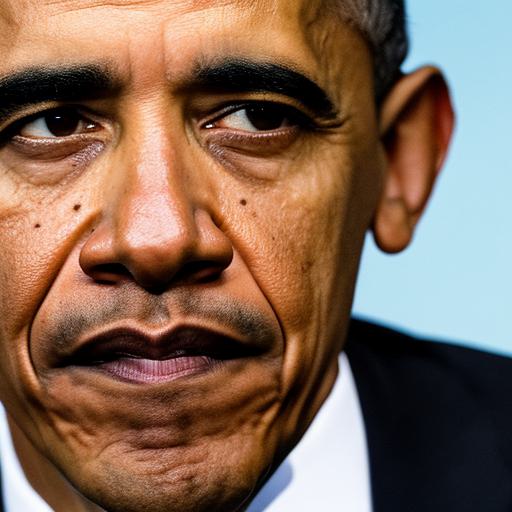} &
        \includegraphics[width=0.1666\textwidth]{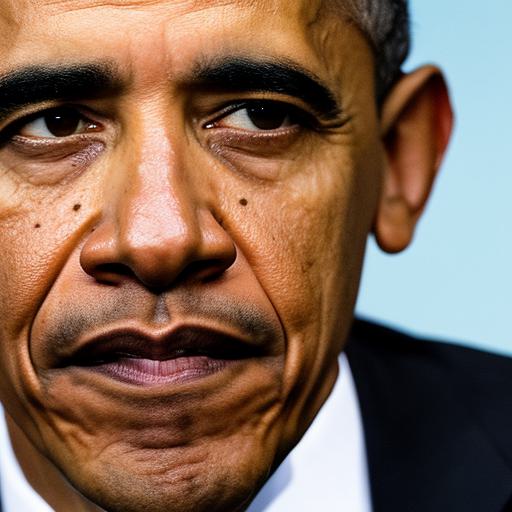} &
        \includegraphics[width=0.1666\textwidth]{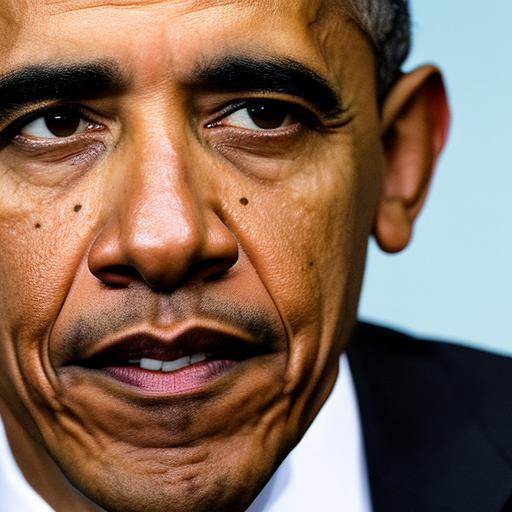} &
        \includegraphics[width=0.1666\textwidth]{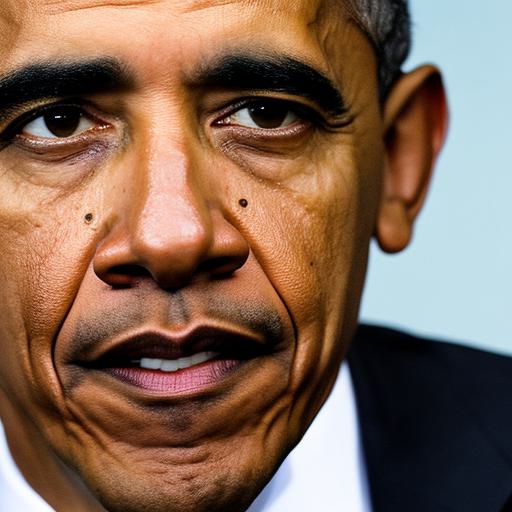} &
        \includegraphics[width=0.1666\textwidth]{figures/aus/resmlp64/obama/AU2/5.jpg}
        \\
        \multicolumn{6}{c}{\qcr{AU4}}  \\
        \\
\end{tabular}
    \end{subfigure}%
\end{center}
\caption{
AU intensity scale from zero to five scale for individual AUs from \qcr{AU1} to \qcr{AU9}.
}
\label{fig:au_intensity2}
\vspace{-3mm}
\end{figure*}

\begin{figure*}
\begin{center}
    \setlength{\tabcolsep}{1pt}
    \begin{subfigure}{0.91\textwidth}
    \hspace{-2mm}
    \begin{tabular}{*{6}{c}}
        \multicolumn{6}{c}{\qcr{AU12}}  \\
        \\
        \includegraphics[width=0.1666\textwidth]{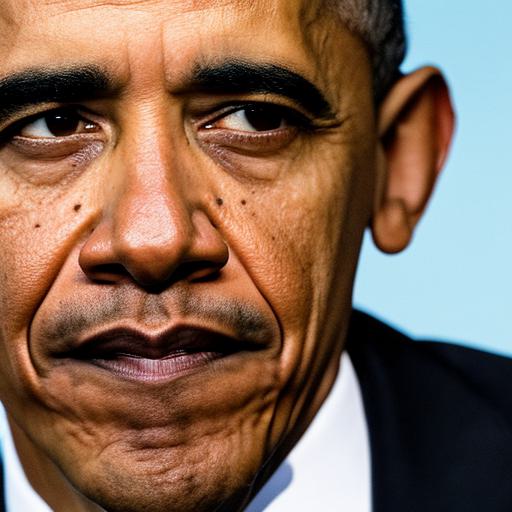} &
        \includegraphics[width=0.1666\textwidth]{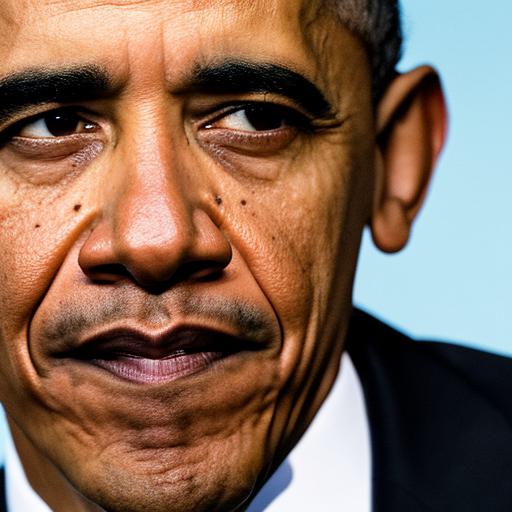} &
        \includegraphics[width=0.1666\textwidth]{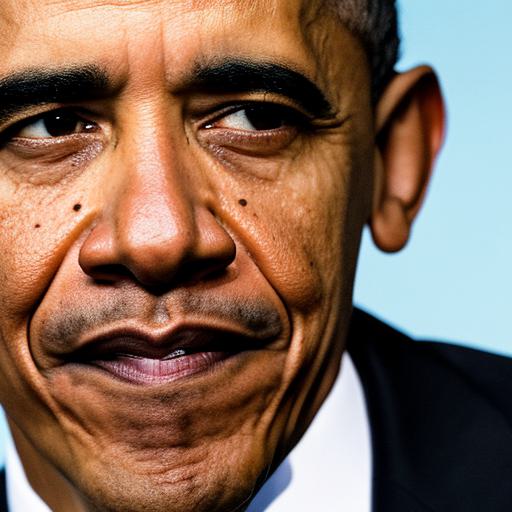} &
        \includegraphics[width=0.1666\textwidth]{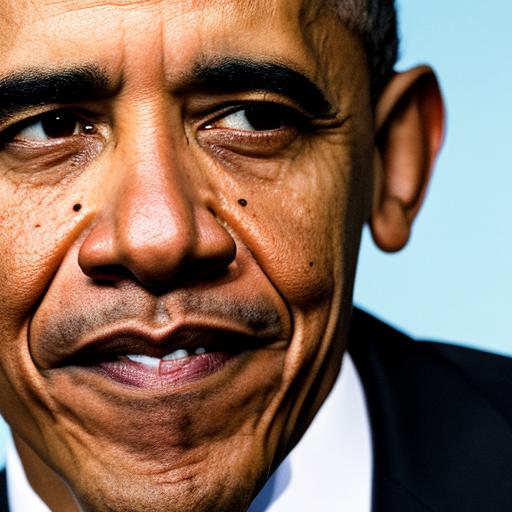} &
        \includegraphics[width=0.1666\textwidth]{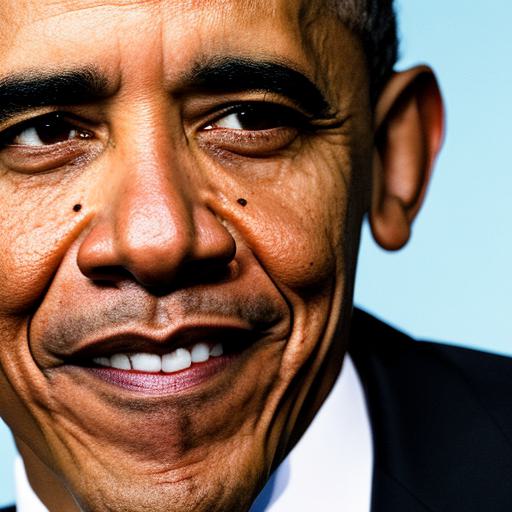} &
        \includegraphics[width=0.1666\textwidth]{figures/aus/resmlp64/obama/AU12/5.jpg}
        \\
        \multicolumn{6}{c}{\qcr{AU15}}  \\
        \\
        \includegraphics[width=0.1666\textwidth]{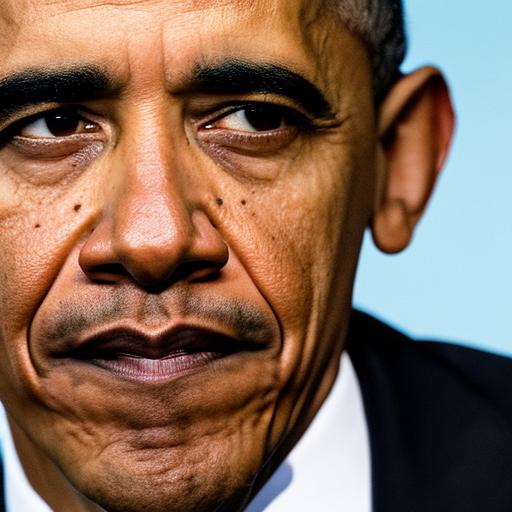} &
        \includegraphics[width=0.1666\textwidth]{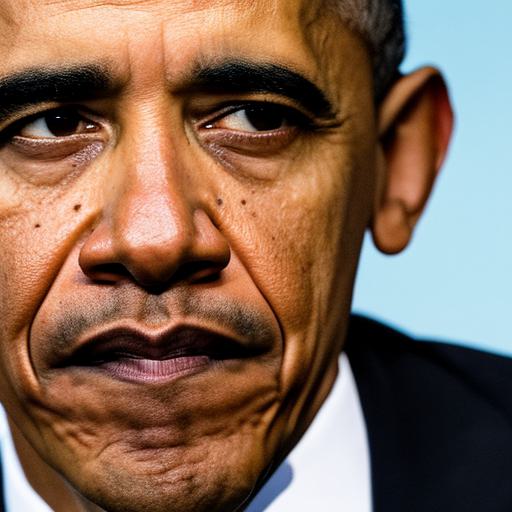} &
        \includegraphics[width=0.1666\textwidth]{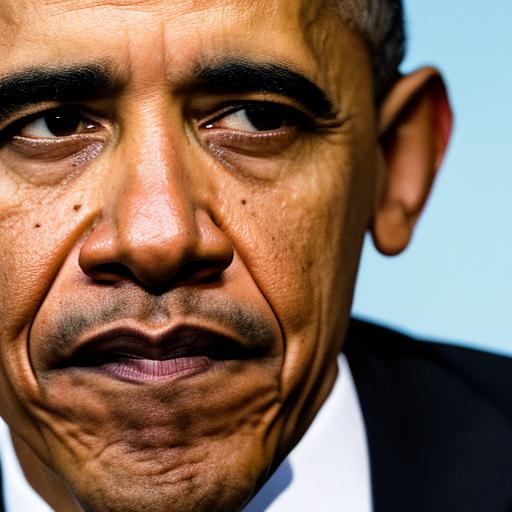} &
        \includegraphics[width=0.1666\textwidth]{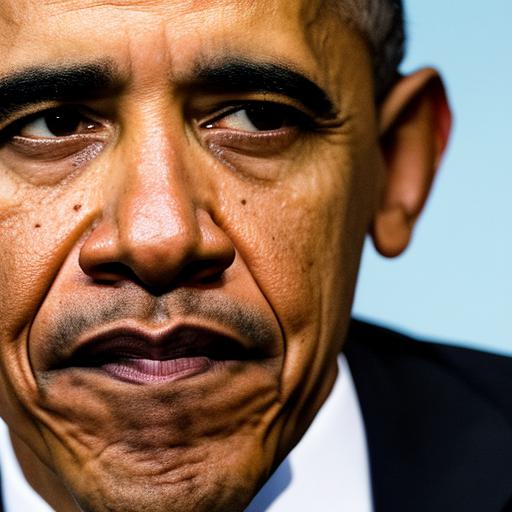} &
        \includegraphics[width=0.1666\textwidth]{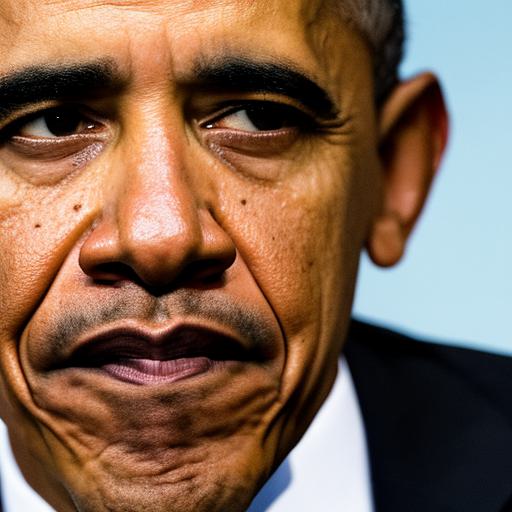} &
        \includegraphics[width=0.1666\textwidth]{figures/aus/resmlp64/obama/AU15/5.jpg}
        \\
        \multicolumn{6}{c}{\qcr{AU17}}  \\
        \\
\end{tabular}
    \end{subfigure}%
\end{center}
\caption{
AU intensity scale from zero to five scale for individual AUs from \qcr{AU12} to \qcr{AU26}.
}
\label{fig:au_intensity3}
\vspace{-3mm}
\end{figure*}

More qualitative results comparing the baseline methods with individual AUs are shown in \cref{fig:au_comparison2,fig:au_comparison3}. For a comparison of combination AUs see \cref{fig:au_combinations2,fig:au_combinations3}. \Cref{fig:au_intensity2,fig:au_intensity3} show continuity of the scaling within the $[0, 5]$ range for individual AUs.

\section*{Societal Impact}
Machine learning models can learn biases from their datasets. This is especially true for human faces and facial expressions where ethnicities and cultures pay a large role. By including a large dataset with subjects from a large variety of ethnicities these challenges can be mitigated. We note that since our model is built upon previous models, it inherits any biases these models may contain. Malicious users may want to mislead viewers with generated images, which is a common issue with existing similar methods. However, recent approaches in detecting fake generated imagery are improving quickly.

\end{document}